\documentclass[11pt,a4paper]{article}

\usepackage[utf8]{inputenc}
\usepackage[T1]{fontenc}
\usepackage{lmodern}
\usepackage[english]{babel}

\usepackage{setspace}
\usepackage{parskip}

\usepackage{amsmath}
\usepackage{amsfonts}
\usepackage{amssymb}
\usepackage{mathtools}

\usepackage{graphicx}
\usepackage{float}
\usepackage{subcaption}

\usepackage{tikz}
\usetikzlibrary{positioning,shapes,arrows,arrows.meta,calc,fit,backgrounds}

\usepackage{pgfplots}
\pgfplotsset{compat=1.18}

\usepackage{booktabs}
\usepackage{array}
\usepackage{multirow}
\usepackage{tabularx}

\usepackage{algorithm}
\usepackage{algorithmic}

\usepackage[backend=biber,style=authoryear,url=true,doi=false,eprint=false,maxnames=99,maxcitenames=2]{biblatex}

\DeclareSourcemap{
  \maps[datatype=bibtex]{
    \map{
      \step[fieldset=file, null]
      \step[fieldset=keywords, null]
      \step[fieldset=abstract, null]
      \step[fieldset=annotation, null]
      \step[fieldset=note, null]
    }
  }
}

\usepackage[colorlinks=true, linkcolor=blue, citecolor=blue, urlcolor=blue]{hyperref}
\usepackage{cleveref}

\usepackage{listings}
\usepackage{geometry}

\AtEveryBibitem{%
  \clearfield{keywords}%
  \clearfield{note}%
  \clearfield{pagetotal}%
  \clearfield{pages}%
  \clearfield{isbn}%
  \clearfield{issn}%
  \clearfield{archivePrefix}%
  \clearfield{eprint}%
  \clearfield{primaryClass}%
  \clearname{editor}%
  \clearfield{urldate}%
}

\DeclareFieldFormat{citehyperref}{%
  \DeclareFieldAlias{bibhyperref}{noformat}%
  \bibhyperref{#1}}

\DeclareCiteCommand{\cite}[\mkbibparens]
  {\usebibmacro{prenote}}
  {\usebibmacro{citeindex}%
   \printtext[citehyperref]{\usebibmacro{cite}}}
  {\multicitedelim}
  {\usebibmacro{postnote}}

\renewcommand{\mkbibparens}[1]{\mkbibbrackets{#1}}

\AtBeginBibliography{\emergencystretch=3em}

\usepackage{xurl}
\makeatletter
\let\allot@origtexttt\texttt
\renewcommand{\texttt}[1]{\allot@origtexttt{\allot@ttbreak#1\@nil}}
\def\allot@ttbreak#1\@nil{\begingroup\def\_{\char`\_\discretionary{}{}{}}\allot@ttscan#1\@nil\endgroup}
\def\allot@ttscan#1\@nil{#1}
\makeatother

\newcommand{\axA}{A1}

\newcommand{\axG}{A7}

\usepackage{xurl}
\graphicspath{{figures/}}

\title{AllotNet: an Interpretable Mixture-of-Experts Choice Model for Moral Judgement in Humanitarian Aid Allocation, Validated against a Planted Synthetic Population}
\author{Mark Nowak\\Conflux Laboratory\\ORCID 0009-0000-2233-1714}
\date{15 November 2026\\\small Public version. The elicitation instrument's scoring specification is withheld under a pre-registration embargo; see Section~\ref{sec:scope}.\\Licensed under CC BY 4.0.}

\begin{document}

\maketitle

\begin{abstract}\small
Official development assistance fell by 23.1 per cent in 2025, and the decision of who is helped first when there is not enough for everyone is moving into software.
A leading published machine-learning targeting system excluded 53 per cent of eligible households as deployed.
If the rule of distribution ends up in code, it should carry the judgement of many cultures, and the machine should execute while a person judges.
AllotNet is an interpretable mixture-of-experts conditional logit for the allocation judgements elicited by Allot, a pre-registered six-language instrument: eight linear taste experts over seven ethical criteria, weighted by a gate over participant context, with a damped residual, an encoder for household intake, an outcome head and a budget-constrained planner that builds a review queue and never rejects anyone.
A model of human judgement has no answer key on real data, so we wrote one first: a planted synthetic population of 5.6 million sessions and 104 million responses (98 million usable choices), generated by the production code path from a ground truth hidden from the loss.
All sixteen pre-specified checks passed: criterion ranking $\rho = 1.000$, hidden segments recovered (ARI 0.671), outcome head at 99 per cent of the noise ceiling, planner at 99.4 per cent of oracle.
The caveats are reported with them: two checks are weaker than their titles, and a live-generation failure cost 260,712 seeds.
The model has 23,550 parameters, is trained quantisation-aware for NVFP4 with its interpretable core in fp32 (welfare-rank parity $\rho = 0.9937$ at W4A16 on the shipped bundle; one parity threshold was relaxed post hoc), and ships as a 125~KB bundle with C11, Java, Rust and Python runtimes.
Nothing has been trained on real data yet; at the study's real scale a hierarchical Bayesian mixed logit is the registered comparator.

\end{abstract}

\section{Introduction}
\label{sec:introduction}

This paper is about one decision: who gets help first when there is not enough for everyone.
Aid budgets have contracted sharply, and the decision is moving into software, where two dangers wait for it.
One is cold cost-effectiveness, which finds the most vulnerable not worth helping; the other is the in-group favouritism people bring with or without a machine.
We do not argue that the decision should be kept out of code, since at today's budgets prioritisation at scale has no manual alternative.
We argue that a rule of distribution written in code should carry the judgement of many cultures rather than the preference of whoever wrote it, and that the machine should execute while a person judges.
This paper builds and tests the model half of that programme: a small, readable choice model of allocation judgement, validated against a population whose answers were fixed before the model saw any data.

\subsection{Allocation under contraction}
\label{sec:contraction}

Official development assistance from the members of the OECD Development Assistance Committee fell by 23.1 per cent in real terms in 2025, to USD~174.3 billion, or 0.26 per cent of the members' combined gross national income \cite{oecdHistoricDeclineForeign2026,oecdPreliminaryOfficialDevelopment2026}.
In the OECD's words, it is ``the largest annual contraction recorded in the history of ODA''.
Humanitarian ODA fell by 35.8 per cent, and the OECD projects a further 5.8 per cent decline for 2026 \cite{oecdODAProjections20262026}.
The 2025 figures are preliminary, released on 9 April 2026, and are quoted here as reported; the final statistics may revise them.

The appeals tell the same story from the receiving end.
The Global Humanitarian Overview 2026 counts 239 million people in need, targets 135 million of them and requests USD~33 billion; its hyper-prioritised core plan narrows the target to 87 million people for USD~23 billion \cite{ochaGlobalHumanitarianOverview2025,ochaLifeLifeLaunches2025}.
The 2025 appeals received USD~12 billion of the USD~47.4 billion requested, which OCHA calls ``the lowest in a decade''.
Less money, not less need: the smaller in-need figure ``reflects more focused analysis, rather than a reduction in suffering'' \cite{ochaGlobalHumanitarianOverview2025,alnapStateHumanitarianSystem2026}.
Every mistake now costs more.

\subsection{The hardest decision is moving into software}
\label{sec:two-dangers}

The hardest decision in the sector is who gets help first when there is not enough for everyone.
That decision is increasingly encoded: in proxy means tests, in vulnerability scores, in machine-learning models that rank households from mobile-phone data \cite{aikenMachineLearningPhone2022,hannaUniversalBasicIncomes2018,athiasHitMissAssessment2020}.
With reduced budgets, prioritisation at this scale has no manual alternative.
Two dangers wait here.

The first is cold logic.
A machine without moral values concludes that helping the most vulnerable is `not cost-effective'.

The second is us, and it needs no machine.
We help our own tribe more willingly: our culture, our language, people who look like us.
Meta-analysis puts in-group bias at $d = 0.32$, 0.19 without interdependence and 0.42 with it \cite{ballietIngroupFavoritismCooperation2014}; arbitrary group labels suffice to produce the effect \cite{tajfelExperimentsIntergroupDiscrimination1970,tajfelIntegrativeTheoryIntergroup1979}.
The quiet ones stay invisible.

Nor does the machine, so far, do much better.
The best published machine targeting, the Novissi programme in Togo, ranked households from phone data to reach the poorest 29 per cent.
In the rural evaluation precision and recall were both 47 per cent, an exclusion error of 53 per cent as deployed, and 50 per cent in the nationwide simulation \cite{aikenMachineLearningPhone2022}.
A fully calibrated proxy means test, the best feasible alternative, would have excluded about 37 per cent, but it requires a comprehensive social registry that Togo lacked; the government's geographic options excluded 55 to 67 per cent.
Across 42 poverty-targeted schemes in 25 countries no scheme had an exclusion error below 44 per cent \cite{athiasHitMissAssessment2020}; the two errors of targeting, exclusion and inclusion, were named three decades ago \cite{corniaTwoErrorsTargeting1993}.

Both are unacceptable.
If the rule of distribution ends up in code anyway, let it carry the judgement of many cultures, not the personal preference of whoever wrote the code.
The machine gets the execution, not the judgement.

\subsection{Machine executes, human judges}
\label{sec:governance-stance}

The governance stance follows from the two dangers, and it was stated as one rule at DIHAD 2026 and on the laboratory's website: ``the system may build a queue for human review. It may not reject anyone on its own.''
Technology changes our work fastest where a mistake costs a wasted trip.
Where a mistake costs a person their aid, it must change slowly.
The last word stays with a human being, a human one can ask ``why?''.
Figure~\ref{fig:hitl} sketches the resulting division of labour: the machine ranks, plans and builds the queue; the person judges, vetoes, sets the weights $\beta$, selects the mode and asks why.

\ifdefined\allottikzstyles\else
\def\allottikzstyles{loaded}%
\definecolor{mainblue}{rgb}{0.2, 0.4, 0.7}
\definecolor{accentred}{rgb}{0.8, 0.2, 0.2}
\definecolor{softgreen}{rgb}{0.2, 0.6, 0.3}
\definecolor{warmgray}{rgb}{0.5, 0.5, 0.5}
\tikzset{
  block/.style={rectangle, draw=mainblue!70!black, fill=mainblue!10, rounded corners=2pt,
    minimum height=2.4em, minimum width=3em, align=center, inner sep=3pt, font=\footnotesize},
  data/.style={trapezium, trapezium left angle=80, trapezium right angle=100,
    trapezium stretches=true, draw=softgreen!70!black, fill=softgreen!12,
    minimum height=2.4em, align=center, inner sep=3pt, font=\footnotesize},
  decision/.style={diamond, draw=warmgray!80!black, fill=yellow!15, aspect=2,
    align=center, inner sep=1pt, font=\footnotesize},
  heldback/.style={rectangle, draw=accentred, dashed, fill=accentred!6, rounded corners=2pt,
    minimum height=2.4em, align=center, inner sep=3pt, font=\footnotesize},
  config/.style={rectangle, draw=warmgray!80!black, fill=warmgray!12, rounded corners=2pt,
    minimum height=2.4em, align=center, inner sep=3pt, font=\footnotesize},
  group/.style={rectangle, draw=warmgray, dashed, rounded corners=4pt, inner sep=6pt},
  solidgroup/.style={rectangle, draw=warmgray, rounded corners=4pt, inner sep=6pt},
  bluegroup/.style={rectangle, draw=mainblue, rounded corners=4pt, inner sep=6pt, fill=mainblue!4},
  graygroup/.style={rectangle, draw=warmgray, rounded corners=4pt, inner sep=6pt, fill=warmgray!8},
  grouplabel/.style={font=\footnotesize\itshape, text=warmgray!40!black, inner sep=2pt},
  annotation/.style={font=\footnotesize\itshape, text=warmgray!40!black, align=center},
  note/.style={font=\scriptsize\itshape, text=warmgray!40!black, align=left},
  flag/.style={font=\scriptsize\bfseries, text=accentred, align=left},
  edgelabel/.style={font=\scriptsize, fill=white, inner sep=1.5pt, align=center},
  arrow/.style={-{Stealth[length=2mm]}, thick, draw=black!75},
  dashedarrow/.style={-{Stealth[length=2mm]}, thick, dashed, draw=black!75},
  biarrow/.style={{Stealth[length=2mm]}-{Stealth[length=2mm]}, thick, draw=black!75},
  cut/.style={draw=accentred, dashed, thick},
}
\newsavebox{\allotfigbox}
\newenvironment{allotfit}{\begin{lrbox}{\allotfigbox}}{\end{lrbox}%
  \typeout{ALLOTFIT natural width \the\wd\allotfigbox\space (limit \the\dimexpr0.95\textwidth\relax)}%
  \ifdim\wd\allotfigbox>0.95\textwidth
    \typeout{ALLOTFIT scaled down to 0.95\string\textwidth}%
    \resizebox{0.95\textwidth}{!}{\usebox{\allotfigbox}}%
  \else
    \usebox{\allotfigbox}%
  \fi}
\fi

\begin{figure}[htbp]
  \centering
  \begin{allotfit}
  \begin{tikzpicture}[node distance=1.0cm and 0.6cm]
    \node[data, text width=2.2cm] (intake) {Intake data\\household forms;\\budget $B$};
    \node[block, text width=3.0cm, right=of intake] (machine) {\textbf{MACHINE}\\ranks by clean score;\\predicts outcomes;\\builds a\\review queue;\\never rejects};
    \node[block, draw=softgreen!70!black, fill=softgreen!12, text width=3.0cm, right=1.5cm of machine] (human) {\textbf{HUMAN}\\judges and vetoes;\\sets $\beta$ and the clean mode;\\answers ``why?''};
    \node[block, text width=2.5cm, right=of human] (decision) {Distribution decision\\(who, how much,\\how often, how)};
    \node[config, text width=3.2cm] (config) at ($(machine.south)!0.5!(human.south)+(0,-1.2)$) {Configuration (set by the human)\\$\beta$, clean mode, budget $B$};
    \draw[arrow] (intake) -- (machine);
    \draw[biarrow] (machine) -- node[edgelabel, above] {queue\\$\leftrightarrow$ verdicts} (human);
    \draw[arrow] (human) -- (decision);
    \draw[arrow] (human.south) |- (config.east);
    \draw[arrow] (config.west) -| (machine.south);
    \node[edgelabel] at ($(config.west)!0.5!(config.west -| machine.south)+(0,0.3)$) {constrains};
    \node[edgelabel] at ($(config.east)!0.5!(config.east -| human.south)+(0,0.3)$) {feedback};
  \end{tikzpicture}
  \end{allotfit}
  \caption{Human-in-the-loop division of labour.
    The machine ranks households by their clean-mode score, predicts outcomes, and assembles a queue for review; it never rejects.
    A human judges and vetoes, sets $\beta$ and the clean mode, and can ask `why?' of any decision because the weights are directly readable.
    The governing rule: ``the system may build a queue for human review; it may not reject anyone on its own.''}
  \label{fig:hitl}
\end{figure}
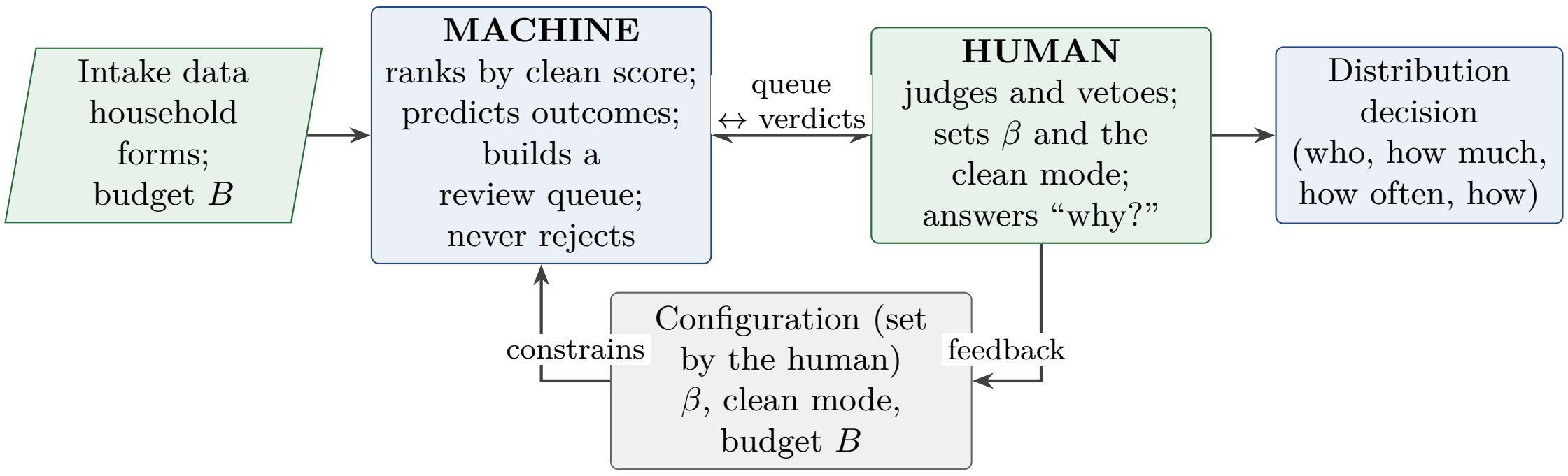

The legal anchor is Regulation (EU) 2024/1689.
Annex III, point 5(a), classifies as high-risk the AI systems ``intended to be used by public authorities or on behalf of public authorities to evaluate the eligibility of natural persons for essential public assistance benefits and services'', and Article 14(1)--(2) requires effective human oversight of such systems \cite{eu_RegulationEU20242024}.
Point 5(a) is scoped to public authorities and to those acting on their behalf, so a tool run by a non-governmental organisation on its own caseload falls squarely within it only when the organisation acts on behalf of a public authority.
We adopt the obligations by design regardless (Section~\ref{sec:discussion-governance}).
We do not treat human oversight as a guarantee either; the literature on the limits of oversight policies \cite{greenFlawsPoliciesRequiring2022,greenDisparateInteractions2019} and on contestability \cite{lyonsConceptualisingContestability2021} shapes how the queue is presented to the person who works through it.

\subsection{The problem this paper solves: no answer key}
\label{sec:no-answer-key}

A model of human allocation judgement faces a problem that most supervised models do not.
``On real preference data there is no answer key: any model output is a hypothesis.''
A model trained on real allocation decisions that reports that one criterion outweighs another cannot be checked against anything, and a model that produces plausible hypotheses is then indistinguishable from one that produces confident nonsense.

So the answer key was written first.
A ground truth fixes a synthetic population's hidden structure: latent moral types, regional cohorts and deliberately planted biases.
A generator plays that population through the production code path of the elicitation instrument.
The model never sees the ground truth and is scored on whether it recovers numbers that were decided in advance.
This is blind analysis and pre-registration applied to a model rather than to a hypothesis \cite{maccounBlindAnalysisHide2015,nosekPreregistrationRevolution2018}, and planted-truth validation is the only way to test the instrument before the data exist.

\subsection{Contributions}
\label{sec:contributions}

Four contributions follow.
\begin{enumerate}
  \item \textbf{AllotNet}, an interpretable mixture-of-experts conditional-logit model: $K = 8$ linear taste experts over the instrument's seven criteria, a gate over participant context and a damped residual, 23{,}550 parameters in all, together with an encoder that maps standard household intake data onto the seven criteria, an outcome head and a budget-constrained cycle planner that builds a review queue and never rejects anyone (Section~\ref{sec:allotnet}).
  \item \textbf{Validation against a planted synthetic population} of 5.6 million sessions and 104 million responses (98 million usable choices), generated by the production code path from a ground truth written down before training and hidden from the loss.
  Sixteen pre-specified checks all passed, with a world control run before training, and every caveat is disclosed (Sections~\ref{sec:synthetic}, \ref{sec:evaluation} and~\ref{sec:results}).
  \item \textbf{Deployable artefacts}: NVFP4 quantisation-aware training with the interpretable core kept in fp32, and a 125~KB portable bundle with C11, Java, Rust and Python runtimes verified against golden vectors on six machines and benchmarked under one protocol (Sections~\ref{sec:allotnet-nvfp4}, \ref{sec:allotnet-bundle} and~\ref{sec:results-bench}).
  On four open household surveys the clean score agrees with the directly observed food-consumption score at $\rho_s$ 0.70 to 0.80 on three files and 0.29 where income is wholly imputed (Section~\ref{sec:results-realdata}); the food-consumption score is itself an input to the clean score, and the data-quality weights were chosen after the first run on the same surveys, so the agreement is partly built in and is not an out-of-sample test.
  \item \textbf{An honest boundary.}
  The pre-launch traffic of the instrument holds 46 sessions and 206 responses from 12 people, six of whom completed the protocol; all of it is quarantined as internal testing.
  Nothing has been trained on real data, and at the study's real scale a hierarchical Bayesian mixed logit is the registered comparator (Section~\ref{sec:discussion-scale}).
\end{enumerate}

A companion paper reports the designed society simulation built on the same synthetic households and the comparison of allocation rules run in it; its results are not repeated here.

\subsection{Embargo and scope}
\label{sec:scope}

Allot is pre-registered on OSF, under embargo until 1 August 2030.
The embargo exists so that the criteria, the planted ground truth and the scenario scoring do not prime future participants: participants who know the scoring key are no longer naïve, and non-naïveté, crosstalk between participants and order effects all distort responses to moral dilemmas \cite{chandlerNonnaiveteAmazonMechanical2014,edlundPrevalencePreventionCrosstalk2014,wiegmannOrderEffectsMoral2012}.
The documented rationale for OSF embargoes is protection against scooping \cite{helpWelcomeRegistrationsPreregistrations2026}; the anti-priming justification is our own.

This public version therefore follows three rules.
The seven criteria are referred to as \axA{} to \axG{} and never named; nothing in the paper reports a per-criterion weight, a per-criterion ranking or which criterion came out highest.
The planted values of the synthetic population, the instrument's scoring and session mechanics, and the recovered per-criterion and per-expert values are withheld; the checks are described by what they test, not by the direction or size of the planted effect, and recovery is reported through aggregate statistics.
All of it appears in the full version after the embargo, together with the instrument specification and the first dataset.
The numbers that are reported here are properties of the model: rank and cosine agreement, segment-recovery indices, outcome-head accuracy against its ceiling, planner efficiency, quantisation parity and runtime measurements.

On provenance: the framing of this introduction follows the author's featured presentation at DIHAD 2026 in Dubai on 24 August 2026, ``The Intersection of Humanitarian Shifts and Technology'', given in his capacity as Chief Technology Officer of OMNIAID and covering the laboratory's Allot and AllotNet work \cite{programme2026ConferenceProgram2026}.
The laboratory's website attributes the quoted lines to a panel on 25 August whose programme entry does not list the author; that attribution is a website error, and no recording of either session was found.

Section~\ref{sec:related} reviews the background; Section~\ref{sec:allot} describes the elicitation instrument at the level already public; Section~\ref{sec:allotnet} describes AllotNet; Section~\ref{sec:synthetic} covers the synthetic population and training; Section~\ref{sec:evaluation} the evaluation protocol; Section~\ref{sec:results} the results; Section~\ref{sec:discussion} the discussion; Section~\ref{sec:conclusion} concludes, and the appendices hold the supporting tables.

\section{Background and related work}
\label{sec:related}

\subsection{Targeting and cash}
\label{sec:targeting}

Targeting has a literature of its own, and it is largely a literature of error rates.
Proxy means tests (PMTs) predict household welfare from observable characteristics and score households against a cut-off; their accuracy has been examined in detail, with sobering results \cite{brownPoorMeansTest2018,coadyTargetingTransfersDeveloping2004,groshRevisitingTargetingSocial2022}.
Community-based targeting trades accuracy for legitimacy \cite{alatasTargetingPoorEvidence2012}, and the cost-effectiveness of phone-based machine learning against a PMT has since been assessed at national scale \cite{aikenScalableTargetingSocial2025}.
Humanitarian programmes have their own instruments: UNHCR's Vulnerability Assessment Framework in Jordan \cite{unhcrVulnerabilityAssessmentFramework2015} and WFP's Consolidated Approach for Reporting Indicators of Food Security \cite{wfpConsolidatedApproachReporting2021}.
The outcome indicators AllotNet's encoder and outcome head use are the standard ones: the Food Consumption Score \cite{wiesmannValidationWorldFood2009}, the reduced Coping Strategies Index \cite{maxwellMeasuringFoodInsecurity2008,caldwellCopingStrategiesIndex2008}, the minimum expenditure basket \cite{networkMinimumExpenditureBasket2020} and household equivalence scales \cite{buhmannEQUIVALENCESCALESWELLBEING1988,deatonGuidelinesConstructingConsumption2002}.
The cash-transfer evidence is by now large: randomised evaluations of unconditional transfers and their long-run follow-ups \cite{haushoferShorttermImpactUnconditional2016,shapiroLongTermImpactUnconditional2018,blattmanGeneratingSkilledSelfEmployment2014}, a review of spending on temptation goods \cite{evansCashTransfersTemptation2017}, cross-programme syntheses \cite{handaMythBustingConfrontingSix2018,al.CashTransfersWhat2016} and a direct comparison of cash with vouchers \cite{akerComparingCashVoucher2015}.

The seven criteria of the instrument are not our invention: they are drawn from established traditions of distributive justice, and the full specification, with its sources, is published with the first dataset; they are not named here for the reason given in Section~\ref{sec:scope}.
One corollary of the social-choice literature governs how they are used.
Arrow's impossibility theorem rules out an aggregation procedure that satisfies every desirable axiom at once \cite{arrowDifficultyConceptSocial1950}, so any decomposition of a choice into seven criteria is a normative choice, not a computation of `the truth'.

\subsection{Discrete choice and taste heterogeneity}
\label{sec:discrete-choice}

AllotNet is a discrete-choice model before it is anything else.
Its base is the random-utility softmax, McFadden's conditional logit, with roots in Luce's choice axiom and Thurstone's comparative judgement \cite{mcfaddenConditionalLogitAnalysis1974,luceIndividualChoiceBehavior2005,thurstoneLawComparativeJudgment1927,trainDiscreteChoiceMethods2003}.
Taste heterogeneity enters as a finite mixture of logits \cite{mcfaddenMixedMNLModels2000}, the latent-class choice model \cite{kamakuraProbabilisticChoiceModel1989,greeneLatentClassModel2003,trainEMAlgorithmsNonparametric2008}; when class membership depends on observed covariates the model becomes a concomitant-variable latent-class model \cite{daytonConcomitantVariableLatentClassModels1988,bhatEndogenousSegmentationMode1997}, which is what AllotNet's gate is.
In machine-learning terms the same object is a mixture of experts \cite{jacobsAdaptiveMixturesLocal1991,shazeerOutrageouslyLargeNeural2017}.
The damped residual follows the hybrid neural--logit models that add a learned correction to a linear utility \cite{sifringerEnhancingDiscreteChoice2020,wongResLogitResidualNeural2021,hanNeuralembeddedDiscreteChoice2022}, and RUMnets prove that a neural mixture of linear utilities is itself a random-utility model and give sample-complexity bounds for it \cite{aouadRepresentingRandomUtility2026}.
Smaller pieces have their own precedents: the permutation-equivariant scorer \cite{zaheerDeepSets2017}, the Bradley--Terry equivalence for two-option sets \cite{bradleyRankAnalysisIncomplete1952}, Gumbel-max sampling in the generator \cite{maddisonSampling2014} and the identifiability of latent-class models \cite{allmanIdentifiabilityParametersLatent2009}.

Sample size is where the small-data reality bites.
Hierarchical Bayesian mixed logit is the established alternative for small samples \cite{allenbyMarketingModelsConsumer1998,rossiBayesianStatisticsMarketing2005,reveltMixedLogitRepeated1998}.
Practitioner guidance asks for at least 500 appearances per attribute level and around 200 respondents per subgroup \cite{softwareSampleSizeIssues}, and neural choice models need at least 50 observations per network weight \cite{alwosheelYourDatasetBig2018}.
The real study will yield several thousand comparisons per stratum of the sample, several orders of magnitude fewer than the synthetic corpus, and AllotNet's residual has 2{,}353 weights, so the rule is not met at the study's scale.
This is the argument for the strong $\ell_2$ damping on the residual and for the hierarchical-Bayes comparator (Section~\ref{sec:discussion-scale}).

\subsection{Preference learning and hidden context}
\label{sec:hidden-context}

Learning a reward from pairwise preferences is now routine \cite{christianoDeepReinforcementLearning2017,ouyangTrainingLanguageModels2022,rafailovDirectPreferenceOptimization2023}, and so is the catalogue of its limitations \cite{casperOpenProblemsFundamental2023}.
One of them decides AllotNet's shape.
When annotators' preferences depend on context the model does not observe, the utility recovered by Bradley--Terry preference learning, and hence by RLHF and DPO, equals a Borda-count aggregate of the per-context utilities \cite{siththaranjanDistributionalPreferenceLearning2023}.
Borda count can rank a dominated option, or one that harms a minority, highest, and it violates independence of irrelevant alternatives.
A single reward fitted to allocation judgements from many cultures would average those cultures away and could return a rule none of them holds.
AllotNet's answer is $K$ interpretable expert vectors and a gate over observed covariates, so that context is modelled rather than averaged away.

Remedies proposed elsewhere include distributional preference learning, from the same paper, mixture-of-preference reward models \cite{chakrabortyMaxMinRLHFAlignmentDiverse2024}, gated multi-objective reward heads, the closest architectural analogue \cite{wangInterpretablePreferencesMultiObjective2024}, and latent-context reward models \cite{poddarPersonalizingReinforcementLearning2024}.
The social-choice reading of the problem, that Bradley--Terry reward learning violates axioms one would want \cite{geAxiomsAIAlignment2024} and that social choice should guide alignment \cite{conitzerPositionSocialChoice2024}, joins the wider case for pluralistic alignment \cite{sorensenRoadmapPluralisticAlignment2024a,kirkPRISMAlignmentDataset2024}.
The closest precedents for carrying the judgement of many into a decision rule are the voting-based ethical decision systems built on Moral Machine data and on kidney-exchange value alignment \cite{noothigattuVotingBasedSystemEthical2018,freedmanAdaptingKidneyExchange2020,leeWeBuildAI2019}.
The instrument's design ancestor is the Moral Machine experiment itself, with 40 million decisions from 233 countries and territories \cite{awadMoralMachineExperiment2018}, and the lineage matters because most of moral psychology's findings come from Western, educated, industrialised, rich and democratic populations \cite{henrichWeirdestPeopleWorld2010}.

\subsection{Synthetic populations and planted ground truth}
\label{sec:planted}

Training on a synthetic population with a known answer key is a validation method with a lineage of its own.
Simulation-based calibration and parameter recovery test whether an estimator returns what was planted \cite{taltsValidatingBayesianInference2018,cookValidationSoftwareBayesian2006,modrakSimulationBasedCalibrationChecking2025,schadPrincipledBayesianWorkflow2021,wilsonTenSimpleRules2019,heathcoteIntroductionGoodPractices2015}; recovering the planted number of latent classes is a classic test in its own right \cite{nylundDecidingNumberClasses2007}; and planted structure can be provably unrecoverable below a detectability threshold \cite{decelleAsymptoticAnalysisStochastic2011,condonAlgorithmsGraphPartitioning2001}.
Noise ceilings set the bar a model can be held to: the explainable variance $R^2_{\text{ceiling}} = 1 - \mathrm{Var}(\text{noise})/\mathrm{Var}(\text{observed})$ \cite{hsuQuantifyingVariabilityNeural2004,schoppeMeasuringPerformanceNeural2016,lage-castellanosMethodsComputingMaximum2019,spearmanProofMeasurementAssociation1904}, for which unbiased estimators exist \cite{pospisilUnbiasedEstimationFraction2021}.
Writing the answer key first is blind analysis by another name \cite{maccounBlindAnalysisHide2015} and pre-registration by a third \cite{nosekPreregistrationRevolution2018}.
The synthetic population itself sits in the traditions of synthetic-data fidelity and utility \cite{jordonSyntheticDataWhat2022}, population synthesis \cite{beckmanCreatingSyntheticBaseline1996,farooqSimulationBasedPopulation2013} and validated agent-based models of humanitarian populations \cite{suleimenovaGeneralizedSimulationDevelopment2017}.

\subsection{Low-precision inference}
\label{sec:low-precision}

NVFP4 stores E2M1 elements in blocks of 16 with an E4M3 block scale and an FP32 tensor scale; its designers keep selected layers in higher precision, which is the precedent for keeping AllotNet's interpretable core in fp32 \cite{nvidiaPretrainingLargeLanguage2025,nvidiaHowQuantizationAware2025}.
The format belongs to the microscaling family \cite{rouhaniMicroscalingDataFormats2023,projectOCPMicroscalingFormats2023} and follows FP8 \cite{micikeviciusFP8FormatsDeep2022}.
Quantisation-aware training rests on the straight-through estimator \cite{bengioEstimatingPropagatingGradients2013,jacobQuantizationTrainingNeural2018,nagelWhitePaperNeural2021}, and NVFP4 has been run on low-power edge devices \cite{senCharacterizingImpactNVFP42026}.
Two findings temper the enthusiasm.
Quantisation can alter a model's explanations while preserving its accuracy \cite{rabbiSamePredictionsDifferent2026}, and compression harms the long tail of the data distribution first \cite{hookerWhatCompressedDeep2019}.
Cross-backend numerical validation \cite{phamCRADLECrossBackendValidation2019} is the precedent for the golden vectors shared by the C11, Java, Rust and Python runtimes, and the runtimes answer to the connectivity constraints of crisis settings \cite{gsmaConnectivityCrisisHumanitarian2024,hartungOpenDataKit2010}.

\section{The elicitation instrument}
\label{sec:allot}

Allot is the instrument whose sessions AllotNet is built to learn from.
It is a cross-cultural study of how people allocate scarce humanitarian aid when every choice has a cost, and its openly published record of judgements is meant to be the training signal for an allocation-support model that learns from what people judge to be right, not from cold optimisation.
This section describes the instrument only as far as the laboratory's public website already does; the specification itself is released with the first dataset (Section~\ref{sec:scope}).

\subsection{What a participant sees}
\label{sec:allot-session}

A session is anonymous and takes roughly forty minutes; the privacy notice (revision 3, 10 May 2026) still says ``A session takes about 20 minutes'', and on this point it lags behind the rest of the study's documentation.
A participant works through twenty-seven core dilemmas, each a concrete allocation choice with no costless answer.
The instrument runs in six languages (English, Russian, Spanish, French, Arabic and Portuguese) and draws on a bank of 75 scenarios and six meta-scenarios across 27 classes, with 56 beneficiary archetypes in five groups.
Every beneficiary figure has a name, one or two sentences of history and capabilities, ``not economics alone'', and every numeric attribute is traced to published field evidence.

Three design commitments carry the rest.
Sacrifice is mandatory: the budget is always smaller than the sum of the assistance packages, and there are no win-win options.
No right answer is ever asserted during play, and each scenario is written with arguments on more than one side, in descriptive language without evaluative adjectives, to keep the instrument from steering the participant.
And the participant answers from an assigned participant role.

Two features let the study see more than a first instinct.
The same dilemma is presented differently with the numbers held fixed, which separates the effect of presentation from the effect of the trade-off itself.
A mid-session change of conditions shows how people revise under new information.
For every decision the instrument records the option taken, the time taken to decide and the participant's stated confidence.

Each response falls into one of four categories, and these are the categories AllotNet's feature contract encodes (Section~\ref{sec:allotnet-features}): choosing a candidate (or a group of figures that form one alternative), splitting the budget between candidates where the scenario allows it, refusing all candidates where the scenario allows it, and skipping with a stated reason.
A skip is not a choice and is excluded from training.

Every alternative is decomposed, on the server, into signed contributions on seven criteria.
Participants are never shown that decomposition while playing, and no aggregate score is displayed during play.
Each participant sees their own decomposition in the debrief, once the last decision is submitted, and the debrief tells them that any decomposition of a choice into seven criteria is a normative choice rather than a computation of `the truth' \cite{arrowDifficultyConceptSocial1950}.

\subsection{Pre-registration, data governance and ethics}
\label{sec:allot-governance}

Specification v2.3 was frozen on 11 May 2026, with its SHA-256 digest published in the public repository before any session existed.
Version 2.4 of 16 August 2026, the canonical text, was deposited with the OSF registration.
The registration is embargoed until 1 August 2030.

Collected are an anonymous session identifier, the choices, time on screen, the browser language, a coarse region, a device class, a consent event and optional demographics, each with ``prefer not to say''.
Not collected are the IP address, name, email, phone, precise location, a device fingerprint or third-party cookies.
Identity questions (religion, ethnicity or citizenship) are optional and isolated from the narrative a participant sees; they are used only to measure bias on the server side, never to change what is shown.
Every record expires five years after creation, a ``Delete my session'' control erases a session immediately, and the age gate is 18 and over.
Code is released under Apache-2.0 and the dataset under CC-BY-4.0.

An ethics application for the study is a first submission, self-funded, unpaid and recruiting no vulnerable populations; the documents audited for this paper record the application but no decision on it, and the training of a deployed allocation model would need a separate ethics review.
The application declares a conflict of interest, the intent to train an allocation-support model, mitigated by pre-registration.
The protocol also fixes the model's limits in advance: it is meant to support allocation, never to compute a ``minimum necessary'' survival level, never to ration knowledge or medicine for institutional convenience, and never to be optimised to reduce anyone's decline.

The platform is a stateless single-page client in six locales behind a content-delivery layer, with a serverless back end and managed tables.
That back end matters to this paper for one reason: the synthetic population of Section~\ref{sec:synthetic} was generated by driving its production handler, not a re-implementation of it.

\section{AllotNet}
\label{sec:allotnet}

\subsection{Goals and design constraints}
\label{sec:allotnet-goals}

AllotNet solves the inverse problem to the instrument.
Given observed voting sessions, each a choice set of allocation alternatives, the participant's selection and the participant's context, it recovers the hidden structure of a population's moral preferences: the weights of the seven criteria, a segmentation of participants, regional gradients, group effects and in-group bias, in an interpretable form, ``so it can serve as an auditable reward substrate for a later allocation-support system rather than an opaque scorer''.
We wrote the model from scratch, with no pretrained weights and no fine-tuning; JAX is used only as a tensor library with automatic differentiation.
A pipeline was the alternative: fit a conditional logit, cluster its residuals, train a network on the clusters.
That is three models with three sets of assumptions and ``no single object you can audit''.
AllotNet is one forward pass, one loss and one training run.

Deliberately, the model is tiny.
Trained, it has 23,550 parameters, fitted on 93,032,246 training choices drawn from the 97,928,680 usable choices in a corpus of 103,937,259 responses.
Against the corpus total that is roughly one parameter to 4,400 observations; against the training split, roughly one to 3,950.
Its interpretable core is a single $8 \times 15 = 120$-number matrix; ``everything else routes to it or predicts outcomes''.
Overfitting ``is something the construction makes implausible'', and the validation negative log-likelihood was flat to four decimals between the last two epochs; the binding constraint was compute, not data.
Even the smallest planted effects rest on millions of observations.

Why not a scalar reward?
Because standard preference learning aggregates hidden context by Borda count \cite{siththaranjanDistributionalPreferenceLearning2023}, a scalar reward fitted across a heterogeneous population ``would silently distort the result and create an incentive to misreport''; an instrument that decomposes every choice into seven criteria and invokes Arrow's theorem points instead to a multi-objective reward with vector output and deferred aggregation \cite{rameRewardedSoupsParetooptimal2023,sorensenRoadmapPluralisticAlignment2024a}.
Two further constraints frame what follows: scale, since AllotNet ``is justified by the $10^8$ synthetic observations it was trained on, not by the study's actual sample size'' (Section~\ref{sec:discussion-scale}); and regulation, since under the EU AI Act human-in-the-loop operation, decision logging and directly readable weights are ``the compliance argument'', not a nice-to-have.

\subsection{Feature contract}
\label{sec:allotnet-features}

Nothing in the model is hard-wired to the data schema.
A versioned feature contract defines every input, and its version travels into every export.
There are $F = 15$ alternative features: seven criterion features, five group features, two virtual flags and an in-group affinity.
Criterion features are the scenario's server-side decomposition divided by a fixed scale, with the set of criteria discovered from the content.
Group features are a one-hot of the beneficiary archetype's group, again discovered from the content.
The virtual flags mark the refusal and split alternatives, which are added to the choice set whenever the scenario allows them and carry fixed implicit decompositions of their own.
In-group affinity is one minus an identity distance computed from the optional identity items; a player without an identity profile scores 0.

What the gate sees is the participant, not the alternative.
Its context is eight categoricals (the assigned role, language, country, operating system, device type and three onboarding answers) and seven numerics: whether an identity profile was submitted, the mean and standard deviation of the log decision time, the share of skipped choices, the mean number of reconsideration events, the mean confidence (filled with 50.0 when absent) and the share of very fast responses.
Numerics are $z$-scored with $\sigma + 10^{-6}$; missing or out-of-vocabulary categories map to a last index.
The eight embedding tables hold 968 parameters, and the gate input has dimension $8 \times 8 + 7 = 71$.
Drift is tolerated: a missing column is filled neutrally with a warning, an unknown category goes to the out-of-vocabulary index, and extra columns are ignored.
Skips are excluded (``SKIP responses are not choices''), as are responses whose scenario or (scenario, choice) pair is not in the content lookup, and the choice set is capped at eight alternatives.
Of 103,937,259 responses, 97,928,680 are usable.

\subsection{Forward pass}
\label{sec:allotnet-forward}

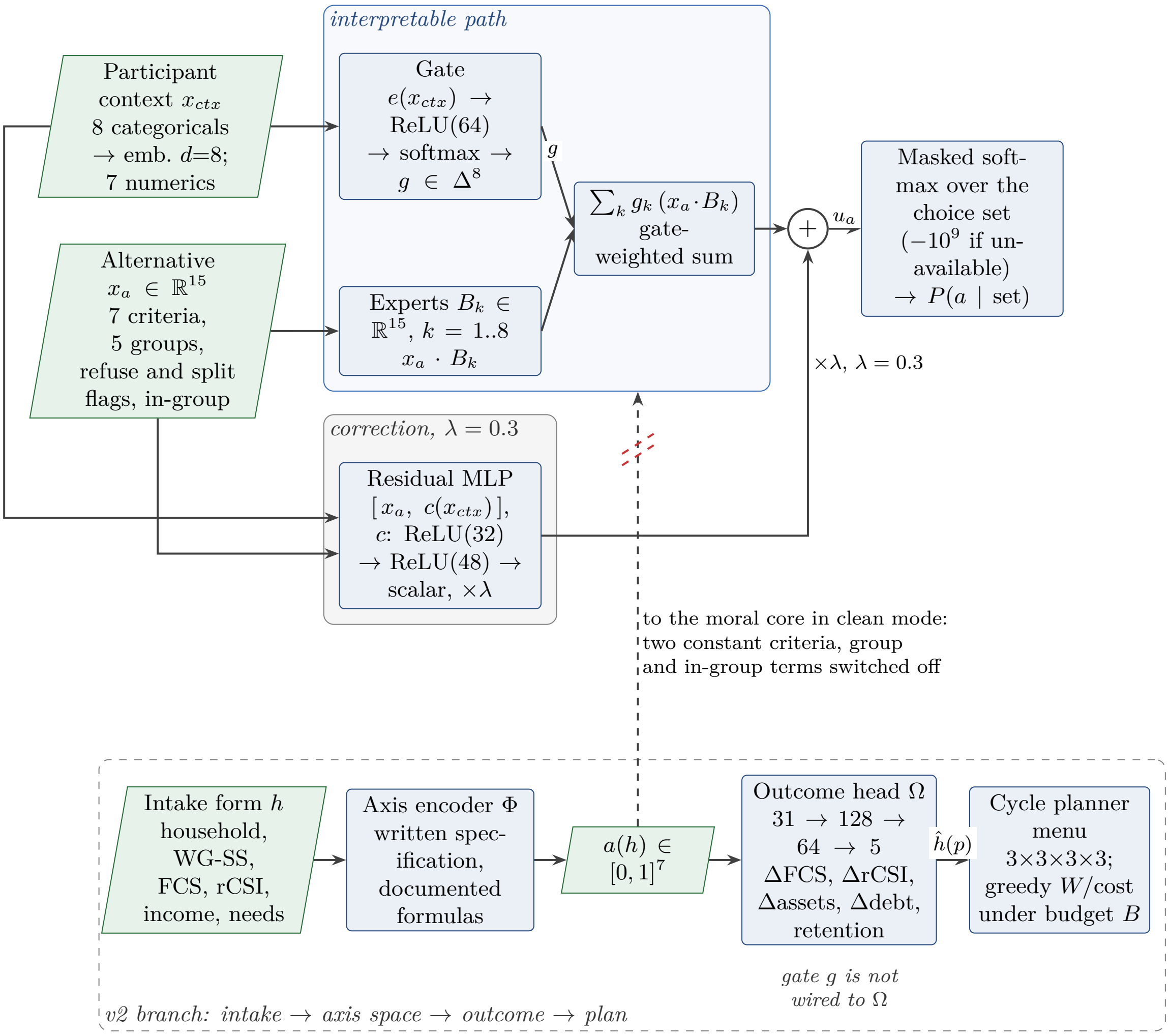
\begin{figure}[htbp]
  \centering
  \begin{allotfit}
  \begin{tikzpicture}[node distance=1.2cm and 0.45cm]
    \node[data, text width=2.5cm] (xctx) {Participant context $x_{ctx}$\\8 categoricals $\to$ emb.\ $d{=}8$;\\7 numerics};
    \node[block, text width=2.6cm, right=0.95cm of xctx] (gate) {Gate\\$e(x_{ctx}) \to \mathrm{ReLU}(64)$\\$\to \mathrm{softmax} \to g \in \Delta^{8}$};
    \node[block, text width=2.6cm, below=of gate] (experts) {Experts $B_k \in \mathbb{R}^{15}$, $k = 1..8$\\$x_a \cdot B_k$};
    \node[data, text width=2.5cm, left=0.95cm of experts] (xa) {Alternative $x_a \in \mathbb{R}^{15}$\\7 criteria, 5 groups,\\refuse and split flags, in-group};
    \node[block, text width=2.3cm, anchor=west] (mix) at ($(gate.east)!0.5!(experts.east)+(0.45,0)$) {$\sum_k g_k\,(x_a \!\cdot\! B_k)$\\gate-weighted sum};
    \node[circle, draw=black!75, thick, minimum size=0.55cm, inner sep=0pt, font=\small, right=of mix] (sum) {$+$};
    \node[block, text width=2.6cm, right=of sum] (soft) {Masked softmax over the choice set\\($-10^{9}$ if unavailable)\\$\to P(a \mid \text{set})$};
    \node[block, text width=2.6cm, below=of experts] (res) {Residual MLP\\$[\,x_a,\ c(x_{ctx})\,]$, $c$: ReLU(32)\\$\to \mathrm{ReLU}(48) \to$ scalar, $\times\lambda$};
    \node[block, text width=2.4cm, below=2.5cm of res] (phi) {Axis encoder $\Phi$\\written specification,\\documented formulas};
    \node[data, text width=2.2cm, left=of phi] (intake) {Intake form $h$\\household, WG-SS, FCS, rCSI, income, needs};
    \node[data, text width=1.6cm, right=of phi] (ah) {$a(h) \in [0,1]^{7}$};
    \node[block, text width=2.5cm, right=of ah] (omega) {Outcome head $\Omega$\\$31 \to 128 \to 64 \to 5$\\$\Delta$FCS, $\Delta$rCSI, $\Delta$assets, $\Delta$debt, retention};
    \node[block, text width=2.3cm, right=of omega] (planner) {Cycle planner\\menu $3{\times}3{\times}3{\times}3$;\\greedy $W/\mathrm{cost}$ under budget $B$};
    \node[note, text width=2.7cm, below=0.2cm of omega, align=center] (gnote) {gate $g$ is not wired to $\Omega$};
    \begin{scope}[on background layer]
      \node[bluegroup, fit={(gate)(experts)(mix)($(gate.north)+(0,0.45)$)},
        label={[grouplabel, text=mainblue!70!black, anchor=north west]north west:interpretable path}] (interp) {};
      \node[graygroup, fit={(res)($(res.north)+(0,0.45)$)},
        label={[grouplabel, anchor=north west]north west:correction, $\lambda = 0.3$}] (corr) {};
      \node[group, fit={(intake)(phi)(ah)(omega)(planner)(gnote)($(intake.south)+(0,-0.45)$)},
        label={[grouplabel, anchor=south west]south west:v2 branch: intake $\to$ axis space $\to$ outcome $\to$ plan}] (v2) {};
    \end{scope}
    \draw[arrow] (xctx) -- (gate);
    \draw[arrow] (xa) -- (experts);
    \draw[arrow] (gate.east) -- node[edgelabel, above, pos=0.35] {$g$} (mix.west);
    \draw[arrow] (experts.east) -- (mix.west);
    \draw[arrow] (mix) -- (sum);
    \draw[arrow] (sum) -- node[edgelabel, above] {$u_a$} (soft);
    \coordinate (trunk) at ($(xa.bottom left corner)+(-0.35,0)$);
    \draw[arrow] (xctx.west) -- (xctx.west -| trunk) |- ([yshift=0.25cm]res.west);
    \draw[arrow] (xa.south) |- ([yshift=-0.25cm]res.west);
    \draw[arrow] (res.east) -| node[edgelabel, pos=0.8, right] {$\times\lambda$, $\lambda = 0.3$} (sum.south);
    \draw[arrow] (intake) -- (phi);
    \draw[arrow] (phi) -- (ah);
    \draw[arrow] (ah) -- (omega);
    \draw[arrow] (omega) -- node[edgelabel, above] {$\hat h(p)$} (planner);
    \draw[dashedarrow] (ah.north) -- node[edgelabel, right, pos=0.42, align=left] {to the moral core in clean mode:\\two constant criteria, group\\and in-group terms switched off} (ah.north |- interp.south);
    \draw[cut] ($(ah.north)!0.88!(ah.north |- interp.south)+(-0.22,-0.14)$) -- ++(0.44,0.28);
    \draw[cut] ($(ah.north)!0.88!(ah.north |- interp.south)+(-0.22,-0.30)$) -- ++(0.44,0.28);
  \end{tikzpicture}
  \end{allotfit}
  \caption{AllotNet architecture.
    v1 (upper part): the participant context is embedded and gated into a distribution over eight segments; each alternative's fifteen features are scored by eight linear experts $B_k$ and mixed by the gate (interpretable path, blue), and a small residual MLP is added at weight $\lambda = 0.3$ (correction, grey).
    Utilities pass through a masked softmax over the alternatives actually present in the choice set.
    v2 (lower part): a specification-driven encoder $\Phi$ maps an intake questionnaire into the same seven-criterion space $a(h)$, which feeds the moral core in clean mode (two criteria held constant on intake, group and in-group terms switched off; dashed cut) and the outcome head $\Omega$; the cycle planner searches the $3{\times}3{\times}3{\times}3$ plan menu greedily by value per cost under budget $B$.
    The gate is deliberately not wired to $\Omega$.}
  \label{fig:allotnet-arch}
\end{figure}

Figure~\ref{fig:allotnet-arch} shows the architecture.
Its moral core is the v1 architecture, carried into v2 verbatim.
Let $c_1, \ldots, c_8$ be the categorical codes and $z \in \mathbb{R}^7$ the standardised numerics of a session.
The gate input concatenates the eight embeddings with the numerics,
\begin{equation}
  \label{eq:gate-input}
  e(x_{\mathrm{ctx}}) = \big[\, \mathrm{emb}_1[c_1];\; \ldots;\; \mathrm{emb}_8[c_8];\; z \,\big] \in \mathbb{R}^{71},
\end{equation}
and the gate is a one-hidden-layer network with a softmax over $K$ segments,
\begin{equation}
  \label{eq:gate}
  g(x_{\mathrm{ctx}}) = \operatorname{softmax}\!\big( W_2\, \mathrm{ReLU}(W_1 e + b_1) + b_2 \big) \in \Delta^{K}, \qquad W_1 \in \mathbb{R}^{71 \times 64},\; W_2 \in \mathbb{R}^{64 \times 8}.
\end{equation}
For alternative $a$ with feature vector $x_a \in \mathbb{R}^{15}$, utility is a gate-weighted mixture of $K$ linear experts $B_k \in \mathbb{R}^{15}$ plus a small residual,
\begin{equation}
  \label{eq:utility}
  u_a = \sum_{k=1}^{K} g_k\, (x_a \cdot B_k) \;+\; \lambda\, \mathrm{MLP}\big(x_a, c(x_{\mathrm{ctx}})\big),
  \qquad c(x_{\mathrm{ctx}}) = \mathrm{ReLU}(W_{\mathrm{ctx}}\, e + b_{\mathrm{ctx}}) \in \mathbb{R}^{32},
\end{equation}
where $\mathrm{MLP}(x_a, c) = w_2^{\top}\, \mathrm{ReLU}\big(W_1^{\mathrm{res}} [x_a; c] + b_1^{\mathrm{res}}\big) + b_2^{\mathrm{res}}$ with $W_1^{\mathrm{res}} \in \mathbb{R}^{47 \times 48}$ and $w_2 \in \mathbb{R}^{48 \times 1}$.
Choice probability is a softmax over the available set,
\begin{equation}
  \label{eq:choice}
  P(\text{choice} = a \mid \text{set}) = \frac{\exp(u_a)}{\sum_{a' \in \text{set}} \exp(u_{a'})},
\end{equation}
with unavailable alternatives masked to $-10^{9}$ before the softmax, which gives them ``zero probability without special-casing the normaliser''.
One scorer is applied to every alternative in the set, so the model is permutation-equivariant in the manner of Deep Sets \cite{zaheerDeepSets2017}.

Hyperparameters are $K = 8$, $F = 15$, embedding width 8, gate hidden width 64, context width 32, residual hidden width 48 and $\lambda = 0.3$.
Initialisation is Gaussian: embeddings with standard deviation 0.05, the gate, context and first residual weights 0.08, the experts 0.02 and the second residual weights 0.05.
In the v1 trainer the residual is switched off for the first 40 per cent of steps, implemented as $\max(1, \lfloor 0.4 \times \text{epochs} \rfloor)$ warm-up epochs, so with six epochs the first two ran at $\lambda = 0$; the linear experts capture the main signal first ``so interpretability does not `leak' into the MLP''.
Since v2 warm-starts from v1, it has no warm-up phase.

A conditional logit \cite{mcfaddenConditionalLogitAnalysis1974} is the special case $K = 1$, $\lambda = 0$,
\begin{equation}
  \label{eq:clogit}
  P(\text{choice} = a \mid \text{set}) = \frac{\exp(x_a \cdot \beta)}{\sum_{a' \in \text{set}} \exp(x_{a'} \cdot \beta)},
\end{equation}
and we keep it as a diagnostic probe, fitted globally or by subgroup with Adam at learning rate 0.3 for at most 400 steps, an $\ell_2$ penalty of $10^{-6}$ and early stopping when the loss changes by less than $10^{-7}$.
Each fit is capped at 20,000,000 observations (a seeded random subsample), because the logit materialises the full observations-by-alternatives-by-features tensor and the whole corpus would need about 50~GB; groups below 3,000 observations are skipped.
One of the checks (Section~\ref{sec:evaluation-c}) concerns an estimator rather than a property of the gate and is computed from these probes.

\subsection{Loss and building blocks}
\label{sec:allotnet-loss}

Four terms make up the objective,
\begin{equation}
  \label{eq:loss}
  \mathcal{L} = -\frac{1}{N} \sum_{i=1}^{N} \log P(y_i) \;+\; \ell_2^{B}\, \lVert B \rVert^2 \;+\; \ell_2^{r} \big( \lVert W_1^{\mathrm{res}} \rVert^2 + \lVert W_2^{\mathrm{res}} \rVert^2 \big) \;+\; \gamma \sum_{k=1}^{K} \bar{g}_k \log \bar{g}_k,
\end{equation}
with $\ell_2^{B} = 3 \times 10^{-5}$ (``weak, the experts are meant to carry signal''), $\ell_2^{r} = 10^{-3}$ (``strong, the residual is meant to stay small''), $\gamma = 0.01$ and $\bar{g}_k$ the batch-mean gate mass on segment $k$.
That last term is negative entropy and is added, so minimising pushes the batch-mean gate distribution towards higher entropy; it is a penalty against gate collapse, in which every participant is routed to one segment and $K - 1$ experts stay untrained.
AdamW optimises the choice stream at learning rate $3 \times 10^{-3}$ and weight decay $10^{-6}$, batch 16,384 on the GPU, with a validation split of the first $\max(2000, 0.05 N)$ rows of a seeded permutation.
When the set has two alternatives the choice likelihood is the Bradley--Terry model \cite{bradleyRankAnalysisIncomplete1952}, ``the standard for preference-learned rewards'' \cite{christianoDeepReinforcementLearning2017}.
Table~\ref{tab:building-blocks} names the lineage of each component.

\begin{table}[htbp]
  \setstretch{1}
  \centering
  \footnotesize
  \caption{Building blocks of AllotNet and their basis in the discrete-choice and preference-learning literature.}
  \label{tab:building-blocks}
  \begin{tabularx}{\textwidth}{@{}l X X@{}}
    \toprule
    Block & What it does & Basis \\
    \midrule
    Softmax over the choice set & Probability of choosing among available alternatives & Random utility maximisation \cite{mcfaddenConditionalLogitAnalysis1974} \\
    $K$ linear taste experts $B_k$ & Interpretable feature-weight vectors (criteria, group and in-group terms) & Latent-class discrete choice \cite{kamakuraProbabilisticChoiceModel1989,trainDiscreteChoiceMethods2003} \\
    Gate over participant context & Segment membership as a function of observed covariates & Concomitant-variable latent class \cite{daytonConcomitantVariableLatentClassModels1988}; mixture of experts \cite{jacobsAdaptiveMixturesLocal1991} \\
    Small residual MLP ($\lambda = 0.3$, strong $\ell_2$) & Non-linear corrections that do not steal signal from the linear part & L-MNL, linear kernel plus representation \cite{sifringerEnhancingDiscreteChoice2020} \\
    One scorer per alternative & Permutation equivariance over the set & Deep Sets \cite{zaheerDeepSets2017} \\
    Choice NLL as objective & Bradley--Terry when $|\text{set}| = 2$; preference-learned rewards & \cite{christianoDeepReinforcementLearning2017} \\
    \bottomrule
  \end{tabularx}
\end{table}

\subsection{Encoder for household intake}
\label{sec:allotnet-encoder}

Scenario alternatives already carry a decomposition on the seven criteria; a real household does not.
An encoder $\Phi$ therefore maps an intake questionnaire $h$ into $a(h) \in [0,1]^7$, the same seven-dimensional space in which the scenarios' own decompositions live.
This vector is the interlingua of the system: the moral core and the outcome head connect only through $a(h)$ and a structural vector $s(h)$.
The encoder is a written, versioned specification rather than learned weights, so it can be read and argued with.
Each criterion is a weighted list of terms, each term a field, an operator, parameters and a weight; the operators are min--max scaling (clipped), its complement, an indicator, a threshold, a constant, the identity and a within-population percentile.
Each criterion's value is the clipped weighted mean
\begin{equation}
  \label{eq:axis-term}
  a_j(h) = \operatorname{clip}_{[0,1]}\!\left( \frac{\sum_{t \in \mathcal{T}_j} w_t\, t(h)}{\sum_{t \in \mathcal{T}_j} w_t} \right).
\end{equation}
The terms read the standard sector fields: the food-consumption deficit, the coping index, the household's gap against its basket, shelter and WASH conditions, coverage by other programmes, a deprivation index, displacement status, unaccompanied children, protection concerns, Washington Group disability, market efficiency and remoteness.
Two of the seven criteria are constants on household intake, so they carry no information about a household; the per-criterion assignment of the other terms is part of the withheld specification.

Derived fields are economic.
With $\mathrm{MEB}_{\mathrm{hh}}$ the household's minimum expenditure basket per month, income $y$ and other assistance $o$,
\begin{equation}
  \label{eq:gap-ratio}
  \mathrm{gap}_{\mathrm{month}} = \max\big(0,\, \mathrm{MEB}_{\mathrm{hh}} - y - o\big), \qquad
  \mathrm{gap\_ratio} = \operatorname{clip}_{[0,1]}\!\left( \frac{\mathrm{gap}_{\mathrm{month}}}{\mathrm{MEB}_{\mathrm{hh}}} \right).
\end{equation}
Food-security fields are normalised as
\begin{equation}
  \label{eq:food-norms}
  \mathrm{fcs\_deficit} = \operatorname{clip}_{[0,1]}\!\big(1 - \tfrac{\mathrm{FCS}}{60}\big), \qquad
  \mathrm{rcsi\_norm} = \operatorname{clip}_{[0,1]}\!\big(\tfrac{\mathrm{rCSI}}{40}\big),
\end{equation}
and market efficiency is 1.0, 0.85 or 0.6 for good, limited and poor markets.
One of the encoder's terms is the within-population percentile of a deprivation index,
\begin{equation}
  \label{eq:depr-index}
  \mathrm{depr\_index} = 0.4\, \mathrm{gap\_ratio} + 0.3\, \mathrm{fcs\_deficit} + 0.2\, \mathrm{rcsi\_norm} + 0.1\, \operatorname{clip}_{[0,1]}(\mathrm{shelter\_cond}/3).
\end{equation}

Household baskets use an equivalence scale \cite{networkTransferValuesHow2021,associationSphereHandbookHumanitarian2018},
\begin{equation}
  \label{eq:equivalence-scale}
  n_{\mathrm{eq}} = \max\big(1,\, \text{adults} + 0.6\, \text{children} + 0.5\, \text{children}_{u5} + 0.9\, \text{elderly}\big)^{\theta}, \qquad \theta = 0.85,
\end{equation}
\begin{equation}
  \label{eq:meb-hh}
  \mathrm{MEB}_{\mathrm{hh}} = 38.0\, n_{\mathrm{eq}} + 6.0\, n_{\mathrm{disability}} + 5.0\, n_{\mathrm{PLW}} + 3.0\, n_{\mathrm{elderly}} + 4.0\, [\text{female-headed}],
\end{equation}
with a survival basket at 0.70 of the full basket.
Delivery overheads are 0.08 for cash, 0.15 for vouchers and 0.35 for in-kind food; administrative cost is 15.0 per household plus 2.0 per payment cycle (monthly pays every month, quarterly $\lceil \text{duration}/3 \rceil$ times, lump sum once); packages last at least three months, transfer at least 5.0 per month, in-kind food covers the food gap only, and a lump sum pays the total up front.
Admissible plans are the product of coverage of gap $\{0.5, 0.8, 1.0\}$, frequency $\{\text{monthly}, \text{quarterly}, \text{lump sum}\}$, duration $\{3, 6, 12\}$ months and modality $\{\text{cash}, \text{voucher}, \text{in-kind food}\}$, $3 \times 3 \times 3 \times 3 = 81$ combinations per household, minus rows whose monthly value falls below 5.0.
These rules are configuration, not code, so a programme can encode its own working group's decisions rather than inherit ours.

The structural vector $s(h)$ has 14 entries: log household size, equivalence size, gap ratio, normalised FCS, rCSI and debt, an asset score, three market indicators, urban, the vulnerability rate, the dependency ratio and female headship; missing names are filled with zero.
A percentile needs a population to be taken against, which a portable runtime does not have, so the bundle ships a 2,001-point quantile grid of the training population's deprivation index against which the runtime interpolates, ``deterministic and independent of input size''.
Intake follows a minimal household form covering composition, location (OCHA P-codes) \cite{ochaCommonOperationalDatasets2024}, protection (including the Washington Group disability count) \cite{slomanWashingtonGroupShort2018}, economy, food security (FCS, rCSI, HDDS, coping stage) \cite{wfpConsolidatedApproachReporting2021}, shelter and WASH, consent, and an extension prefix for programme-specific fields.

\subsection{Outcome head}
\label{sec:allotnet-outcome-head}

An outcome head $\Omega$ predicts how a household responds to a plan $p$,
\begin{equation}
  \label{eq:outcome-head}
  \Omega:\; \big[a(h), s(h), p\big] \in \mathbb{R}^{31} \;\longmapsto\; y = \big(\Delta\mathrm{FCS}, \Delta\mathrm{rCSI}, \Delta\mathrm{Assets}, \Delta\mathrm{Debt}, \text{retention}\big) \in \mathbb{R}^{5},
\end{equation}
as a ReLU network $31 \to 128 \to 64 \to 5$ initialised with standard deviation $1.4/\sqrt{d_{\mathrm{in}}}$.
Its input concatenates the seven criteria, the 14 structural features, a modality one-hot over cash, voucher, in-kind food and none, the coverage, a frequency one-hot over monthly, quarterly and lump sum, the duration divided by 12, and the monthly value relative to the basket; in all $7 + 14 + 4 + 1 + 3 + 1 + 1 = 31$.
Targets are divided by $(20.0, 10.0, 0.2, 1.0, 1.0)$ before a mean-squared-error loss with an $\ell_2$ penalty of $10^{-6}$, optimised with Adam at learning rate $2 \times 10^{-3}$.
Deliberately, the gate is not wired to $\Omega$: ``whether assistance improves a household's food consumption cannot depend on who is looking at the household''.
Hence the two blocks have disjoint parameters and two optimisers over one bundle.
$\Omega$ has $31 \times 128 + 128 + 128 \times 64 + 64 + 64 \times 5 + 5 = 12{,}677$ parameters.

\subsection{Cycle planner}
\label{sec:allotnet-planner}

A planner allocates the budget over households.
For household $i$ the plan set $\mathcal{P}(h_i)$ is the 81-row menu above after the constraints; the objective is
\begin{equation}
  \label{eq:planner-objective}
  \max_{p_i \in \mathcal{P}(h_i)} \sum_i W_i(p_i) \quad \text{s.t.} \quad \sum_i \mathrm{cost}(p_i) \leq B,
\end{equation}
with
\begin{equation}
  \label{eq:plan-cost}
  \mathrm{cost}(p) = \text{value}_{\mathrm{month}} \cdot \text{duration} \cdot (1 + \text{overhead}_{\mathrm{modality}}) + \text{admin}(\text{freq}, \text{duration}).
\end{equation}
Welfare of a plan is the need it closes, weighted by the moral core and discounted by persistence,
\begin{equation}
  \label{eq:welfare}
  W_i(p) = \big(0.5 + 0.5\, \text{retention}\big) \sum_{j=1}^{7} \beta^{+}_{j}\, \operatorname{clip}_{\geq 0}\!\Big( a_{j}(h_i) - a_{j}\big(\hat{h}_i(p)\big) \Big),
\end{equation}
where $\hat{h}_i(p)$ is the post-assistance state predicted by $\Omega$ (post columns clipped to their valid ranges, and post criteria re-encoded with percentiles taken against the pre-assistance population) and $\beta^{+}$ are the clipped-positive clean-mode criterion weights of the moral core.
In the architecture document's words, ``the value of a plan is how much of the household's need it closes, weighted by how much humanity cares about that kind of need, discounted by whether the gain persists''.
The weights are the recovered clean-mode weights of a core trained on the planted corpus, so `what humanity cares about' is the document's phrase and not a measurement.

Solving is lazy greedy on marginal value per dollar: menu rows are sorted within household by descending $W/\mathrm{cost}$, a heap holds each household's best row, the top is popped and taken if affordable or replaced by that household's next row, and rows with $W \leq 0$ are skipped.
A naive baseline gives every household with a positive gap the same monthly cash package for three months with coverage capped at 1.2.
We do not claim the $(1 - 1/e)$ guarantee \cite{nemhauserAnalysisApproximationsMaximizing1978}: submodularity holds within a household through the saturating dose-response but ``does not hold cleanly across the fixed enrolment cost'', which makes the problem a knapsack with set-up costs, so the greedy solution is reported against an oracle (check D6) rather than a bound.
An LP-relaxation mode ``is available for comparison'', according to the architecture document; no such code path exists in the staged planner.
Every decision carries its top drivers among the criteria, and the planner writes household, modality, monthly value, frequency, duration, total cost, welfare, predicted $\Delta$FCS, predicted retention and monthly gap per row.

Two properties of this objective belong here, where it is defined, rather than in a caveat.
Because the heap orders menu rows, the object being ranked is the pair of a household and a plan, never the household: the moral score orders need, the queue orders purchasable closure of need per dollar, and Section~\ref{sec:results-queue-score} shows how far the two orderings come apart.
And the objective is per household while the cost is not.
The core's feature vector carries no size or dependants term, and all seven criteria are intensive (rates, percentiles, indicators), so the criterion space has no extensive dimension; $W$ multiplies by no headcount, so closing the same relative gap for eight people scores exactly what it scores for two; and the cost side is scaled by the basket's adult-equivalent size ($\theta = 0.85$).
Per capita, a large household is therefore penalised by roughly $\text{size}^{0.85}$, which is the mechanism behind the basket split of Section~\ref{sec:results-queue-score}.

\subsection{Clean mode and governance flags}
\label{sec:allotnet-clean}

Findings are read off the weights, with no post-hoc attribution.
Rows $B_k$ are segment profiles; contrasts between group coefficients measure group effects; the gate posteriors give segment membership (purity, adjusted Rand index); $\mathbb{E}[g \cdot B \mid \text{country}]$ gives effective per-country weights and hence cohort gradients; and $\pi_k = \mathbb{E}[g_k]$ gives response-weighted segment sizes.
Clean mode zeroes the group and in-group coefficients, leaving the seven criteria and the structural features.
Ordering matters here: the model must first demonstrate that it can see the bias, which checks C4 and C5 test, and only then is the switch to exclude it meaningful.
``A model that never represented [the bias] could not prove it had removed it.''
Clean mode is the recorded default for operational decisions and the full model is the research instrument.
Two of the criteria are off by default on real intake, ``a programme ethics decision, not an engineering one''; since the encoder emits them as constants, their coefficients multiply a constant, and because the planner uses $\beta^{+} \cdot \Delta a$ with $\Delta a = 0$ on a constant criterion they contribute nothing to $W$.
On household intake the switch itself has nothing to zero: the encoder emits no group or in-group feature for a household, so the group and in-group coefficients have no input to multiply and the protection on that path is structural rather than selected.
The switch bites on the instrument's own scenarios, where archetype group membership and in-group status are real features.
Bundle metadata carries the flags: default decision mode clean, human in the loop, and the note that the group and in-group coefficients ``take no part in operational decisions''.
Every runtime exposes the same three levels: a criterion profile (``pure needs assessment, no judgment''), clean-mode target-group selection with a per-decision rationale, and cycle planning.

\subsection{NVFP4 quantisation-aware training}
\label{sec:allotnet-nvfp4}

A 23,550-parameter model does not need compressing.
We quantised it for two other reasons.
Blackwell-class accelerators, including the DGX Spark GB10 used for training, execute matrix products natively in NVFP4, so training in that format means ``the numbers the chip computes are the numbers the evaluation was run on'' \cite{nvidiaPretrainingLargeLanguage2025}.
And the $8 \times 15$ audit matrix must not be perturbed by 4-bit execution, which gives the policy ``quantise the machinery, never the conclusion''.

NVFP4 stores each element as E2M1 (one sign, two exponent and one mantissa bit), whose representable magnitudes are $\{0, 0.5, 1, 1.5, 2, 3, 4, 6\}$, two per byte, under two levels of scaling: an FP8 E4M3 scale (maximum 448) per block of 16 elements and one FP32 scale per tensor \cite{projectOCPMicroscalingFormats2023}.
For a tensor $x$ with block $b$,
\begin{equation}
  \label{eq:nvfp4}
  \hat{x} = \text{code} \times s_{\mathrm{block}} \times s_{\mathrm{tensor}}, \qquad
  s_{\mathrm{tensor}} = \frac{\max |x|}{6 \times 448}, \qquad
  s_{\mathrm{block}} = \mathrm{E4M3}\!\left( \frac{\max_{b} |x|}{6\, s_{\mathrm{tensor}}} \right).
\end{equation}
MXFP4, by contrast, uses blocks of 32 with a power-of-two E8M0 scale.
Training uses fake quantisation with a straight-through estimator: the forward pass sees the quantised weights and the activations feeding each GEMM (W4A4 during training), rounding to nearest on the E2M1 grid (a stochastic-rounding variant exists in the reference implementation but is not used by the training path), and the backward pass uses the identity gradient.
Weights of a GEMM $x @ w$ are blocked along the contraction axis, and activations along their last axis, because that is the axis the chip blocks along.
Table~\ref{tab:precision} states what is quantised.

\begin{table}[htbp]
  \setstretch{1}
  \centering
  \footnotesize
  \caption{Precision of each component under NVFP4 quantisation-aware training.}
  \label{tab:precision}
  \begin{tabularx}{\textwidth}{@{}l l X@{}}
    \toprule
    Component & Precision & Reason \\
    \midrule
    Gate, both layers & NVFP4 & routing machinery \\
    Context projection & NVFP4 & machinery \\
    Residual MLP, both layers & NVFP4 & a small correction term to begin with \\
    Outcome head, first two layers & NVFP4 & the largest GEMMs in the model \\
    Expert matrix ($8 \times 15$) & full precision & the interpretable output a human audits \\
    Categorical embeddings & full precision & lookup tables, not GEMM inputs \\
    Final logits and softmax & full precision & last layer before a probability \\
    Final 5-output layer of $\Omega$ & fp32 in training; packed in export & tiny; sets the outcome scale \\
    \bottomrule
  \end{tabularx}
\end{table}

One tensor needs a note.
The final layer of $\Omega$ is in the list of quantisation-aware tensors and is therefore packed into the NVFP4 parameter file, but the outcome forward pass applies the quantiser only to the first two layers and multiplies by the third unquantised.
The documentation lists it once as NVFP4 and once, as ``final 5-output GEMM'', as full precision; both rows refer to the same tensor, and its training-time treatment is full precision.
Hidden widths were chosen as multiples of 16 (gate 64, context 32, residual 48, $\Omega$ 128 and 64); the contraction axes that cannot fit, the $\Omega$ input of 31 and the 15 alternative features, are zero-padded inside the quantiser and stripped on the way out, as the runtime does.
A self-test asserts that packing then unpacking equals the fake-quantised tensor to $10^{-6}$.
Quantisation-aware training repeats the v2 schedule, warm-started from the v2 parameters, with two choice epochs and twelve outcome epochs at batches 16,384 and 32,768 and the quantisation flag set.

\subsection{Bundle and runtimes}
\label{sec:allotnet-bundle}

Table~\ref{tab:param-budget} gives the parameter budget and the three sub-totals that answer different questions: the full trained model (25 arrays), the inference bundle (19 tensors), and the interpretable core.
A fourth figure, 14,501 packed words, is the size of the quantisation-aware GEMMs alone in the NVFP4 parameter file.

\begin{table}[htbp]
  \setstretch{1}
  \centering
  \footnotesize
  \caption{Parameter budget of AllotNet v2. The inference bundle omits the context projection and the residual MLP, so the portable runtimes evaluate the linear mixture only.}
  \label{tab:param-budget}
  \begin{tabular}{@{}l l r@{}}
    \toprule
    Block & Shape & Parameters \\
    \midrule
    Categorical embeddings & eight tables of width 8 & 968 \\
    Gate, layer 1 & $71 \times 64$, $64$ & 4{,}608 \\
    Gate, layer 2 & $64 \times 8$, $8$ & 520 \\
    Expert matrix & $8 \times 15$ & 120 \\
    Context projection & $71 \times 32$, $32$ & 2{,}304 \\
    Residual, layer 1 & $47 \times 48$, $48$ & 2{,}304 \\
    Residual, layer 2 & $48 \times 1$, $1$ & 49 \\
    Outcome head, layer 1 & $31 \times 128$, $128$ & 4{,}096 \\
    Outcome head, layer 2 & $128 \times 64$, $64$ & 8{,}256 \\
    Outcome head, layer 3 & $64 \times 5$, $5$ & 325 \\
    \midrule
    Interpretable core & & 120 \\
    v1 moral core (all but $\Omega$) & & 10{,}873 \\
    Inference bundle (19 tensors) & & 18{,}893 \\
    Full trained model (25 arrays) & & 23{,}550 \\
    \bottomrule
  \end{tabular}
\end{table}

An exporter writes three files: a human-readable JSON twin with both specifications and the weights as nested lists, a binary bundle, and a golden-vector file of eight reference cases at tolerance $10^{-4}$.
The metadata includes both specifications, the global weights, the experts ordered by $\pi$, the gate vocabularies, the percentile reference grid and the governance flags.
The binary layout (Appendix~\ref{app:anb}) is a four-byte magic, a 32-bit little-endian metadata length, the metadata as UTF-8 JSON without tensors, a 32-bit tensor count, and per tensor a 16-bit name length, the name, an 8-bit rank, 32-bit dimensions and the data as little-endian float32 in C order.
The 19 tensors are the outcome head, the gate, the eight embeddings and the expert matrix, 18,893 floats in all.
Not carried are the context projection and the residual MLP, so the L-MNL residual cannot be evaluated by the portable runtimes.
The fp32 v2 bundle is 124,690 bytes (metadata 48,814 bytes, 19 tensors) and its fp4 twin 124,688 bytes; the golden-vector files are 18,516 and 18,521 bytes.
Header arithmetic closes (Appendix~\ref{app:anb}).
Earlier records give sizes of 105,651 and 105,649 bytes, and the website speaks of a file of ``about 600 KB''; the bundles were since re-exported with English specification text, tensors unchanged, and we cite the measured sizes.

Each golden case runs an intake through the derived economics, the seven criteria, the clean score, three candidate plans (cash at 0.8 monthly for six months; in-kind food at 0.5 monthly for three months; a voucher at 1.0 as a lump sum for twelve months), their outcomes and their total cost.
Tolerances are $10^{-4}$ on economics, criteria, clean score and plan costs, $10^{-3}$ on predicted outcomes, plus planner budget discipline.
``A runtime that reproduces all 8 cases has implemented the model, not merely the matrix multiplications.''
Four interchangeable runtimes read the same bundle: a Python reference on numpy, a dependency-free Java library for JDK 17 and later, a C11 library needing only libc and libm, and a Rust crate on the standard library alone.
Each exposes the three levels of Section~\ref{sec:allotnet-clean}, with single-plan prediction and menu enumeration beside them.
At the first snapshot only the C11 runtime had been re-verified against the current bundles; a second phase measured all four on every benchmark machine (Sections~\ref{sec:evaluation-bench} and~\ref{sec:results-runtime}).
The bundle's metadata carry the criteria and the expert vectors, so the runtimes and bundles are released together with the specification after the embargo, not before.

\section{Synthetic population and training}
\label{sec:synthetic}

\subsection{Why synthetic}
\label{sec:synthetic-why}

Real data could not train anything.
The audit of 16 August 2026 and the read-only snapshot of 29 August 2026 both count 46 sessions and 206 responses, all team-and-acquaintance traffic, ``not a research dataset but test traffic'' (Section~\ref{sec:results-real-data}).
There is a deeper reason as well.
``On real preference data there is no answer key: any model output is a hypothesis. So the answer key was written first. A ground truth fixes the population's hidden structure; the generator plays it through the production API; the model is then scored on whether it recovers numbers that were decided in advance.''
We wrote the ground truth before training and kept it out of training: the per-session ground-truth sidecar (segment, weights, cohort) is read by the evaluation code only, the ground-truth directory is read by the evaluation script only, ``never by a training script'', and the feature code states ``Ground truth is not used here.''
Everything in the corpus is synthetic and contains no real personal data; every session is generated deterministically from an integer seed.
``The obstacle to publishing it is priming, not privacy.''

\subsection{Two generation paths}
\label{sec:synthetic-paths}

Both generation paths drive the Allot platform's real production handler, built from the instrument's own repository; ``the business logic is not forked''.
Table~\ref{tab:generation-paths} contrasts them.

\begin{table}[htbp]
  \setstretch{1}
  \centering
  \footnotesize
  \caption{The two generation paths. Both execute the same production handler; the throughput figures are those stated in the training documentation and are revisited in Section~\ref{sec:synthetic-hardware}.}
  \label{tab:generation-paths}
  \begin{tabularx}{\textwidth}{@{}l r X l r@{}}
    \toprule
    Path & Sessions & Route & Throughput & Errors \\
    \midrule
    Live & 505{,}288 & generator $\to$ CDN emulator $\to$ function-URL emulator $\to$ in-memory database emulator $\to$ compressed export & $\sim$5/s & 260{,}712 \\
    Direct & 4{,}500{,}000 & generator $\to$ the same handler in-process, with a mocked cloud SDK over an in-memory store & $\sim$243/s per worker & 0 \\
    \bottomrule
  \end{tabularx}
\end{table}

Live generation runs in batches and resumes from a manifest; if more than half of a batch errors the manifest is not advanced, and ``the seeds are regenerated from scratch''.
Direct generation queues chunks of 25,000 seeds to 14 parallel workers and writes the same export format, plus the ground-truth sidecar.
An equivalence tool compares a live batch with a direct batch on schema and on response distributions; its output for the production run is not among the staged inputs, so what is staged is the tool and the statement that equivalence ``is tested rather than asserted''.

\subsection{Planted ground truth}
\label{sec:synthetic-ground-truth}

The ground truth is a frozen, versioned specification, created on 27 August 2026 before any training and hidden from the loss; ``all distributions below are hidden generator parameters''.
Its values are the answer key, and their publication before the study closes would prime participants (Section~\ref{sec:scope}), so this subsection gives its structure only.

The population consists of eight latent moral segments.
Each segment carries a weight vector over the seven criteria, a choice temperature and a behaviour profile: how engaged its members are, how often they split, refuse or skip, how carefully they answer, how often they reflect and how often they submit identity and demographic data.
Over the segments sit five regional cohorts, each with a language mix over the instrument's six languages and a weighted geography of countries (42 distinct country codes in all), and each tilting the segment shares, so that cohort membership shifts the mix of moral types without determining it.
Into that structure the specification plants biases on purpose: cultural gradients, in which the effective weight of a criterion differs between cohorts; group effects attached to some groups of beneficiary archetypes; and an in-group effect that is weak across the population and concentrated in a minority of it.
The specification also fixes the timing of answers, reconsiderations and dropout through the session, so that the corpus has the shape of real traffic as well as its preferences.

A derived target file, recomputed from the specification, holds the numeric consequences the checks are scored against: the tilted global segment mix, the global effective weights and their ranking, and the per-cohort effective weights.
The specification also states the conclusions the evaluation is meant to recover; three of them have no coded check, which Section~\ref{sec:evaluation-c} discusses.

\subsection{Persona sampler and decision core}
\label{sec:synthetic-decision}

Randomness comes from a seeded stream, deterministic per seed, with separate streams for the persona and for in-session behaviour.
A persona is sampled in a fixed order: cohort, language, country, timezone and region, segment (from share times tilt), device type, operating system, browser, screen and viewport, and browser language (occasionally mismatched).
Then come the optional identity items, the onboarding answers, demographics and expert mode, each with segment-dependent probabilities and with ``prefer not to say'' drawn as a real answer.

Each alternative is scored as a weighted sum of the persona's effective criterion weights times the scenario's scaled decomposition, plus a group term when the alternative has a beneficiary archetype, plus an in-group gain times the identity affinity when an identity distance exists.
How the effective weights are formed is fixed by the specification and withheld with it.
Choice is Gumbel-max sampling at the segment's temperature $\tau$,
\begin{equation}
  \label{eq:gumbel}
  \tilde{u}(a) = u(a) + \tau\, G_a, \qquad G_a = -\log\big(-\log(1 - U_a)\big), \quad U_a \sim \mathcal{U}(0,1),
\end{equation}
with the gap between the best and second-best noise-free utilities deciding, in a fixed order, whether the persona skips (a near tie and a coin flip), refuses (all alternatives unattractive and a segment-dependent propensity), splits (a near tie and a segment-dependent split rate, the allocation proportional to need) or picks the Gumbel-max winner.
Timing is modelled too: decision times are log-normal around an engagement-dependent median, lengthened by fatigue and by close calls, with a share of very fast responses among low-engagement personas; reconsideration events are Poisson with a higher rate on close calls; and a reflection, when submitted, carries a confidence that rises with the utility gap.

A session runs the production lifecycle: create the session, possibly bounce or abandon onboarding, post onboarding, confirm the role and post identity if present, then loop over scenarios under a per-slot dropout hazard, fetching each scenario and posting each response, and finally fetch the debrief and post demographics.
The sidecar records seed, cohort, language, country, segment, weights, temperature, device, role, outcome, answered count and drop slot.

\subsection{Intake and outcome world}
\label{sec:synthetic-outcome-world}

The outcome head needs households and consequences, and the moral ground truth provides neither.
An intake generator draws a synthetic population of household forms over the same five cohorts and geography, with cohort-specific urbanisation, market access, income relative to the basket, family size and displacement mix (withheld with the cohorts); household composition, disability, chronic illness, protection concerns, debt, assets, FCS, rCSI, HDDS, shelter and WASH are then drawn conditionally on the household's gap ratio.
A designed severity for check D7 is written alongside,
\begin{equation}
  \label{eq:gt-severity}
  \mathrm{gt\_severity} = 0.5\, \mathrm{gap\_ratio} + 0.3\, \mathrm{fcs\_deficit} + 0.2\, \operatorname{clip}_{[0,1]}(\mathrm{rCSI}/40).
\end{equation}
We generated 2,000,000 training households and a holdout of 400,000 from a different seed, 2.4 million questionnaires in all; 67.1 per cent have a positive gap, the median gap ratio is 0.28, and 12.9 per cent face poor markets (13.0 per cent in the holdout).

Outcomes come from a separate outcome ground truth.
For a plan with modality $m$, coverage $c$, frequency and duration $d$ months, modality efficiency and the food share of the gap are
\begin{equation}
  \label{eq:eta}
  \eta_m = a_m + b_m\, \mathrm{market\_eff}, \qquad
  f = \operatorname{clip}_{[0.30, 0.85]}\big(0.45 + 0.35\, \mathrm{fcs\_deficit}\big),
\end{equation}
so cash is fully market-dependent ($a = 0$, $b = 1$), vouchers half ($0.50$, $0.50$) and in-kind food not at all ($0.95$, $0$).
Effective doses, with $v$ the monthly value and $g$ the monthly gap, are
\begin{equation}
  \label{eq:dose}
  c_{\mathrm{total}} = \begin{cases} \min(1.2,\, \eta v / g) & \text{cash, voucher} \\ \min(1.2 f,\, \eta v / g) & \text{in-kind food} \end{cases}
  \qquad
  c_{\mathrm{food}} = \begin{cases} c_{\mathrm{total}} & \text{cash, voucher} \\ \min(1.2,\, \eta v / (f g)) & \text{in-kind food,} \end{cases}
\end{equation}
and a household random effect $r = \exp(\mathcal{N}(0, 0.15^2))$ multiplies the dose-response.
Four changes and retention follow,
\begin{align}
  \Delta\mathrm{FCS} &= (62.0 - \mathrm{FCS})_{+}\, \big(1 - e^{-2.2\, r\, c_{\mathrm{food}}}\big)\, m^{\mathrm{fcs}}_{\mathrm{freq}}, \label{eq:dfcs} \\
  \Delta\mathrm{rCSI} &= -\mathrm{rCSI}\, \big(1 - e^{-2.0\, r\, c_{\mathrm{total}}}\big)\, m^{\mathrm{fcs}}_{\mathrm{freq}}, \label{eq:drcsi} \\
  \Delta\mathrm{Assets} &= 0.16\, \min\!\big(2,\, c_{\mathrm{total}}\, d / 6\big)\, m^{\mathrm{assets}}_{\mathrm{freq}}\, (1 - \mathrm{assets}), \label{eq:dassets} \\
  \Delta\mathrm{Debt} &= -\mathrm{debt} \times 0.35\, \big(1 - e^{-1.5\, c_{\mathrm{total}}}\big)\, m^{\mathrm{debt}}_{\mathrm{freq}}, \label{eq:ddebt} \\
  \text{retention} &= \sigma\!\big( -1.2 + 1.4\, [d \geq 3] + 0.10\, (d - 3)_{+} + 2.0\, \Delta\mathrm{Assets} + 0.6\, c_{\mathrm{total}} - 0.8\, \text{shock} \big), \label{eq:retention}
\end{align}
with frequency multipliers (FCS / assets / debt) of 1.10 / 0.60 / 1.15 for monthly, 1.00 / 0.90 / 1.00 for quarterly and 0.90 / 1.35 / 0.80 for lump-sum payment, and a shock-prone share that differs by cohort.
Without assistance retention is zero and consumption drifts, $\Delta\mathrm{FCS} = -4.0\, \mathrm{gap\_ratio}$ and $\Delta\mathrm{rCSI} = 5.0\, \mathrm{gap\_ratio}$.
Measurement noise is then added: Gaussian with standard deviations 3.5, 2.5, 0.04 and 0.15 on the four changes, and retention replaced by a uniform draw with probability 0.03.

Simulation draws five plans per training household, $2{,}000{,}000 \times 5 = 10{,}000{,}000$ rows, uniformly over modality, coverage $\{0.5, 0.8, 1.0\}$, frequency and a duration pool of the menu's $\{3, 6, 12\}$ months plus $\{1, 2\}$, ``so the head learns the $\geq$3 months threshold (Dioptra) that the planner respects'' \cite{ircDioptraCostefficiencyAnalysis2019}; 12 per cent of rows are no-assistance controls, because ``the model must learn the drift''.
Both the noisy observed columns and noiseless true-value columns are written; the mean true $\Delta$FCS at positive coverage is 7.71 and the realised control share 12.0 per cent.
Mechanisms follow the cash-transfer literature \cite{haushoferShorttermImpactUnconditional2016,gentiliniRevisitingCashFood2015}, but the simulator is literature-calibrated, not measured, and nine of the citations in its source register are unverified: ``a mechanism drawn from a paper and parameterised by hand is a hypothesis about the world, not an observation of it. D-checks verify the model recovered the hypothesis.''
That three-month rule is practitioner guidance rather than an estimate, and the extension of the Omega Value to cash is ``a designed assumption verified by check D4, not a sector-validated comparison''.

\subsection{Corpus}
\label{sec:synthetic-etl}

An extract-transform-load step unmarshals the database exports into flat columnar files.
It yields 5,621,417 sessions, 103,937,259 responses, 1,840,198 demographics rows and 5,266,000 ground-truth sidecars.
Session counts are consistent: live and direct together give $505{,}288 + 4{,}500{,}000 = 5{,}005{,}288$ primary sessions, and optional sub-sessions carry their own identifiers, which accounts for the remainder.
Sidecar rows number 5,266,000, that is 766,000 attempted live seeds plus the 4,500,000 direct seeds, since every attempted seed writes a sidecar, including the 260,712 error envelopes.
After the skip exclusion and the lookup match, 97,928,680 choice observations remain, of which 5 per cent, 4,896,434 rows, form the validation split and 93,032,246 the training split.
An earlier internal note said ``5.0 M sessions \ldots\ $\approx$105 M responses''; the measured figures are the ones above.

\subsection{Training schedule}
\label{sec:synthetic-training}

Three training runs produced the three checkpoints (Table~\ref{tab:training-schedule}); the full settings are in Appendix~\ref{app:hyper}.

\begin{table}[htbp]
  \setstretch{1}
  \centering
  \footnotesize
  \caption{Training schedule. All runs used one seed and the NVIDIA JAX 25.04 container image; times are UTC on 27--28 August 2026.}
  \label{tab:training-schedule}
  \begin{tabularx}{\textwidth}{@{}l X X X@{}}
    \toprule
    & v1 moral core & v2 unified model & NVFP4 QAT \\
    \midrule
    Initialisation & random & warm start from v1 & warm start from v2 \\
    Choice stream & 6 epochs, batch 16{,}384, AdamW lr $3 \times 10^{-3}$, wd $10^{-6}$; residual off for the first 2 epochs & 2 epochs, batch 16{,}384, AdamW as v1 & as v2 \\
    Outcome stream & -- & 12 epochs, batch 32{,}768, Adam lr $2 \times 10^{-3}$, $N = 10{,}000{,}000$, split 9{,}500{,}000 / 500{,}000 & as v2 \\
    Finished / pipeline window & 22:19:29 & 05:20:57 $\to$ 05:29:32 (8 m 35 s, six stages) & 05:49:08 $\to$ 05:59:25 (10 m 17 s, four stages) \\
    \bottomrule
  \end{tabularx}
\end{table}

Run v1 trained the moral core alone for six epochs with the residual warm-up described in Section~\ref{sec:allotnet-forward}, within about 10~GB of host memory, finishing at 22:19:29 UTC on 27 August 2026 with a final validation negative log-likelihood of 0.5958 and accuracy of 0.7057.
Lazy batching explains the memory figure: the alternative-feature tensor is never materialised (``at 104 M responses that would be $\sim$50 GB'') but gathered per batch from a per-scenario table indexed by a 16-bit scenario id.
Conditional-logit probes were fitted alongside, for the checks that need them.

Run v2 warm-started from the v1 parameters and trained two optimisers over one bundle: the choice stream for two epochs and the outcome stream for twelve, on 10,000,000 simulated outcome rows of input dimension 31 split 9,500,000 to 500,000.
Its six stages ran from 05:20:57 to 05:29:32 UTC on 28 August 2026: generate 2 million intakes plus the 400,000 holdout; verify the world (7 of 7); simulate 10 million rows; train on the GPU; evaluate the D-checks and the C-regression; and a demonstration of target-group selection at quota 0.15 followed by cycle planning at a budget of 12,000,000.
In one place the training documentation states the outcome batch as 16,384; the command that ran passes 32,768, as the documentation's own command listing and the script agree.
Run NVFP4 warm-started from v2 with the same epochs and batches plus the quantisation flag; its four stages ran from 05:49:08 to 05:59:25 UTC: re-check the v2 baseline with the logit probes in place; train under quantisation; run the C- and D-checks on the quantised model; and run the inference-parity check on 20,000 households.

\subsection{Hardware and wall-clock}
\label{sec:synthetic-hardware}

Everything ran on one NVIDIA DGX Spark with a GB10 Grace-Blackwell processor: Linux 6.17 on aarch64, 20 CPUs, 121~GiB of RAM, driver 580.173.02 and CUDA 13.0 at the time of the snapshot (29 August 2026).

For the live phase no end-to-end wall-clock exists, and its documented throughput of ``$\sim$5/s'' is not reproducible from the manifests, whose first batch implies about 44 sessions per second per generator instance.
The direct phase needs a correction: the documentation presents the sum of the per-chunk durations, 18,567~s, as ``wall clock'', but a progress file written from a single process start records 1,365~s at 3,296.8 sessions per second ($3{,}296.8 \times 1{,}365 \approx 4{,}500{,}000$), so the whole direct run took about 22~min~45~s with 14 workers.
The documentation's summary that ``the whole v2 pipeline \ldots\ completes in under ten minutes'' is consistent with the timestamped GPU windows.

\section{Evaluation protocol}
\label{sec:evaluation}

Evaluation ``is a contract, fixed before training''.
It has two families: the C-checks test the moral core against the derived targets of the planted specification, and the D-checks test the encoder, the outcome head and the planner against the outcome ground truth, with the C-checks re-run on the later model as a regression guard.
Both evaluation scripts take the ground-truth directory as an argument so that the corpus ``can be scored blind by someone who does not read them'', and each writes a JSON and a Markdown report in which ``every number in one appears in the other''.
We count sixteen checks: eight coded C-checks (C1--C6, C9s and C10), seven D-checks (D1--D7) and the C-regression; a seventeenth, D6w, is a design control run before training.
Appendix~\ref{app:checks} tabulates all of them.

\subsection{C-checks}
\label{sec:evaluation-c}

Table~\ref{tab:c-checks-generic} states what each C-check tests.
Directional floors and bands that would reveal the size of a planted effect are withheld with the specification; the thresholds that do not are given.

\begin{table}[htbp]
  \setstretch{1}
  \centering
  \footnotesize
  \caption{The coded C-checks, stated generically. Each threshold was fixed before training; ``floor withheld'' marks a threshold that would reveal the size of a planted effect.}
  \label{tab:c-checks-generic}
  \begin{tabularx}{\textwidth}{@{}l X X@{}}
    \toprule
    Check & What it tests & Coded condition \\
    \midrule
    C1 & the global ranking of the seven criteria & Spearman $\rho$ of the recovered against the planted global weights $\geq 0.85$, and the top three criteria identical \\
    C2 & a planted cultural gradient on one criterion between two cohorts & recovered cohort ratio above a directional floor (floor withheld) \\
    C3 & a second planted cultural gradient, on another criterion & recovered cohort ratio above a directional floor (floor withheld) \\
    C4 & planted group effects on groups of beneficiary archetypes & recovered group contrast beyond a threshold with the planted sign (threshold withheld) \\
    C5 & an in-group effect that exists and is localised & global in-group coefficient above a floor and its maximum over the experts above a higher floor (floors withheld) \\
    C6 & an estimator comparison between two groups of sessions defined by the withheld instrument specification & cosine of one group's fit to the truth $\geq$ cosine of the other's $- 0.005$ \\
    C9s & that the latent classes match the hidden segments & adjusted Rand index $> 0.25$ \\
    C10 & the corpus completion rate & inside a coded band around the design target (band withheld) \\
    \bottomrule
  \end{tabularx}
\end{table}

C1 to C5 read the effective weights from the exported production model; if the export is absent they fall back to the conditional-logit probes and say so in the report header.
C6 needs a two-population logit fit, because the claim concerns an estimator and is not a property of the gate.
C9s needs the gate posteriors of every session.

Two titles overstate their conditions, and the evaluation document says so.
C6's title claims more than its condition: the condition permits the first group to be up to 0.005 \emph{worse} in cosine and still pass, and the measured difference is well inside that tolerance, so C6 confirms `not worse', which is a much weaker statement than its title.
C10's coded band is wider than the design target its title states, so a result can pass C10 while sitting outside the design target, ``which is exactly what happened''.
``Neither is a bug in the code, both bands were chosen to tolerate ordinary sampling variation. But a `\checkmark' against a loose band should not be read as confirmation of a tight claim, and this document exists so it is not.''
Three of the conclusions stated in the specification have no coded check; C9s is the coded substitute for part of one of them.
``8/8'' therefore means eight of eight coded checks, not every conclusion stated in the specification.

\subsection{D-checks}
\label{sec:evaluation-d}

For the D-checks the judge is the noiseless simulator, run without measurement noise and without the household random effect, on the holdout population of 400,000 households; ``the model predicts; the simulator states the truth''.
Probes are 3,000 households sampled from those with a gap ratio above 0.10 and an FCS below 55.
D1 tests a saturating dose-response (rank agreement of predicted with true $\Delta$FCS, and monotonicity in coverage); D2 a frequency contrast (monthly payment raises FCS more, a lump sum raises assets more); D3 a duration threshold at three months in retention; D4 a modality-by-market interaction (in-kind food per dollar beats cash in poor markets, and not the reverse in good ones); D5 economies of scale through the equivalence scale; D6 the planner against an oracle and a naive split; and D7 the encoder (its output against the planted severity of Equation~\ref{eq:gt-severity}, and its percentile term against an exact percentile).
Two design choices deserve a sentence each.
D5 fixes the absolute transfer value and conditions within gap-ratio quartiles ``so that the economies-of-scale mechanism is isolated from the trivial correlation between household size and need''.
D4 evaluates at coverage 0.5, ``outside the dose-response saturation zone, which is where modality choice is actually decided''.
D6's oracle is the same greedy algorithm with the true simulator substituted for the learned head: the welfare of every menu row is computed from simulated outcomes and the greedy allocation is run on it.
``D6 measures search quality, not real-world efficacy \ldots\ The judge is the planted world.''
The planner's $\beta$ is the clean-mode $\beta$ of the export; the oracle and the world verification use the planted global weights.

\subsection{The D6w world control}
\label{sec:evaluation-d6w}

Before training, the v2 pipeline runs the D-checks against the simulator itself, using the planted weights and no model, and adds D6w: the oracle's welfare must exceed 1.3 times the naive equal split.
It answers the question ``does the planted world reward targeting at all?''.
``Without D6w, D6's `99.4\% of oracle' would be unfalsifiable.''

\subsection{The $R^2$ ceiling}
\label{sec:evaluation-ceiling}

Because the outcome targets carry measurement noise by construction, a perfect predictor of the true value would not score $R^2 = 1$ against them.
That ceiling is
\begin{equation}
  \label{eq:r2-ceiling}
  R^2_{\mathrm{ceiling}} = 1 - \frac{\operatorname{Var}(y_{\mathrm{obs}} - y_{\mathrm{true}})}{\operatorname{Var}(y_{\mathrm{obs}})},
\end{equation}
``the $R^2$ a perfect predictor of the true value would score against the noisy target''.
We estimate it two ways.
A measured estimate uses the true-value columns the simulator writes alongside the noisy ones, on exactly the trainer's validation rows (the first $\max(2000, \lfloor 0.05 N \rfloor) = 500{,}000$ rows of the seeded permutation of the training outcome file).
A derived estimate uses the noise parameters of the outcome ground truth: for the additive dimensions the noise variance is the squared standard deviation, and for the retention flip it is $p(1-p)\, \mathbb{E}[(1 - 2r)^2]$ with $p = 0.03$, an approximation of a probability flip.
Agreement between the two is reported as their absolute difference.

\subsection{NVFP4 parity protocol}
\label{sec:evaluation-nvfp4}

Inference parity is checked on 20,000 holdout households: the check builds the full plan menu (1,040,742 plans) and runs the outcome head with fp32 weights and with unpacked NVFP4 weights in two modes: W4A16, four-bit weights with unquantised activations, ``the standard Blackwell production mode'' and the mode on which the verdict rests; and W4A4, activations quantised too, reported for information.
Per outcome dimension it records Pearson and Spearman correlations, the root-mean-square error and the fp32 scale; at the decision level it records the Spearman correlation of plan welfare $W$ under clean-mode $\beta$ and the overlap of the top-decile selections by $W/\mathrm{cost}$.
Originally the coded predicate was uniform,
\begin{equation}
  \label{eq:parity-original}
  \rho_W \geq 0.98 \;\wedge\; \mathrm{overlap}_{10\%} \geq 0.90 \;\wedge\; \forall k:\; \rho_k \geq 0.95,
\end{equation}
and the predicate now in the code is
\begin{equation}
  \label{eq:parity-relaxed}
  \rho_W \geq 0.98 \;\wedge\; \mathrm{overlap}_{10\%} \geq 0.90 \;\wedge\; \forall k \neq \text{retention}:\; \rho_k \geq 0.95 \;\wedge\; \rho_{\text{retention}} \geq 0.70.
\end{equation}
This relaxation is post hoc, and the code says so: ``this leniency for retention was introduced AFTER an initial run failed the uniform $\geq$0.95 bar on retention alone (measured 0.7703), with every other metric passing and byte-identical numbers before and after. The reasoning above stands on its own merits, but the threshold was set post-hoc''.
Retention, the reasoning goes, is sigmoid-compressed, so its rank is unstable among near-equal values, while its influence on decisions is bounded by the $0.5 + 0.5\,\text{retention}$ multiplier and already covered by the welfare rank and the top-decile overlap.
We report both predicates and the order in which they were set.

\subsection{Runtime parity}
\label{sec:evaluation-runtime}

Each runtime must reproduce the eight golden vectors: the derived economics (equivalence scale, household basket, gap), all seven criteria and the clean score at $10^{-4}$; the predicted outcomes of the cash, voucher and in-kind plans at $10^{-3}$; the plan costs at $10^{-4}$; and planner sanity (the budget respected).
The Java and C planners reproduce the Python greedy allocation on shared fixtures.

What the harness cannot tell us is stated in the evaluation document under three headings, which we adopt.
``Nothing about human values'': every check scores recovery of a planted structure.
``Nothing about real-world outcomes'': the head learned a simulator.
``Nothing about the planner in the field'': D6 and D6w live inside the planted world.
Operational use requires retraining the core on live study data and the head on real post-distribution monitoring data.

\subsection{Benchmark protocol}
\label{sec:evaluation-bench}

The first staging of this work found no timing measurement anywhere in the inputs, and the website claimed ``microseconds per household'' without a log behind it.
A second phase closes the gap with one harness for the inference runtimes: every runtime is re-verified on golden sets of 8 and 2,000 cases, and speed is measured on six machines under one protocol.
We follow the practice of the tiny-model benchmarking literature in fixing the workload, the measurement points and the reporting format before any number is quoted \cite{banburyMLPerfTinyBenchmark2021}.

\subsubsection{Harness and machines}

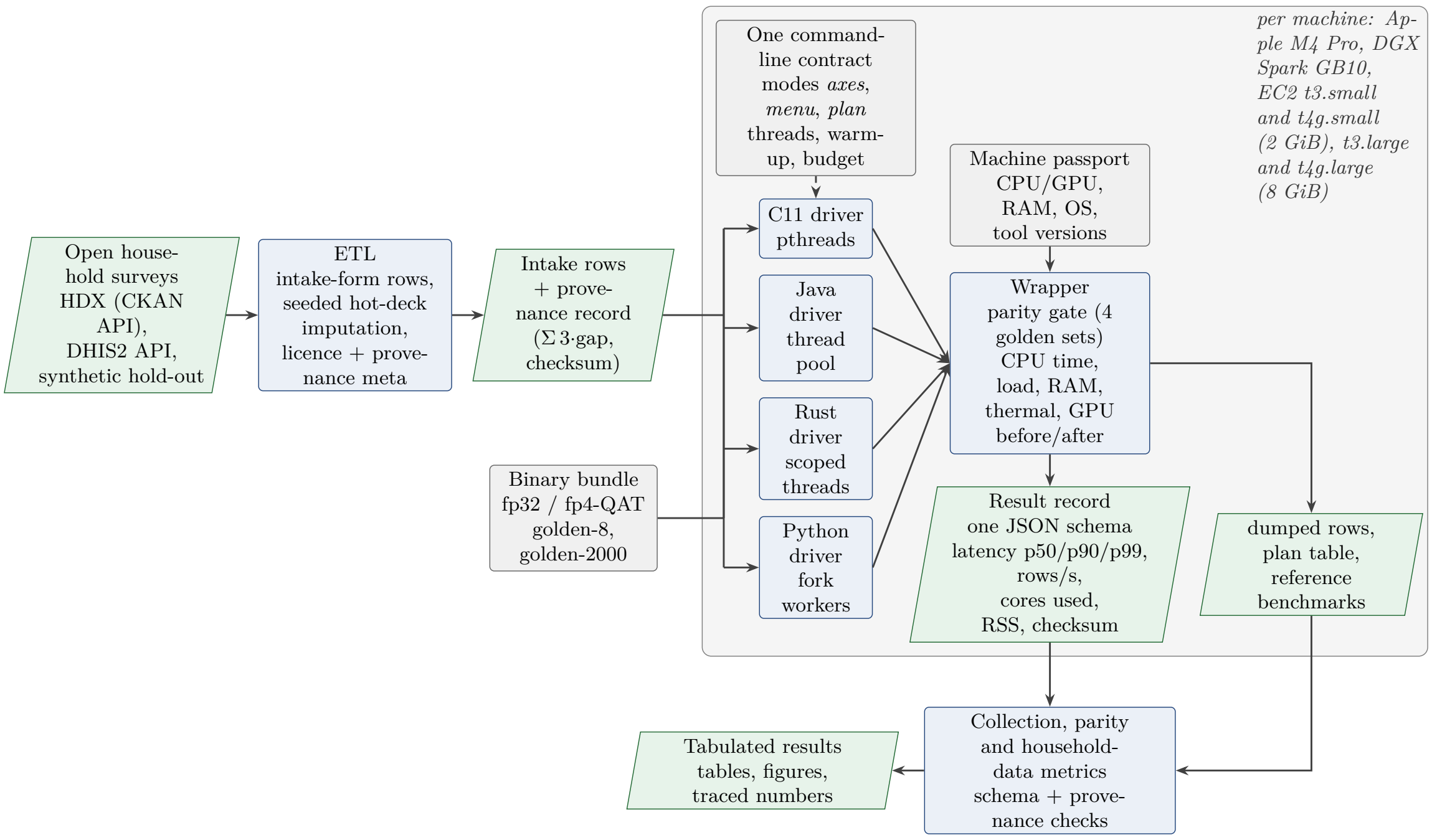
\begin{figure}[htbp]
  \centering
  \begin{allotfit}
  \begin{tikzpicture}[node distance=0.9cm and 0.5cm]
    \node[data, text width=2.6cm] (surveys) {Open household surveys\\HDX (CKAN API), DHIS2 API,\\synthetic hold-out};
    \node[block, text width=2.8cm, right=of surveys] (etl) {ETL\\intake-form rows,\\seeded hot-deck imputation,\\licence + provenance meta};
    \node[data, text width=2.2cm, right=of etl] (jsonl) {Intake rows\\+ provenance record\\($\Sigma\,3\cdot$gap, checksum)};
    \node[config, text width=2.4cm, below=1.3cm of jsonl] (bundle) {Binary bundle\\fp32 / fp4-QAT\\golden-8, golden-2000};
    \node[block, text width=1.55cm, right=1.5cm of jsonl, yshift=1.35cm] (c) {C11 driver\\pthreads};
    \node[block, text width=1.55cm, below=0.25cm of c] (java) {Java driver\\thread pool};
    \node[block, text width=1.55cm, below=0.25cm of java] (rust) {Rust driver\\scoped threads};
    \node[block, text width=1.55cm, below=0.25cm of rust] (py) {Python driver\\fork workers};
    \node[config, text width=2.9cm, above=0.35cm of c] (contract) {One command-line contract\\modes \emph{axes}, \emph{menu}, \emph{plan}\\threads, warm-up, budget};
    \node[block, text width=2.9cm, right=1.2cm of java, yshift=-0.55cm] (wrap) {Wrapper\\parity gate (4 golden sets)\\CPU time, load, RAM,\\thermal, GPU before/after};
    \node[config, text width=2.9cm, above=0.4cm of wrap] (fp) {Machine passport\\CPU/GPU,\\RAM, OS, tool versions};
    \node[data, text width=3.3cm, below=0.5cm of wrap] (res) {Result record\\one JSON schema\\latency p50/p90/p99, rows/s,\\cores used, RSS, checksum};
    \node[data, text width=2.7cm, right=of res] (dump) {dumped rows,\\plan table,\\reference benchmarks};
    \node[block, text width=3.7cm, below=1.0cm of res] (collect) {Collection, parity\\and household-data metrics\\schema + provenance checks};
    \node[data, text width=3.6cm, left=of collect] (csv) {Tabulated results\\tables, figures,\\traced numbers};
    \begin{scope}[on background layer]
      \node[graygroup, fit={(c)(java)(rust)(py)(wrap)(fp)(res)(dump)(contract)},
        label={[grouplabel, anchor=north east, align=left, text width=2.6cm]north east:per machine: Apple M4 Pro, DGX Spark GB10, EC2 t3.small and t4g.small (2~GiB), t3.large and t4g.large (8~GiB)}] (machines) {};
    \end{scope}
    \draw[arrow] (surveys) -- (etl);
    \draw[arrow] (etl) -- (jsonl);
    \coordinate (bus) at ($(c.west)+(-0.55,0)$);
    \draw[thick, draw=black!75] (jsonl.east) -- (jsonl.east -| bus);
    \draw[thick, draw=black!75] (bundle.east) -- (bundle.east -| bus);
    \draw[thick, draw=black!75] (bus |- c.west) -- (bus |- bundle.east);
    \draw[thick, draw=black!75] (bus |- c.west) -- (bus |- py.west);
    \foreach \d in {c,java,rust,py} { \draw[arrow] (bus |- \d.west) -- (\d.west); }
    \draw[dashedarrow] (contract.south) -- (c.north);
    \foreach \d in {c,java,rust,py} { \draw[arrow] (\d.east) -- (wrap.west); }
    \draw[arrow] (fp) -- (wrap);
    \draw[arrow] (wrap) -- (res);
    \draw[arrow] (wrap.east) -| (dump.north);
    \draw[arrow] (res) -- (collect);
    \draw[arrow] (dump.south) |- (collect.east);
    \draw[arrow] (collect) -- (csv);
  \end{tikzpicture}
  \end{allotfit}
  \caption{Benchmark harness.
    Open household surveys and the synthetic hold-out are transformed once into rows of the intake form with provenance metadata; four drivers written against one command-line contract exercise the C, Java, Rust and Python runtimes on the same bundle and rows.
    On every machine a wrapper first runs the parity gate (four golden sets) and then the matrix, recording the machine passport and the resources each invocation actually consumed; results share one JSON schema and are folded into the tables from which every benchmark table, figure and number in this paper is traced.}
  \label{fig:bench-harness}
\end{figure}

Four runtimes read the same binary bundles: the C11 library, the Java library, the Rust crate and a Python package written for the second phase as a row-wise port of the same algorithms on numpy alone.
The Python package also has a vectorised batch path, which we measure separately.
Both bundles carry fp32 tensors; the fp4-QAT bundle differs in its weights and in a flag.
Three modes exercise the three levels of the model.
Mode \emph{axes} derives the economics, encodes the seven criteria and computes the clean score for one household (levels 1 and 2, the moral core on intake).
Mode \emph{menu} adds the full admissible plan menu with predicted outcomes, cost and welfare: $3 \times 3 \times 3 \times 3 = 81$ plans for a household with a positive gap and none otherwise, the per-household part of cycle planning.
Mode \emph{plan} is one planning call over the whole population under a budget, level 3 proper; the greedy is sequential by construction.
Figure~\ref{fig:bench-harness} shows how the pieces fit.

Every driver obeys one command-line contract, and the order of operations is fixed.
A driver loads the bundle, parses the whole intake file into memory before any timing starts (parsing is never inside a latency sample), processes the warm-up rows without recording them, then times every row with a monotonic clock around the model call only.
Statistics are medians over three repeated passes; with $T$ threads the rows are sharded into $T$ contiguous chunks over one immutable model, and \emph{plan} mode ignores $T$.
As a cheap cross-language hint, each driver writes a SHA-256 over its outputs formatted to three decimals; parity proper is decided by the tolerance test below on the dumped rows.
Results share one JSON schema with the build, the machine passport, the bundle, the dataset, the timing block and a resources block; a wrapper runs every driver under the system's resource timer, so that the cores kept busy during the timed passes are recorded, and samples load, memory, thermal state and the GPU before and after every invocation.

Before any speed number is quoted, the wrapper runs a gate: all four runtimes must pass the eight-case golden file and a 2,000-case golden set (2,000 intakes drawn from the training population) on both bundles, and the harness refuses to run the matrix otherwise.

Table~\ref{tab:bench-machines} lists the six machines: four planned ones and, added after the Java out-of-memory failures of Appendix~\ref{app:bench}, two 8~GiB twins of the small cloud instances for the 100,000- and 400,000-row datasets.
We chose them to span a developer laptop, a desk-side accelerator and the two smallest general-purpose cloud instances a field office might rent; no large server is included by design.
The cloud instances were created and terminated under a tagged, scripted lifecycle, and the total on-demand cost stayed within the planned bound of five dollars.

\begin{table}[htbp]
  \setstretch{1}
  \centering
  \footnotesize
  \caption{Benchmark machines.}
  \label{tab:bench-machines}
  \begin{tabularx}{\textwidth}{@{}>{\raggedright\arraybackslash}p{2.3cm} >{\raggedright\arraybackslash}X >{\raggedright\arraybackslash}X@{}}
    \toprule
    Machine & Hardware & OS and toolchain \\
    \midrule
    Apple M4 Pro & MacBook Pro, Apple M4 Pro, 12 cores (8 performance, 4 efficiency), 24~GiB unified memory; matrix run on mains power & macOS 26.6.2; clang 21.0.0, OpenJDK 21.0.9, rustc 1.97.1, Python 3.14.7 with numpy 2.4.2 \\
    DGX Spark GB10 & NVIDIA GB10 (Grace Blackwell): 20 cores (10 Cortex-X925, 10 Cortex-A725), 1,378--3,900~MHz, 121.7~GiB unified; GPU GB10, compute capability 12.1, CUDA 13.0 & Ubuntu 24.04.4 LTS, kernel 6.17; gcc 13.3.0, OpenJDK 21.0.12, rustc 1.98.0, Python 3.12.3 with numpy 2.5.2 on OpenBLAS \\
    EC2 t3.small & 2 vCPU x86-64 (two hyper-threads of one Intel Xeon Platinum 8259CL core), 2~GiB, burstable with unlimited credits & Ubuntu 24.04 LTS; gcc 13.3, OpenJDK 21.0.12, rustc stable, Python 3.12.3 \\
    EC2 t4g.small & 2 vCPU arm64 (two Graviton2 cores, Neoverse-N1), 2~GiB, burstable with unlimited credits & as t3.small \\
    EC2 t3.large & as t3.small with 8~GiB; added for the 100k and 400k cells; JVM heap 5~GiB & as t3.small \\
    EC2 t4g.large & as t4g.small with 8~GiB; as above & as t4g.small \\
    \bottomrule
  \end{tabularx}
\end{table}

Because the machines differ in everything from instruction set to memory bandwidth, each also runs reference benchmarks (sysbench, 7-Zip, OpenSSL and a numpy FP64 GEMM, three repetitions with medians).
The only dense component of AllotNet is the outcome head, so the one GPU measurement is its batched forward pass in JAX on the GB10 for batch sizes $2^{12}$ to $2^{19}$, over the 624,078 admissible (household, plan) rows of 12,000 held-out households, with parity against numpy on the first 100,000 rows.
Not measured, and not claimed: GPU inference of the runtimes (none has a GPU path), GPU speed of the moral core or the planner, and NVFP4 kernel throughput; JAX 0.5 exposes no FP4 GEMM, so the NVFP4 checks of Section~\ref{sec:evaluation-nvfp4} are numpy emulations and establish parity, not speed.

\subsubsection{Parity gate and the 2,000-case golden set}
\label{sec:evaluation-bench-gate}

Eight golden cases never exercised a percentile tie, which is why a larger set was needed.
We exported the 2,000-case set from the v2 and v2-fp4 models after checking that all 19 tensors and the percentile grid of the re-exported bundles are byte-identical to the committed ones; the golden intakes carry none of the ground-truth columns.

The larger set surfaced two portability findings, which we report rather than absorb.
The first is FMA contraction in C.
With clang's default floating-point contraction on arm64, the deprivation index of Equation~\ref{eq:depr-index} is compiled into fused multiply-adds and lands one unit in the last place above the value the Python, Java and Rust runtimes compute.
That would not matter, except that 529 of the 2,000 golden households tie exactly with a point of the 2,001-value percentile grid (which itself has 322 duplicate values), so a one-ulp nudge moves the percentile rank by one or two steps, 0.0005 to 0.001 on that criterion, and 42 of the 2,000 cases failed the $10^{-4}$ test until contraction was disabled in the source and the build.
The second is locale in Java: two formatting routines printed decimal commas in rationales under a decimal-comma locale; the fix has no numeric effect.

\subsubsection{Speed criteria}

Let $\ell_i$ be the per-row latency of the model call.
Percentiles are nearest-rank on the pooled samples of the median pass,
\begin{equation}
  \label{eq:latency-pct}
  p_q = \ell_{(\lceil q\,n \rceil)}, \qquad q \in \{0.5,\, 0.9,\, 0.99\},
\end{equation}
reported with the mean, minimum and maximum.
Throughput is rows per second of the median pass,
\begin{equation}
  \label{eq:throughput}
  \Theta = \frac{n_{\mathrm{timed}}}{T_{\mathrm{wall}}}.
\end{equation}
Speed-ups are taken across languages on one machine, with the row-wise Python package as the reference, and across machines for one language, with the smallest instance as the reference,
\begin{equation}
  \label{eq:speedup}
  S_{L,M} = \frac{\Theta_{L,M}}{\Theta_{\mathrm{python},M}}, \qquad
  S^{M}_{L} = \frac{\Theta_{L,M}}{\Theta_{L,\text{t3.small}}}.
\end{equation}
Thread-scaling efficiency is reported next to the cores kept busy,
\begin{equation}
  \label{eq:scaling-eff}
  E(T) = \frac{\Theta(T)}{T \cdot \Theta(1)},
\end{equation}
and normalised throughput divides by the machine's single-thread 7-Zip rating $Z_M$ in MIPS (and, as a second normaliser, by its FP64 GEMM rate in GFLOPS),
\begin{equation}
  \label{eq:norm-throughput}
  \tilde{\Theta}_{L,M} = \frac{\Theta_{L,M}}{Z_M}.
\end{equation}
The sysbench cpu test, planned as a normaliser, is excluded: its Homebrew build for macOS/arm64 optimises the prime loop away and reports about 16.7 million events per second, three orders of magnitude off the Linux machines.
GEMM is second rather than first because the BLAS back-ends differ (Accelerate on the Mac, OpenBLAS elsewhere).
We set no threshold on speed; the numbers are reported, not judged.

\subsubsection{Parity criteria}
\label{sec:evaluation-bench-parity}

For each language $L$ against the Python package as reference, and for each quantity family $q$ (the eight derived economics, the seven criteria, the clean score, total cost, welfare and the five outcomes), the dumped rows give a worst-case error and a pass rate,
\begin{equation}
  \label{eq:parity-bound}
  e^{\max}_{L,q} = \max_{i,k} \big| y^{L}_{ik} - y^{\mathrm{ref}}_{ik} \big|, \qquad
  \mathrm{PR}_{L,q} = \frac{1}{N} \sum_{i} \mathbf{1}\Big[ \forall k:\; \big| y^{L}_{ik} - y^{\mathrm{ref}}_{ik} \big| \leq \max\big(\tau_q,\; \tau_q\, |y^{\mathrm{ref}}_{ik}|\big) \Big],
\end{equation}
with $\tau_q = 10^{-4}$ for derived fields, criteria, clean score and cost and $10^{-3}$ for outcomes and welfare; menu plans are matched across languages by (modality, coverage, frequency, duration).
The thresholds are a pass rate of 1.000 on the 2,000-case set for every language, identical checksums as a hint, and, for fp32 against fp4, $\rho_s \geq 0.98$ on clean scores with a top-decile overlap of at least 0.90, the acceptance rule of the quantisation check.
One residual is expected and should not be mistaken for a defect.
C, Java and Rust read float32 weights and accumulate in double, whereas the training code that wrote the golden files computed the outcome head in float32, so the golden export differs from the double-precision runtimes at about $10^{-5}$ on outcomes (the worst golden case is $8.6 \times 10^{-6}$ on an outcome and at most $5 \times 10^{-7}$ elsewhere), inside the $10^{-3}$ tolerance.

\subsubsection{Quality criteria on household data}
\label{sec:evaluation-bench-quality}

On the synthetic world the recovery criteria are the checks above.
Real household data carry no answer key, so the clean score $s_i$ can only be compared with the indicators a programme would otherwise rank on: the Food Consumption Score with WFP weights and its 21.5 and 35 thresholds \cite{vamFoodConsumptionScore2024,wiesmannValidationWorldFood2009}, the reduced Coping Strategies Index \cite{maxwellHowIndicatorsHousehold2014,caldwellCopingStrategiesIndex2008}, a CARI-style phase built from FCS and rCSI \cite{wfpConsolidatedApproachReporting2021}, and an asset proxy.
We report Spearman $\rho_s(s, -\mathrm{FCS})$, $\rho_s(s, \mathrm{rCSI})$, $\rho_s$ against the phase, the asset proxy and the gap, Kendall $\tau_b$ against FCS and rCSI, and precision at $k$, $\mathrm{P@}k = |T_k(s) \cap T_k(I)| / k$ with $k = \lfloor 0.15\,n \rfloor$.
Against a reference poor set $P$ (CARI phase 3 or 4, or FCS below 35) and a target set $T$ we report the two classical targeting errors \cite{corniaTwoErrorsTargeting1993,coadyTargetingTransfersDeveloping2004},
\begin{equation}
  \label{eq:ee-ie}
  \mathrm{EE} = \frac{|P \setminus T|}{|P|}, \qquad \mathrm{IE} = \frac{|T \setminus P|}{|T|},
\end{equation}
once at $|T| = k$ and once at $|T| = |P|$, where the two errors coincide.
Monotonicity is checked by perturbation,
\begin{equation}
  \label{eq:monotonicity}
  M_{\mathrm{fcs}} = \frac{1}{n} \sum_i \mathbf{1}\big[ s(h_i \oplus \{\mathrm{fcs} + 5\}) \leq s(h_i) \big], \qquad
  M_{\mathrm{gap}} = \frac{1}{n} \sum_i \mathbf{1}\big[ s(h_i \oplus \{\mathrm{income} - 5\}) \geq s(h_i) \big],
\end{equation}
with income in USD; both are expected to be exactly 1.000 because the encoder is monotone by construction, and anything less flags an ETL or clipping error rather than a model property.
Imputation sensitivity is a bootstrap over the hot-decked fields: $B = 100$ seeded re-draws, each giving a top-15 per cent target set $T_b$; the membership probability $p_i = B^{-1} \sum_b \mathbf{1}[i \in T_b]$, the share of households with $0.1 < p_i < 0.9$ (unstable), and the mean Jaccard index $J(T_b, T_{\mathrm{ref}})$ with a percentile 95 per cent interval.
Finally, fp32 against fp4 on each dataset: $\rho_s(s^{32}, s^{4})$ on clean scores and the top-decile overlap $O_{10}$.

\subsubsection{Cost criteria for cycles and modalities}

With the bundle's overheads $o_m$ (cash 0.08, voucher 0.15, in-kind 0.35) and administrative cost of 15 per household plus 2 per payment cycle, $V_i = \mathrm{value\_month}_i \cdot d_i$ is the value transferred and $C_i$ the cost.
The budget per dataset is a quarter of the population's three-month gap, $B_D = 0.25 \sum_{\mathrm{gap} > 0} 3\,\mathrm{gap}_i$, with $B_D/2$ and $2B_D$ as sensitivity runs.
Budget utilisation is $U = \sum_{i \in T} C_i / B$.
The cost--transfer ratio follows the IRC's cost-efficiency method \cite{ircCostEfficiencyAnalysis2016,ircDioptraCostefficiencyAnalysis2019},
\begin{equation}
  \label{eq:ctr}
  \mathrm{CTR} = \frac{\sum_i V_i}{\sum_i C_i},
\end{equation}
and cost per household-month is $C_{hm} = \sum_i C_i / \sum_i d_i$.
WFP's alpha value for in-kind assistance, the market value of the basket per dollar spent \cite{programmeCashVouchersManual2014}, is $\alpha_i = V_i / C_i$ under the convention that the in-kind basket is worth the gap share it covers.
The omega value \cite{ryckembuschEnhancingNutritionNew2013} is not computable, because the model carries no nutrient composition; the proxy is predicted food-consumption gain per dollar.
Population gap coverage is $\mathrm{GC} = \sum_{i \in T} V_i / \sum_{\mathrm{gap} > 0} 3\,\mathrm{gap}_i$ \cite{networkMinimumExpenditureBasket2020}, model-native welfare per dollar is $W/\$ = \sum_{i \in T} W_i / \sum_{i \in T} C_i$, and modality--market alignment is $A = P(\text{in-kind} \mid \text{poor market}) - P(\text{in-kind} \mid \text{good market})$ \cite{wfpMarketFunctionalityIndex2020}.

The baselines run at the same budget.
Equal cash is the naive baseline as coded: the same monthly cash to every household with a positive gap for three months.
FCS-ranked cash pays households in ascending order of FCS, 100 per cent of the gap, monthly, for three months, until the budget is exhausted.
Random-order cash uses the same package in a seeded random order.
One caveat governs the reading of every welfare column: $W$ is the model judging itself, so on real data the comparison rests on the non-model metrics, CTR, $C_{hm}$, GC and the targeting errors, and $W/\$$ is reported for completeness.

\subsubsection{Household datasets}
\label{sec:evaluation-bench-data}

Four open household surveys, a health-information-system harness and the synthetic hold-out are transformed into intake-form rows by one ETL per source, each recording the source, licence, checksum, raw and clean counts with drop reasons, the column map, per-field provenance (direct, derived or imputed, with the share imputed), the exchange rate (World Bank annual average) and the population's three-month gap.
Table~\ref{tab:bench-datasets} summarises them.

\begin{table}[htbp]
  \setstretch{1}
  \centering
  \footnotesize
  \caption{Household datasets used by the benchmark. $N$ is the raw count and the count after cleaning; the last column names what is read from the source and what is imputed.}
  \label{tab:bench-datasets}
  \begin{tabularx}{\textwidth}{@{}>{\raggedright\arraybackslash}p{2.2cm} >{\raggedright\arraybackslash}X >{\raggedright\arraybackslash}p{1.35cm} >{\raggedright\arraybackslash}p{2.4cm} >{\raggedright\arraybackslash}X@{}}
    \toprule
    Name & Source & Licence & $N$ raw $\to$ clean & Direct and imputed fields \\
    \midrule
    Kismayo 2016 & Somalia, Kismayo IDP settlement assessment, REACH, November 2016 & CC BY-IGO & 1,211 $\to$ 1,154 & FCS and rCSI direct; income an expenditure proxy; market access constant (no item) \\
    Cox's Bazar 2019 & Bangladesh, Cox's Bazar Joint MSNA, refugee households, ISCG/REACH, 2019 & CC0 & 4,824 $\to$ 3,385 & FCS direct; rCSI wholly hot-decked from Kismayo donors by FCS category; income from expenditure brackets; market access from travel time \\
    Libya 2015 & Libya MSNA 2015, REACH & CC BY-IGO & 509 $\to$ 500 & FCS 9 per cent and rCSI 35 per cent imputed; income wholly imputed from a synthetic prior by FCS category \\
    GIFMM 2020 & Colombia, GIFMM joint needs assessment, November 2020 round & CC BY & 3,112 $\to$ 3,101 & FCS direct; rCSI wholly imputed; income = earners $\times$ 2020 minimum wage \\
    DHIS2 harness & DHIS2 Sierra Leone demo instance, live API & demo data & 19,029 & only organisation unit, enrolment date and one under-five from the source; every other field from synthetic rows \\
    synth-1k \ldots\ 400k & first $n$ rows of the 400,000-household synthetic hold-out, ground-truth columns dropped & -- & 1,000 / 10,000 / 100,000 / 400,000 & reference for speed; \emph{plan} mode capped at 100,000 rows \\
    \bottomrule
  \end{tabularx}
\end{table}

Cleaning is the same everywhere (invalid household size, no adult, or refused consent drop a row; identifiers are hashed; no names, free text, phone numbers, coordinates or enumerator identifiers are kept), and imputation is seeded and flagged per field.
The DHIS2 set carries the label its metadata gives it, ``DHIS2-shaped ETL/scale harness, not a needs measurement''.
One basket of USD~38 per adult-equivalent-month (Equation~\ref{eq:meb-hh}) is applied to four countries and years.

\subsubsection{Model versions}
\label{sec:evaluation-bench-versions}

Two model versions appear in the Results.
v2, the export of 28 August 2026, is the model of the preceding sections and of the cross-platform speed matrix.
v2.1, the export of 30 August 2026, is the model of the household-data results.
It keeps the same moral core, expert vectors, gate and percentile grid (13 tensors bit-identical) and adds four things: a specification-driven deprivation index with data-quality weights, population-relative percentiles as the default for population-level calls, an eligibility gate in the planner, and an outcome head continued from v2 on transfer coverages 0.1 to 1.2 (Section~\ref{sec:results-v21-head}).
The changes were decided on 30 August 2026, after the v2 household-data run, on the author's rule that an unsatisfactory quality measurement is a reason to fix the model before writing about it; the v2 run is reported in Section~\ref{sec:results-v21-ablation} as the `before'.
Speed and quality are decoupled on purpose: v2.1 changes neither the head's shape nor the menu nor any per-row arithmetic except the optional quality scaling, so the speed matrix, run on the v2 bundle, describes v2.1 as well.

The quality fields are set by the ETL, not by the model.
Income quality is 1 where income is observed, 0.5 for an expenditure proxy (Kismayo, Cox's Bazar), 0.25 for earner count times minimum wage (GIFMM) and 0 where it is imputed (Libya); FCS and rCSI quality are 1 where observed and 0 per row where hot-decked; the DHIS2 set and the synthetic sets carry 1 everywhere.
A weight scaled to 0 removes the term from its criterion and from the deprivation index and renormalises the remaining weights.
The eligibility gate is evaluated as two extra methods in the cost table: level 2 ranks by clean score with population percentiles and keeps the top 15 or 30 per cent; level 3 plans inside that set with the same budget.
Unspent budget is reported through $U$.

\section{Results}
\label{sec:results}

Every number below is quoted as it appears in the evaluation reports, exports and manifests.
Where two documents disagree, both readings are given and the disagreement is stated rather than resolved.
Because the pre-launch traffic (Section~\ref{sec:results-real-data}) produced no fittable dataset, every model result is on the synthetic corpus or, for the deployed runtimes, on open household surveys that carry no answer key.

\subsection{Pre-launch traffic}
\label{sec:results-real-data}

We read the production tables once, on 29 August 2026, and took only item counts and aggregates over non-personal attributes.
No identity records, demographics content, free text or analytics were read.
The tables hold 46 sessions and 206 responses; 12 sessions hold at least one response and 34 hold none, and six completed the 27-slot core.
Twenty-two sessions ran in Russian and 24 in English.
Creation dates run from 11 May to 23 July 2026.
Of the 206 responses, 190 name a candidate, 11 are skips and five are refusals.

The repository audit of 16 August 2026 reads this traffic as ``not a research dataset but test traffic'': team and acquaintances dominate, and its recommendation is to mark all pre-launch traffic as an internal-testing cohort and to exclude it from the dataset by creation date.
The pre-registered analysis rule quarantines exactly that window, so no session belongs to the pilot as defined.
For this paper the consequence is that no real-data model fit is reported.

\subsection{Synthetic corpus and the live-phase failure}
\label{sec:results-corpus}

\begin{figure}[htbp]
  \centering
  \includegraphics[width=0.7\textwidth]{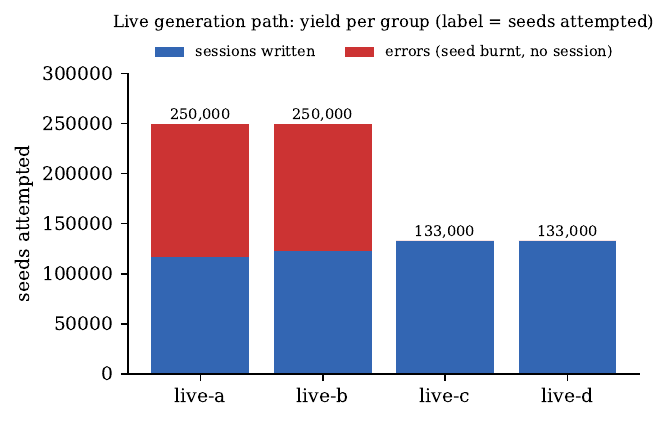}
  \caption{Live generation path: sessions written (blue) and seeds burnt without a session (red) per shard group, stacked to the seeds attempted (label above each bar). live-a and live-b attempted 250,000 seeds each and wrote 117,109 and 122,179 sessions with 132,891 and 127,821 errors; the makeup groups live-c and live-d attempted 133,000 each and wrote 133,000 with zero errors counted.}
  \label{fig:live-yield}
\end{figure}

Two generation paths produced the corpus (Section~\ref{sec:synthetic-paths}).
Table~\ref{tab:corpus} gives the yield of both, and Figure~\ref{fig:live-yield} the live yield per shard group.
The live path was planned as two groups of 250,000 seeds (live-a and live-b) and reached its 505,288 sessions only after two makeup groups (live-c and live-d) of 133,000 seeds each.

\begin{table}[htbp]
  \centering
  \caption{Synthetic corpus yield. Live-phase rows aggregate the per-batch outcomes in each group's manifest; the direct phase ran 180 chunks of 25,000 sessions.}
  \label{tab:corpus}
  \small
  \begin{tabular}{lrrrr}
    \toprule
    Group & Seeds attempted & Sessions written & Errors & Responses \\
    \midrule
    live-a & 250,000 & 117,109 & 132,891 & 2,333,540 \\
    live-b & 250,000 & 122,179 & 127,821 & 2,428,997 \\
    live-c & 133,000 & 133,000 & 0 & 2,654,039 \\
    live-d & 133,000 & 133,000 & 0 & 2,641,595 \\
    \midrule
    live total & 766,000 & 505,288 & 260,712 & 10,058,171 \\
    direct phase & 4,500,000 & 4,500,000 & 0 & 89,662,993 \\
    \bottomrule
  \end{tabular}
\end{table}

Batch-level manifests locate the failure.
live-a's batches 0007--0017 errored on every seed and wrote nothing; live-b errored on 9,358 of 12,000 seeds in batch 0007 and yielded nothing from 0008 to 0017.
The training documentation states that both groups ``produced zero sessions for eleven consecutive batches, 0007 through 0017''; that is exact for live-a, but live-b's batch 0007 still yielded 2,642 sessions.
The cause was the in-memory database emulator restarting under its cgroup memory limit and losing its reference tables, after which every session failed; in the words of the training log, ``On a 124 GB host, V8 chooses an enormous heap ceiling and barely collects''.
The fix capped the heap, set a container memory limit and shrank the live batch from 25,000 to 12,000 seeds, and the makeup groups ran under a supervisor that re-seeds the shards and resumes from the manifest.
The error recurred during the makeup groups (live-d's log records 196,186 orphaned rows exported to a recovery directory), and both groups nevertheless closed at 133,000 sessions with zero errors counted, because the generator does not advance the manifest on an unhealthy exit.
The manifest's completion counter counts seeds attempted, not sessions written, so the per-batch outcome counts are the figures to use; the target was met by repair, not by a clean run; and earlier internal figures of ``0 errors'' have been corrected, since they described a 600-session behaviour validation.
On the direct path, 180 chunks of 25,000 gave 4,500,000 sessions with zero errors and 89,662,993 responses.
After ETL the corpus holds 5,621,417 sessions, 103,937,259 responses and 97,928,680 usable choice observations.

\subsection{Training}
\label{sec:results-training}

The v1 moral core had converged by epoch 4: validation NLL 0.5959 at epoch 4 and 0.5958 at epoch 5, with accuracy identical to four decimals.
Final validation NLL is 0.5958 and validation accuracy 0.7057.
Per-epoch NLL for epochs 0--3 is not among the staged inputs, so no v1 learning curve is shown.
After the two refresh epochs of the choice stream, v2 reports validation NLL 0.5958 and accuracy 0.7057, ``identical to v1 to four decimals''; the exports differ only from the fifth decimal on.

\begin{figure}[htbp]
  \centering
  \includegraphics[width=0.75\textwidth]{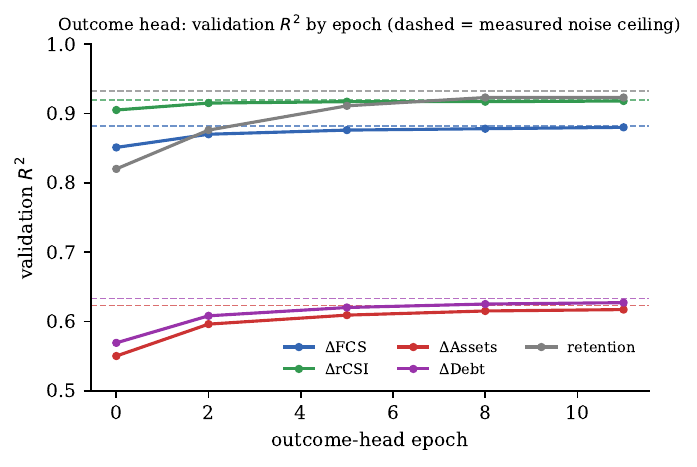}
  \caption{Outcome head: validation $R^2$ at the five recorded epochs (0, 2, 5, 8, 11) for the five outcome dimensions, with the measured noise ceiling of each dimension as a dashed line (0.8815, 0.9195, 0.6230, 0.6328 and 0.9329 for $\Delta$FCS, $\Delta$rCSI, $\Delta$Assets, $\Delta$Debt and retention). The final values are 0.880, 0.918, 0.617, 0.627 and 0.923.}
  \label{fig:v2-r2-curves}
\end{figure}

The v2 outcome head was trained for 12 epochs; Table~\ref{tab:r2-epochs} gives its validation $R^2$ on 500,000 validation rows at the recorded epochs and Figure~\ref{fig:v2-r2-curves} plots them against the measured noise ceilings of Section~\ref{sec:results-ceiling}.
Only epochs 0, 2, 5, 8 and 11 are recorded in the staged sources; the figure plots these five points and interpolates nothing.
At four decimals the export gives the final values as 0.8796, 0.9184, 0.6172, 0.6271 and 0.9228.

\begin{table}[htbp]
  \centering
  \caption{v2 outcome head: validation $R^2$ by epoch on 500,000 validation rows, as recorded in the training documentation.}
  \label{tab:r2-epochs}
  \begin{tabular}{lccccc}
    \toprule
    Epoch & $\Delta$FCS & $\Delta$rCSI & $\Delta$Assets & $\Delta$Debt & retention \\
    \midrule
    0          & 0.851 & 0.905 & 0.550 & 0.569 & 0.820 \\
    2          & 0.870 & 0.915 & 0.596 & 0.608 & 0.876 \\
    5          & 0.876 & 0.917 & 0.609 & 0.620 & 0.911 \\
    8          & 0.878 & 0.917 & 0.615 & 0.625 & 0.923 \\
    11 (final) & 0.880 & 0.918 & 0.617 & 0.627 & 0.923 \\
    \bottomrule
  \end{tabular}
\end{table}

For the NVFP4 quantisation-aware model, the choice stream reports validation NLL 0.6002 and accuracy 0.7024, and outcome $R^2$ of 0.8336, 0.8946, 0.5619, 0.6093 and 0.8627 for $\Delta$FCS, $\Delta$rCSI, $\Delta$Assets, $\Delta$Debt and retention.
These are the QAT model's values under fake-quantised training and sit below the fp32 figures.
Wall-clock, the v2 pipeline ran in 8~min 35~s for six stages, including 2.4~M intakes, 10~M outcomes, training, all checks and the planner demonstration; the fp4 pipeline in 10~min 17~s.

\subsection{Moral core: recovery of the planted structure}
\label{sec:results-core}

All eight coded checks on the moral core passed on the v1 evaluation, with AllotNet as the primary model and the conditional-logit probes as a diagnostic control.
Table~\ref{tab:c-checks} gives the model-side statistics that do not encode the answer key; Section~\ref{sec:results-cregr} reports the same checks re-run on the v2 and NVFP4 models.

\begin{table}[htbp]
  \centering
  \caption{Moral-core checks against the planted ground truth (v1 evaluation). Statistics that would reveal the size or direction of a planted effect are withheld; the verdict and its caveat are given for every check.}
  \label{tab:c-checks}
  \footnotesize
  \begin{tabularx}{\textwidth}{@{}l X l X@{}}
    \toprule
    Check & Quantity & Model & Verdict \\
    \midrule
    C1  & Spearman $\rho$ of the criterion ranking (cosine) & 1.000 (0.997) & pass; ranking exact, $n$ = 97,928,680 \\
    C2  & first planted cultural gradient & withheld & pass; recovered to the planted value at the reported precision \\
    C3  & second planted cultural gradient & withheld & pass; direction right, magnitude overshoots the planted value \\
    C4  & planted group effect & withheld & pass \\
    C5  & in-group effect, global and per expert & withheld & pass; localised in one expert of eight \\
    C6  & cosine to truth, one group of sessions against the other & withheld & pass; a null result (title overstates) \\
    C9s & adjusted Rand index (purity) & 0.671 (0.835) & pass; threshold 0.25 \\
    C10 & corpus completion rate & withheld & pass of the coded band; outside the design target \\
    \bottomrule
  \end{tabularx}
\end{table}

The model's ranking of the seven criteria equals the planted ranking: Spearman $\rho$ = 1.000, cosine 0.997, $n$ = 97,928,680, choice accuracy 0.706.
According to the evaluation report the accuracy is a property of the data rather than a shortfall: it ``reflects the deliberate noise in the segments' decision temperatures''; ``a model that predicted individual choices much better than this would be fitting noise''.
Both planted cultural gradients are recovered in the right direction; the first lands on its planted value at the precision the report gives, and the second overshoots it, a pass only because its coded floor sits well below the planted value.
Per cohort, the recovered criterion vectors agree with the designed cohort weights at cosine 0.995 to 0.998 and Spearman $\rho$ 0.964 to 1.000 across the five cohorts, over the 42 countries with at least 1,500 responses.

The planted group effect is recovered beyond its threshold, and the in-group effect is small globally and concentrated in one of the eight experts; in the architecture note's words, ``the model did not smear the bias across the population, it localised it''.
That is the precondition for clean mode (Section~\ref{sec:allotnet-clean}): an effect the model can point to is an effect it can exclude.
The C6 comparison is a null: the difference lies inside the tolerance.

With $K$ = 8 latent classes, segment recovery reaches purity 0.835 and adjusted Rand index 0.671 over 4,479,696 sessions with a matching ground-truth row.
Segment membership was never a label in the loss; as the report puts it, ``The gate learned it from context covariates alone.''
The confusion concentrates in two pairs of segments.
Completion passes its coded band but sits outside the design target, because truncated live-phase sessions count as non-completions; the direct phase alone lies inside the target, and both readings travel together.
Each of the eight experts reads as one of the planted segments, which is the answer key in another form and is withheld until the embargo lifts.

\subsection{Outcome head against the achievable ceiling}
\label{sec:results-ceiling}

On the fp32 v2 model the D-checks passed eight of eight (D1--D7 plus the C-regression guard); Table~\ref{tab:d-checks} lists them.

\begin{table}[htbp]
  \centering
  \caption{D-checks on the fp32 v2 model. D6 ran on 12,000 households with a budget of 471,115.}
  \label{tab:d-checks}
  \footnotesize
  \begin{tabular}{p{0.16\textwidth}p{0.50\textwidth}p{0.22\textwidth}}
    \toprule
    Check & Result & Threshold \\
    \midrule
    D1 saturating dose-response & Spearman(pred, true) = 0.967; monotonicity 1.000 & $\rho \geq 0.90$; mono $\geq 0.97$ \\
    D2 frequency contrast & FCS(monthly $-$ lump sum) = 3.27; Assets(lump sum $-$ monthly) = 0.0609 & $\geq 0.5$; $\geq 0.015$ \\
    D3 duration threshold & retention(6 mo) $-$ retention(1 mo) = 0.414; jump 3 mo $-$ 2 mo = 0.314 & $\geq 0.1$; $\geq 0.05$ \\
    D4 modality $\times$ market & poor markets ($\Delta$FCS/\$ in-kind) / ($\Delta$FCS/\$ cash) = 1.39; good markets (composite/\$ cash) / (in-kind) = 1.18 & $\geq 1.15$; $\geq 0.95$ \\
    D5 equivalence scale & Spearman($\Delta$FCS/headroom, equivalence size $\mid$ fixed value 50, within gap quartiles) = $-0.777$ & $\leq -0.5$ \\
    D7 encoder & Spearman(encoder output, planted severity) = 0.976; max $|$percentile error$|$ = 0.0000 & $\geq 0.85$; $\leq 0.02$ \\
    D6 planner & planner 918.1, oracle 923.4 (share 0.994), naive 665.4 (ratio 1.38) & $\geq 0.90$; $\geq 1.15$ \\
    C-regr & C-checks re-run, 8/8 & all 8 \\
    \bottomrule
  \end{tabular}
\end{table}

The results document singles out two of these.
D3's 3-month jump of 0.314 against a 0.05 floor ``is the cleanest of these: the head learned a genuine discontinuity at the Dioptra threshold'', because the training duration pool included one and two months.
D7's exact percentile error of 0.0000 ``is not a fitting achievement'': the encoder term is defined as a population percentile, so it is ``a correctness check on the encoder, not evidence about the model''.

\begin{figure}[htbp]
  \centering
  \includegraphics[width=0.75\textwidth]{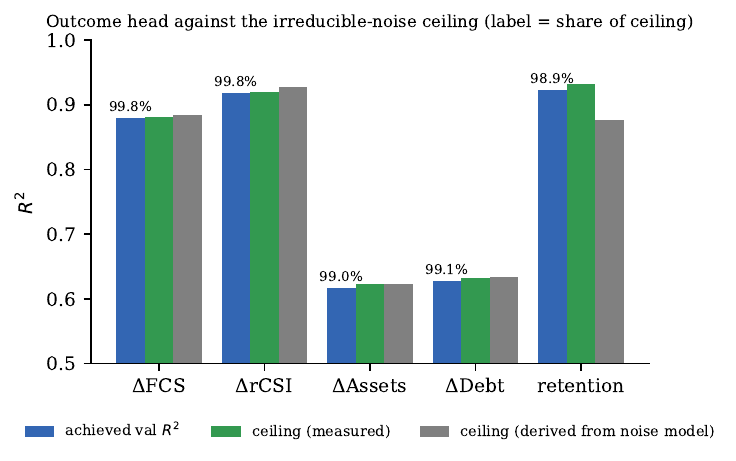}
  \caption{Outcome head against the irreducible-noise ceiling: achieved validation $R^2$ (0.880, 0.918, 0.617, 0.627, 0.923), measured ceiling (0.8815, 0.9195, 0.6230, 0.6328, 0.9329) and ceiling derived from the noise model (0.8847, 0.9275, 0.6229, 0.6332, 0.8762) for $\Delta$FCS, $\Delta$rCSI, $\Delta$Assets, $\Delta$Debt and retention. Labels give the achieved share of the measured ceiling, 99.83, 99.84, 99.04, 99.08 and 98.94 per cent, printed to one decimal.}
  \label{fig:outcome-ceiling}
\end{figure}

Table~\ref{tab:ceiling} sets the final validation $R^2$ of each outcome dimension against the achievable ceiling ($n_{\text{total}}$ = 10,000,000, $n_{\text{val}}$ = 500,000); Figure~\ref{fig:outcome-ceiling} plots the same three quantities.
On $\Delta$FCS, $\Delta$rCSI, $\Delta$Assets, $\Delta$Debt and retention the head reaches 99.8, 99.8, 99.0, 99.1 and 98.9 per cent of the measured ceiling.

\begin{table}[htbp]
  \centering
  \caption{Validation $R^2$ of the outcome head against the achievable ceiling. The measured ceiling uses the simulator's true-value columns on the trainer's 500,000 validation rows; the derived ceiling comes from the noise parameters; agreement is their absolute difference.}
  \label{tab:ceiling}
  \footnotesize
  \setlength{\tabcolsep}{4pt}
  \resizebox{\textwidth}{!}{%
  \begin{tabular}{lrrrrrlr}
    \toprule
    Dimension & Val.\ $R^2$ & Ceiling (measured) & \% of ceiling & Ceiling (derived) & Agreement & Noise model & sd(observed) \\
    \midrule
    $\Delta$FCS    & 0.880 & 0.8815 & 99.8\% & 0.8847 & 0.0033 & additive sd = 3.5  & 10.308506 \\
    $\Delta$rCSI   & 0.918 & 0.9195 & 99.8\% & 0.9275 & 0.0080 & additive sd = 2.5  & 9.286225 \\
    $\Delta$Assets & 0.617 & 0.6230 & 99.0\% & 0.6229 & 0.0000 & additive sd = 0.04 & 0.065141 \\
    $\Delta$Debt   & 0.627 & 0.6328 & 99.1\% & 0.6332 & 0.0004 & additive sd = 0.15 & 0.247668 \\
    retention      & 0.923 & 0.9329 & 98.9\% & 0.8762 & 0.0567 & Bernoulli flip $p$ = 0.03 & 0.264691 \\
    \bottomrule
  \end{tabular}}
\end{table}

The results document reads the table as follows: ``the head reaches 98.9--99.8\% of the achievable $R^2$ on every dimension''; the $\Delta$Assets ceiling is 0.623 because the simulator adds an asset-measurement standard deviation of 0.04 against an observed one of 0.065; and the plateau at epoch 5 ``was the model arriving at the ceiling, not stalling below it''.
On four of the five dimensions the two ceiling estimates agree within 0.008.
Retention differs by 0.057 because the derived estimate models the flip as additive Bernoulli variance; the measured figure is the one to use.

A correction is recorded rather than replaced: an earlier draft called the assets and debt dimensions ``moderate'' with ``38\% of variance unexplained'', which the current text calls ``a misreading: 38\% of the variance is unexplainable by anything''.

\subsection{Planner demonstration and controls}
\label{sec:results-planner}

The world control ran the D-checks against the simulator itself with the planted coefficients, before training, and passed seven of seven: D1 $\rho$ = 1.000 and monotonicity 1.000; D2 3.32 and 0.0592; D3 0.412 and 0.328; D4 poor 1.38 and good 1.17; D5 $-0.831$; D7 0.976 and 0.0000; and D6w oracle 380.4 against naive 274.2, a ratio of 1.39 over a floor of 1.3, on 12,000 households with a budget of 471,115.
So the planted world rewards targeting by a factor of 1.39 before any model is involved.

D6 on the learned head gives planner welfare 918.1 against oracle 923.4, or 99.4 per cent of oracle, and against naive 665.4, or 1.38 times.
The results document restricts the reading: ``as a statement about \emph{search} it is [near-optimal] \ldots What it is not is evidence that the planner would help real households. The oracle knows the planted world; both live inside it.''
D6w establishes 1.39 times of headroom; D6 shows that the planner finds 99.4 per cent of it.
The D6 welfare totals (918.1, 923.4, 665.4) and the D6w totals (380.4, 274.2) are on different scales, because D6 uses the learned clean-mode coefficients and D6w the planted ones; the ratios, not the totals, are comparable.

\begin{figure}[htbp]
  \centering
  \includegraphics[width=0.8\textwidth]{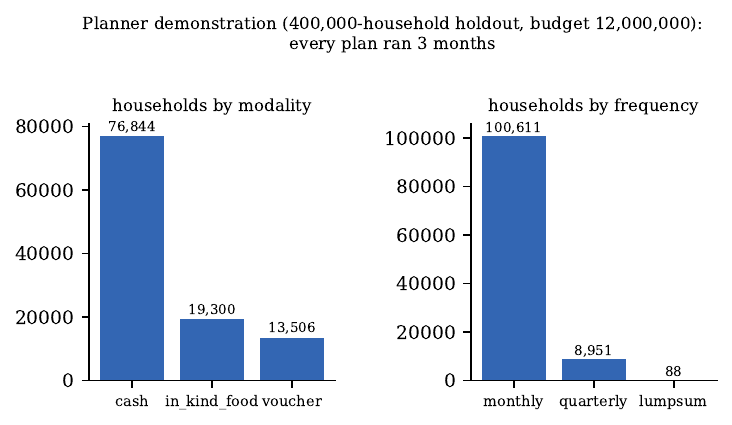}
  \caption{Planner demonstration on the 400,000-household holdout with a budget of 12,000,000: households by modality (cash 76,844; in-kind food 19,300; voucher 13,506) and by frequency (monthly 100,611; quarterly 8,951; lump sum 88). All 109,650 plans ran for 3 months.}
  \label{fig:planner-mix}
\end{figure}

Table~\ref{tab:planner} summarises the planner demonstration itself (v2 clean mode, 400,000-household holdout) and Figure~\ref{fig:planner-mix} shows the modality and frequency mix.

\begin{table}[htbp]
  \centering
  \caption{Planner demonstration on the 400,000-household holdout (v2 clean mode).}
  \label{tab:planner}
  \small
  \begin{tabular}{lp{0.62\textwidth}}
    \toprule
    Metric & Value \\
    \midrule
    Budget & 12,000,000 USD; spent 11,999,996 (100.0\%) \\
    Households with a gap & 272,508 \\
    Households assisted & 109,650 (40.2\%) \\
    Total model welfare & 25,776.77 \\
    Modality mix & cash 76,844; in-kind food 19,300; voucher 13,506 \\
    Frequency mix & monthly 100,611; quarterly 8,951; lump sum 88 \\
    Durations & 3 months: 109,650 (all) \\
    Mean value per month & 25.99 USD; mean duration 3.0 months \\
    Objective & $W = (0.5 + 0.5 \cdot \text{retention}) \cdot \sum_{j} \beta^{+}_j (a_{j,\text{pre}} - a_{j,\text{post}})$; greedy by $W$/cost \\
    \bottomrule
  \end{tabular}
\end{table}

A three-month duration for every plan is an artefact of the budget, not a finding about duration: ``it is the budget constraint binding against a saturating dose-response. Serving twice as many households for 3 months buys more closed need than serving half as many for 6''.
The correct reading is that ``under this budget and this objective, breadth dominates depth, down to the duration floor''.
In-kind food takes 17.6 per cent of assignments (19,300 of 109,650), concentrated in poor markets as D4 predicts.

\subsection{The moral ranking and the allocation queue are different orderings}
\label{sec:results-queue-score}

The planner demonstration ranks plans, not households, and over the full holdout that distinction is large enough to change who is served.
Of the 1{,}000 households carrying the best clean-mode moral scores, 384 received a plan.
The 109{,}650 assisted households span moral ranks 4 to 398{,}017 of 400{,}000, with a median of 105{,}556.
This is not a defect in the allocator.
The greedy ranks menu rows by welfare per dollar, so the object being ordered is the pair of a household and a plan, never the household on its own: the moral score orders need, and the queue orders purchasable closure of need per dollar.

Why the other 616 were passed over is a matter of cost rather than merit.
At an equal moral score (median 3.2533 among those excluded against 3.2484 among those served) and with marginally worse need (gap ratio 0.8858 against 0.8565, deprivation 0.7702 against 0.7598), the excluded households carry a median minimum expenditure basket of 163.34 against 105.41, a ratio of 1.55.
The basket scales with household size while welfare is bounded by the headroom left on the criteria, so at a fixed budget the allocator prefers smaller households systematically.
It is a consequence of the objective rather than of the moral core, which never sees household size.

\begin{table}[htbp]
  \centering
  \caption{Three households from the holdout, each selected by a deterministic rule: A is the best-served household by moral rank, B the most expensive exclusion inside the top 1{,}000, C the best value per dollar in the run. Both served plans are cash over three months. Costs are in USD; MEB is the household's minimum expenditure basket and $W$ the welfare the plan buys. Per-criterion values are withheld.}
  \label{tab:queue-worked-example}
  \small
  \begin{tabular}{lrrrlrrr}
    \toprule
    Case & Rank & Gap & MEB & Served & Cost & $W$ & $W$/100 \\
    \midrule
    A & 4 & 0.818 & 78.50 & 32.1/mo & 125.01 & 0.343 & 0.274 \\
    B & 281 & 0.972 & 271.97 & no & --- & --- & --- \\
    C & 221{,}544 & 0.056 & 106.13 & 5.9/mo & 40.12 & 0.370 & 0.922 \\
    \bottomrule
  \end{tabular}
\end{table}

Table~\ref{tab:queue-worked-example} carries the argument in three rows.
Household B is needier than A on the gap ratio and on the encoder's scores, and is not served, because its basket is three and a half times A's.
Household C sits in the bottom half of the moral ranking with almost no gap to close, and it returns more absolute welfare than the fourth most deserving household in the holdout (0.370 against 0.343) at a third of the cost, because a small transfer closes a small gap completely.
A needs-only score can be progressive while the queue built on top of it is not; they are not the same object.

\subsection{NVFP4: quantisation-aware checks and inference parity}
\label{sec:results-nvfp4}

Under quantisation-aware training the NVFP4 model passed the D-checks eight of eight; Table~\ref{tab:fp4-dchecks} sets its margins beside the fp32 model's.
The results document notes that ``D5 lost roughly half its margin ($-0.777 \rightarrow -0.611$ against a $-0.50$ floor) and D3's jump narrowed by a third \ldots the equivalence-scale mechanism is the one most degraded by 4-bit training.''

\begin{table}[htbp]
  \centering
  \caption{D-checks of the fp32 and NVFP4-QAT models. Bold marks the two margins that narrowed most under QAT.}
  \label{tab:fp4-dchecks}
  \footnotesize
  \resizebox{\textwidth}{!}{%
  \begin{tabular}{p{0.30\textwidth}llp{0.16\textwidth}}
    \toprule
    Check & fp32 & NVFP4-QAT & Threshold \\
    \midrule
    D1 $\rho$ / monotonicity & 0.967 / 1.000 & 0.968 / 0.999 & $\geq 0.90$ / $\geq 0.97$ \\
    D2 FCS(monthly $-$ lump) / Assets(lump $-$ monthly) & 3.27 / 0.0609 & 3.33 / 0.0638 & $\geq 0.5$ / $\geq 0.015$ \\
    D3 retention(6 $-$ 1 mo) / jump 3 $-$ 2 mo & 0.414 / 0.314 & 0.401 / \textbf{0.228} & $\geq 0.1$ / $\geq 0.05$ \\
    D4 poor / good & 1.39 / 1.18 & 1.50 / 1.16 & $\geq 1.15$ / $\geq 0.95$ \\
    D5 $\rho$ & $-0.777$ & $\mathbf{-0.611}$ & $\leq -0.50$ \\
    D7 $\rho$ / percentile error & 0.976 / 0.0000 & 0.976 / 0.0000 & $\geq 0.85$ / $\leq 0.02$ \\
    D6 planner / oracle / naive & 918.1 / 923.4 (0.994) / 665.4 (1.38) & 908.8 / 921.5 (0.986) / 664.0 (1.37) & $\geq 0.90$ / $\geq 1.15$ \\
    \bottomrule
  \end{tabular}}
\end{table}

\begin{figure}[htbp]
  \centering
  \includegraphics[width=0.54\textwidth]{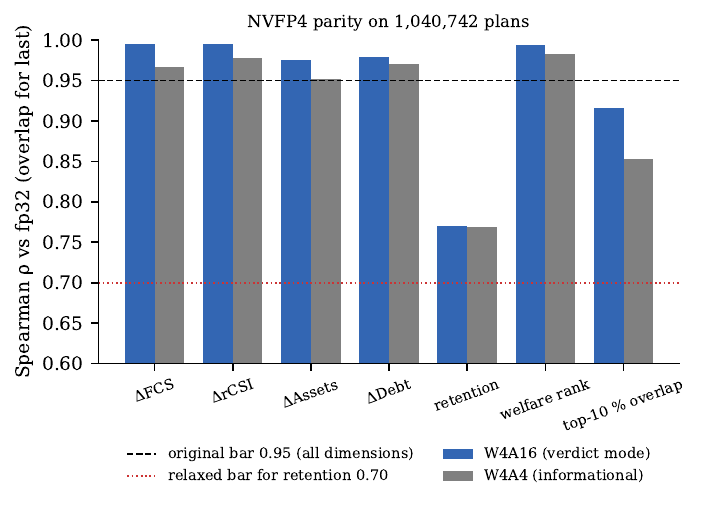}
  \caption{NVFP4 parity on 1,040,742 plans: Spearman $\rho$ against fp32 per outcome dimension, welfare-rank $\rho$ and top-decile overlap, for W4A16 (verdict mode: 0.9949, 0.9954, 0.9751, 0.9795, 0.7703, 0.9938, 0.9159) and W4A4 (informational: 0.9672, 0.9777, 0.9522, 0.9699, 0.7695, 0.9827, 0.8524). The dashed line is the original uniform bar of 0.95, the dotted red line the relaxed bar of 0.70 for retention.}
  \label{fig:nvfp4-parity}
\end{figure}

Inference parity was measured on 1,040,742 holdout plans in two modes: W4A16, with 4-bit weights and unquantised activations, and W4A4, with activations also quantised.
Table~\ref{tab:nvfp4} gives the full report and Figure~\ref{fig:nvfp4-parity} the Spearman columns.
W4A16 ``preserves the decisions that matter: plan welfare ordering at $\rho$ = 0.9938, and 91.6\% of the top-decile selections unchanged''; its retention rank correlation of 0.7703 is ``materially lower than every other dimension''.
Its current report carries a passing verdict.

\begin{table}[htbp]
  \centering
  \caption{NVFP4 inference parity against fp32 on 1,040,742 holdout plans. Bold marks the values the verdict history below turns on.}
  \label{tab:nvfp4}
  \small
  \begin{tabular}{llrrrr}
    \toprule
    Mode & Dimension & Pearson & Spearman & RMSE & fp32 std \\
    \midrule
    W4A16 & $\Delta$FCS    & 0.9955 & 0.9949 & 2.1809 & 10.3314 \\
    W4A16 & $\Delta$rCSI   & 0.9939 & 0.9954 & 1.0156 & 7.1919 \\
    W4A16 & $\Delta$Assets & 0.9866 & 0.9751 & 0.0156 & 0.0706 \\
    W4A16 & $\Delta$Debt   & 0.9897 & 0.9795 & 0.0691 & 0.2063 \\
    W4A16 & retention      & 0.7539 & \textbf{0.7703} & 0.2185 & 0.1278 \\
    W4A16 & welfare rank $\rho$ / top-decile overlap & -- & \textbf{0.9938} / \textbf{0.9159} & -- & -- \\
    \midrule
    W4A4  & $\Delta$FCS    & 0.9658 & 0.9672 & 3.0495 & 10.3314 \\
    W4A4  & $\Delta$rCSI   & 0.9744 & 0.9777 & 1.6880 & 7.1919 \\
    W4A4  & $\Delta$Assets & 0.9705 & 0.9522 & 0.0189 & 0.0706 \\
    W4A4  & $\Delta$Debt   & 0.9821 & 0.9699 & 0.0774 & 0.2063 \\
    W4A4  & retention      & 0.7514 & 0.7695 & 0.2109 & 0.1278 \\
    W4A4  & welfare rank $\rho$ / top-decile overlap & -- & 0.9827 / 0.8524 & -- & -- \\
    \bottomrule
  \end{tabular}
\end{table}

That verdict has a history.
The first quantisation run failed a uniform bar of $\rho \geq 0.95$ on retention alone, at 0.7703; a later run passed with byte-identical metrics, ``No retraining, no re-measurement. Only the verdict predicate changed'' (Equation~\ref{eq:parity-relaxed}).
The results document's own summary: ``A reader who prefers the original uniform bar should record NVFP4 W4A16 parity as failing on retention rank and passing on everything else.''
Neither failing log is among the staged inputs; the history rests on the provenance log, the results document and the code comment.

Under NVFP4 the expert matrix is kept in fp32, but QAT retrains it; the expert-level values of the fp4 export sit within about 0.1 of fp32.

\subsection{Cross-language runtime parity}
\label{sec:results-runtime}

Table~\ref{tab:runtime-parity} records the parity status of the four runtimes after the second phase.
At the first snapshot the picture was uneven: only the C11 runtime had been re-run on the current bundles, and Java and Rust were ``expected to pass; that is an expectation, not a measurement''.
The second phase measured it.
On every benchmark machine all four runtimes passed the eight golden cases and the 2,000-case golden set on both bundles.
On the 2,000-case set the planner assigns 68 households for a budget of 5,000 on fp32 (spent 4,975.96) and 61 on fp4 (spent 4,995.15), identically in C, Java, Rust and Python and on every machine; on the eight-case set it assigns seven (841.99 and 839.34), as before.

\begin{table}[htbp]
  \centering
  \caption{Runtime parity on the golden sets after the second phase (eight and 2,000 cases; tolerance $10^{-4}$, outcomes $10^{-3}$; cells give fp32 / fp4, OK = parity passed) on every machine of Table~\ref{tab:bench-machines}.}
  \label{tab:runtime-parity}
  \footnotesize
  \begin{tabular}{@{}l l c c@{}}
    \toprule
    Runtime & Dependencies & golden-8 & golden-2000 \\
    \midrule
    Python & numpy & OK / OK & OK / OK \\
    Java & none (JDK 17+) & OK / OK & OK / OK \\
    C11 & libc + libm & OK / OK & OK / OK \\
    Rust & none (standard library) & OK / OK & OK / OK \\
    \bottomrule
  \end{tabular}
\end{table}

The 2,000-case set is what made the exercise worth doing: it is the set on which the FMA contraction finding of Section~\ref{sec:evaluation-bench-gate} surfaced.
The website's ``microseconds'' claim is not repeated here; the harness measures p50 and p99 latency per machine and language, and those measurements are what this paper reports.

\subsection{Regression of the moral core across model variants}
\label{sec:results-cregr}

We re-ran the same C-checks on the v2 fp32 and v2 NVFP4-QAT model directories; both passed eight of eight.
Table~\ref{tab:c-regr} sets the three runs side by side.

\begin{table}[htbp]
  \centering
  \caption{C-checks across model variants. Statistics that would reveal a planted effect are reduced to the verdict.}
  \label{tab:c-regr}
  \footnotesize
  \begin{tabular}{p{0.30\textwidth}lll}
    \toprule
    Check & v1 & v2 fp32 & v2 NVFP4-QAT \\
    \midrule
    C1 $\rho$ / cosine / accuracy & 1.000 / 0.997 / 0.706 & 1.000 / 0.997 / 0.706 & 1.000 / 0.997 / 0.702 \\
    C2, C3 cultural gradients & pass / pass & pass / pass & pass / pass \\
    C4 group effect & pass & pass & pass \\
    C5 in-group effect localised in one expert & pass & pass & pass \\
    C6, model-independent probe & pass & pass & pass \\
    C9s purity / ARI & 0.835 / 0.671 & 0.835 / 0.670 & 0.799 / 0.613 \\
    C10 completion & pass & pass & pass \\
    \bottomrule
  \end{tabular}
\end{table}

In the v2 run the per-cohort Spearman values lie between 0.893 and 1.000.
C6 is identical across variants because it is computed from the model-independent v1 logit probes.
QAT costs segment recovery about 0.036 in purity and 0.058 in ARI.

Going from seven of eight to eight of eight on v2 involved no retraining.
The first v2 evaluation reported seven of eight, with the C-regression failing because the logit probe that C6 needs was not yet in the model directory and C6 could not run; the re-run with the probes copied in gave eight of eight, ``a missing-input fix, not a threshold change''.

\subsection{Cross-platform performance}
\label{sec:results-bench}

The speed matrix of Section~\ref{sec:evaluation-bench} ran on 30 August 2026 on the v2 bundles.
The DGX Spark produced 352 results (plus the ten Java out-of-memory cells re-run with a 4~GiB heap), the two 2~GiB instances 280 results each with no failures, and the M4 Pro 632 cells with no failures.
The two 8~GiB instances ran only the 100,000- and 400,000-row sets, so the synth-10k tables of this section draw on the other four machines; the large instances' cells enter the checksum and parity totals and the 400,000-row comparison below.
The environment and the raw tables are in Appendix~\ref{app:bench}.

\subsubsection{Parity gate and cross-language agreement}

The gate came first.
Before any speed cell ran, all four runtimes passed both golden sets on both bundles on all six machines.
The cheap check that travels with every timed cell agrees as well: the SHA-256 over the formatted outputs of every timed row is identical across the languages present in 422 of 422 (machine, dataset, bundle, mode, threads) cells with results from at least two languages, 211 fp32 and 211 fp4 cells over 1,664 result files.
Four cells disagreed in the first pass, three on the DGX and one on the Mac, and none of the four was rounding.
Their Python driver had been rebuilt mid-matrix from the newer source tree and defaulted to population-relative percentiles; the DGX cells were re-run with the grid reference, and the Mac matrix was repeated in full (Appendix~\ref{app:bench}).

\subsubsection{Parity from the dumps}

\begin{figure}[htbp]
  \centering
  \includegraphics[width=\textwidth]{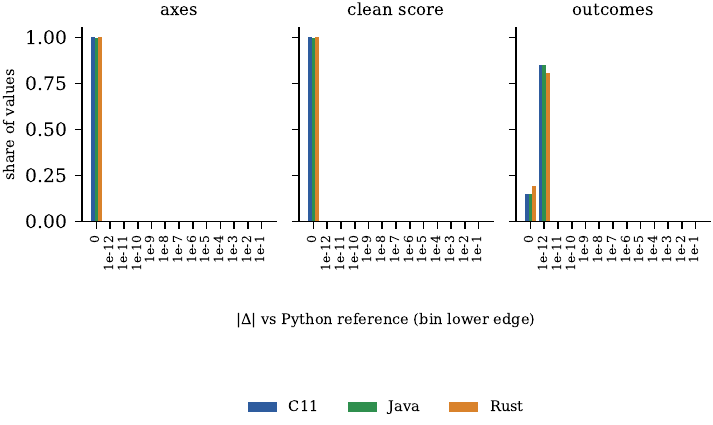}
  \caption{Distribution of $|\Delta|$ against the Python reference by language for the criteria, clean-score and outcome families, pooled over the six machines; the left-most bin is exact agreement, the first non-zero bin collects every non-zero difference below $10^{-11}$, and the bins above it are labelled by their lower edge.}
  \label{fig:bench-parity-error}
\end{figure}

Table~\ref{tab:bench-parity} and Figure~\ref{fig:bench-parity-error} give the tolerance test of Section~\ref{sec:evaluation-bench-parity}.
The first 1,000 rows dumped by every single-thread C, Java and Rust \emph{axes} and \emph{menu} invocation are compared with the Python driver's dump of the same cell, value by value, in six families.
Over the six machines that is 2,226 (cell, family) comparisons, 2,091,000 row comparisons and 87.9 million values.

\begin{table}[htbp]
  \setstretch{1}
  \centering
  \footnotesize
  \caption{Cross-language parity from the dumps (Equation~\ref{eq:parity-bound}): worst-case absolute error $e^{\max}$ and pass rate per language and quantity family against the Python package, pooled over the 2,226 comparisons of the six machines (largest error, smallest pass rate). Tolerances $10^{-4}$ for derived fields, criteria (``axes''), clean score and cost, $10^{-3}$ for outcomes and welfare.}
  \label{tab:bench-parity}
  \begin{tabular}{@{}lrrrrrr@{}}
\toprule
Quantity & \multicolumn{2}{c}{C11} & \multicolumn{2}{c}{Rust} & \multicolumn{2}{c}{Java} \\
\cmidrule(lr){2-3}\cmidrule(lr){4-5}\cmidrule(lr){6-7}
 & $e^{\max}$ & pass rate & $e^{\max}$ & pass rate & $e^{\max}$ & pass rate \\
\midrule
derived & 0.0e+00 & 1.0000 & 0.0e+00 & 1.0000 & 5.7e-14 & 1.0000 \\
axes & 0.0e+00 & 1.0000 & 0.0e+00 & 1.0000 & 1.1e-16 & 1.0000 \\
clean score & 0.0e+00 & 1.0000 & 0.0e+00 & 1.0000 & 8.9e-16 & 1.0000 \\
cost total & 0.0e+00 & 1.0000 & 0.0e+00 & 1.0000 & 1.1e-12 & 1.0000 \\
welfare & 1.3e-15 & 1.0000 & 1.3e-15 & 1.0000 & 1.3e-15 & 1.0000 \\
outcomes & 7.1e-14 & 1.0000 & 7.1e-14 & 1.0000 & 7.1e-14 & 1.0000 \\
\bottomrule
\end{tabular}

\end{table}

The pass rate is 1.000 in every comparison, and the largest error anywhere is $1.137 \times 10^{-12}$.
C and Rust are bit-identical to Python on derived fields, criteria, clean scores and plan costs; their outcomes differ by at most $7.1 \times 10^{-14}$, which is summation order inside the MLP, not precision.
Java is not bit-identical but sits eight to twelve orders of magnitude below $\tau$ (the JVM's mathematical library and the JIT's own contraction rules), so the bound of Equation~\ref{eq:parity-bound} holds with a margin of about $10^{8}$.
The ``float32 weights read as double'' residual anticipated in Section~\ref{sec:evaluation-bench-parity} does not appear between the runtimes at all, because every one of them widens the same float32 tensors to binary64 before computing; it exists only against the golden export.

\subsubsection{Latency and throughput}

\begin{figure}[htbp]
  \centering
  \includegraphics[width=\textwidth]{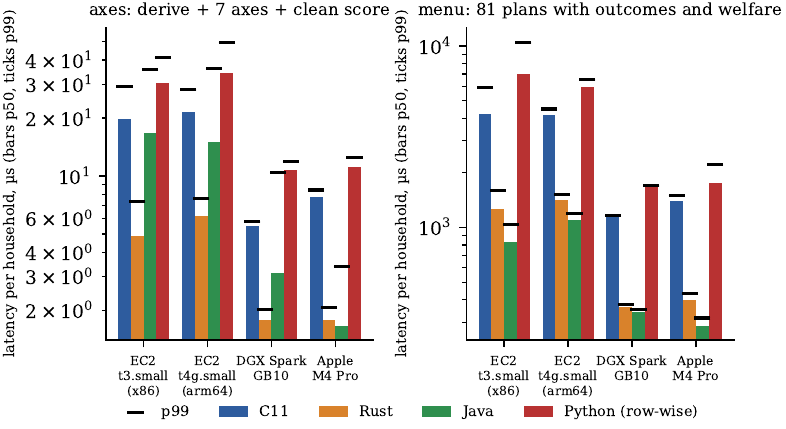}
  \caption{Per-household latency by runtime and machine on synth-10k (fp32, one thread): bars give p50 and ticks p99, in microseconds, for \emph{axes} (derived economics, seven criteria and clean score) and \emph{menu} (81 plans with outcomes, cost and welfare).}
  \label{fig:bench-latency}
\end{figure}

Table~\ref{tab:bench-latency} and Figure~\ref{fig:bench-latency} give per-row latency (Equation~\ref{eq:latency-pct}) and throughput (Equation~\ref{eq:throughput}) on synth-10k, the dataset common to the Mac, the DGX and the two 2~GiB instances, with the fp32 bundle and one thread.

\begin{table}[htbp]
  \setstretch{1}
  \centering
  \footnotesize
  \caption{Per-row latency (p50 and p99 in microseconds, nearest rank on the median pass) and throughput (rows/s) of the four runtimes in \emph{axes} and \emph{menu} mode on synth-10k, fp32 bundle, one thread. The two 8~GiB instances ran only the 100k and 400k cells and are absent.}
  \label{tab:bench-latency}
  \setlength{\tabcolsep}{3pt}
  \resizebox{\textwidth}{!}{%
  \begin{tabular}{@{}llrrrrrrrrrrrr@{}}
\toprule
Mode & Runtime & \multicolumn{3}{c}{EC2 t3.small} & \multicolumn{3}{c}{EC2 t4g.small} & \multicolumn{3}{c}{DGX Spark GB10} & \multicolumn{3}{c}{Apple M4 Pro} \\
\cmidrule(lr){3-5}\cmidrule(lr){6-8}\cmidrule(lr){9-11}\cmidrule(lr){12-14}
 & & p50 & p99 & rows/s & p50 & p99 & rows/s & p50 & p99 & rows/s & p50 & p99 & rows/s \\
\midrule
axes & C11 & 19.7 & 29.1 & 49,716 & 21.4 & 28.2 & 45,723 & 5.5 & 5.8 & 180,287 & 7.8 & 8.5 & 127,624 \\
 & Rust & 4.9 & 7.4 & 197,426 & 6.2 & 7.7 & 156,224 & 1.8 & 2.0 & 547,388 & 1.8 & 2.1 & 544,507 \\
 & Java & 16.6 & 35.6 & 54,090 & 15.0 & 36.0 & 60,964 & 3.1 & 10.5 & 260,808 & 1.7 & 3.4 & 525,197 \\
 & Python & 30.4 & 41.3 & 31,912 & 34.4 & 49.5 & 27,996 & 10.8 & 11.9 & 91,572 & 11.1 & 12.5 & 88,657 \\
\midrule
menu & C11 & 4,193.8 & 5,914.7 & 352 & 4,161.4 & 4,489.6 & 366 & 1,149.4 & 1,163.7 & 1,343 & 1,398.4 & 1,503.8 & 1,098 \\
 & Rust & 1,262.7 & 1,592.3 & 1,158 & 1,410.8 & 1,527.8 & 1,077 & 367.0 & 375.5 & 4,182 & 398.5 & 432.4 & 3,822 \\
 & Java & 828.6 & 1,038.9 & 1,801 & 1,099.2 & 1,190.8 & 1,378 & 342.3 & 354.9 & 4,421 & 287.4 & 317.5 & 5,272 \\
 & Python & 6,993.5 & 10,460.9 & 209 & 5,944.8 & 6,536.3 & 255 & 1,690.6 & 1,704.9 & 911 & 1,752.8 & 2,214.6 & 845 \\
\bottomrule
\end{tabular}
}
\end{table}

Read per row, encoding one household on the seven criteria costs 1.7 to 1.8~$\mu$s in Rust on both workstation-class machines, 5.5 to 7.8~$\mu$s in C and 11~$\mu$s in row-wise Python; on the 2-vCPU instances C, Rust and Python are 2.5 to 3.9 times slower.
The 81-plan menu takes 0.29 to 0.40~ms in Java and Rust, 1.1 to 1.4~ms in C and 1.7 to 1.8~ms in Python on the workstations, and 0.8 to 7.0~ms on the instances.
The ordering of the runtimes is the same on every machine (on \emph{menu} Java $>$ Rust $>$ C $>$ Python); C is the slowest compiled runtime there because it rebuilds a JSON plan object per plan, an implementation choice rather than a language ceiling.
Latency does not depend on the dataset: on the Mac, Rust \emph{axes} p50 is 1.8 to 2.0~$\mu$s on the 16 files, from 500 to 400,000 rows, real and synthetic, with one value above 2.0 (2.25~$\mu$s on Cox's Bazar with quality fields).
Mode \emph{plan}, one single-threaded planning call over synth-100k, takes 29.7 to 37.6~s in Java and Rust, 77.7 to 101.0~s in C and 495.3 to 838.9~s in Python; the planner is $O(n \log n)$ after an $O(81\,n)$ menu pass, so Python's minutes are the price of the reference implementation, not a limit of the algorithm.

\subsubsection{Cross-machine ratios and normalisation}

Relative to the t3.small vCPU the GB10 core is 2.8 to 4.8 times faster and the M4 Pro core 2.6 to 9.7 times on \emph{axes}; the spread across languages on one machine is larger than the spread across machines for one language, which is why the choice of normaliser matters less than the choice of runtime.

\begin{figure}[htbp]
  \centering
  \includegraphics[width=\textwidth]{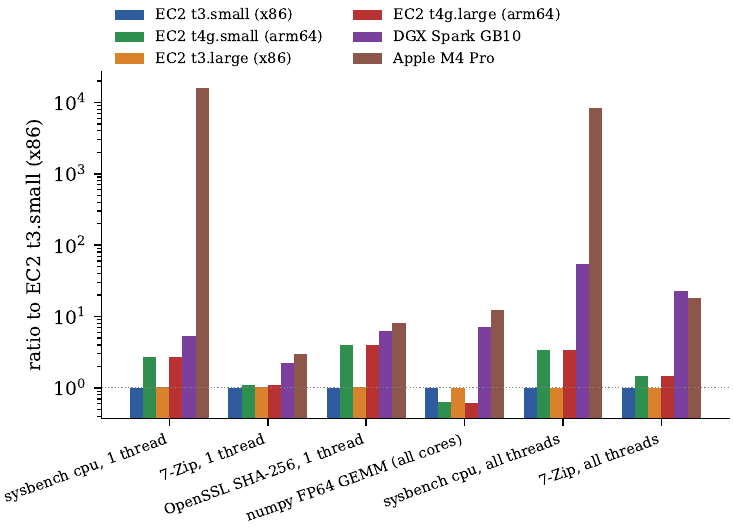}
  \caption{Reference benchmarks of each machine as ratios to the t3.small (log scale): sysbench cpu with one thread and with all threads, 7-Zip with one thread and with all threads, OpenSSL SHA-256 with one thread and numpy FP64 GEMM. The M4 Pro's sysbench cpu bars are not comparable (the Homebrew build optimises the prime loop away; Appendix~\ref{app:bench}).}
  \label{fig:bench-universal-ratio}
\end{figure}

Figure~\ref{fig:bench-universal-ratio} plots the reference benchmarks as ratios to the t3.small.
Normalised by the single-thread 7-Zip rating (Equation~\ref{eq:norm-throughput}), \emph{menu} runs at 0.39 to 0.61 rows per second per MIPS in Java and 0.31 to 0.58 in Rust across the four machines; the GB10 is the most efficient machine per MIPS on every runtime and the Graviton2 the least.
On this evidence 7-Zip predicts the cross-machine ratio to within roughly $\pm 30$ per cent; GEMM does not, because rows per GFLOPS range over five times for the same runtime.

\subsubsection{Thread scaling}

\begin{figure}[htbp]
  \centering
  \includegraphics[width=\textwidth]{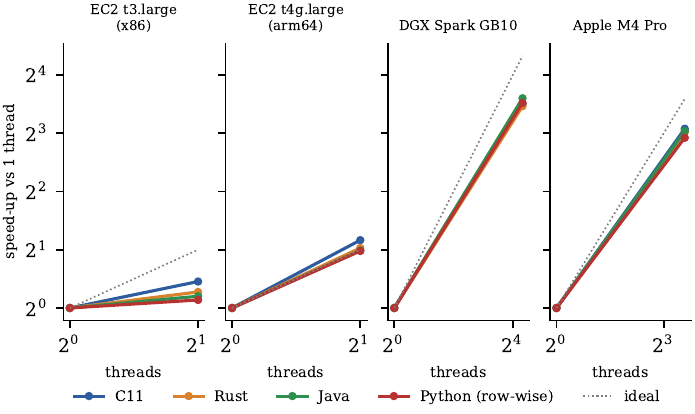}
  \caption{Speed-up of \emph{menu} over one thread against thread count on synth-100k (fp32), one panel per machine, log$_2$ axes; the dotted line is ideal scaling.}
  \label{fig:bench-throughput-threads}
\end{figure}

Table~\ref{tab:bench-scaling} and Figure~\ref{fig:bench-throughput-threads} give $\Theta(T)/\Theta(1)$ and $E(T)$ (Equation~\ref{eq:scaling-eff}) for \emph{menu} with the fp32 bundle and every logical CPU of the machine on synth-100k; the 2~GiB instances did not run that dataset, and their two-thread figures are given in the text.

\begin{table}[htbp]
  \setstretch{1}
  \centering
  \footnotesize
  \caption{Thread scaling of \emph{menu} on synth-100k (fp32): speed-up over one thread with all logical CPUs, efficiency $E(T)$ of Equation~\ref{eq:scaling-eff} and the cores kept busy.}
  \label{tab:bench-scaling}
  \begin{tabular}{@{}llrrrr@{}}
\toprule
Machine & Runtime & threads $T$ & speed-up $\Theta(T)/\Theta(1)$ & $E(T)$ & cores busy \\
\midrule
EC2 t3.large & C11 & 2 & 1.37 & 0.69 & 2.3 \\
EC2 t3.large & Rust & 2 & 1.21 & 0.60 & 2.3 \\
EC2 t3.large & Java & 2 & 1.15 & 0.57 & 2.4 \\
EC2 t3.large & Python & 2 & 1.10 & 0.55 & 2.5 \\
EC2 t4g.large & C11 & 2 & 2.24 & 1.12 & 2.5 \\
EC2 t4g.large & Rust & 2 & 2.04 & 1.02 & 2.5 \\
EC2 t4g.large & Java & 2 & 1.98 & 0.99 & 2.6 \\
EC2 t4g.large & Python & 2 & 1.97 & 0.99 & 2.5 \\
DGX Spark GB10 & C11 & 20 & 11.31 & 0.56 & 19.6 \\
DGX Spark GB10 & Rust & 20 & 11.01 & 0.55 & 20.0 \\
DGX Spark GB10 & Java & 20 & 12.10 & 0.60 & 19.9 \\
DGX Spark GB10 & Python & 20 & 11.45 & 0.57 & 18.6 \\
Apple M4 Pro & C11 & 12 & 8.42 & 0.70 & 13.8 \\
Apple M4 Pro & Rust & 12 & 8.08 & 0.67 & 13.6 \\
Apple M4 Pro & Java & 12 & 8.19 & 0.68 & 13.3 \\
Apple M4 Pro & Python & 12 & 7.57 & 0.63 & 14.1 \\
\bottomrule
\end{tabular}

\end{table}

On the DGX with 20 threads the speed-up is 11.0 to 12.1 times on synth-100k, $E(20)$ from 0.55 to 0.64; the GB10's ten efficiency cores are roughly half a performance core each, so about 0.6 is the ceiling of that heterogeneous cluster rather than a loss to synchronisation.
On the Mac with 12 threads the compiled runtimes reach 7.6 to 8.4 times; the t4g.small's two cores give 1.83 to 1.99 times on synth-10k and the t3.small's two hyper-threads of one core 1.06 to 1.17.

Java needs heap in proportion to the population: with a 4~GiB heap it scales like the others up to 100,000 households, but at 400,000 the heap is full and the serial collector dominates, and with a 16~GiB heap it returns to full speed (Appendix~\ref{app:bench}).
At 400,000 rows the single-thread order on \emph{menu} is Java (16~GiB) at 4.2 to 5.3 thousand rows/s, Rust 3.5 to 3.8 thousand, C 1.1 to 1.2 thousand and Python 0.86 to 0.92 thousand on both workstation-class machines, and every runtime's per-row time is the same at 400k as at 10k.

\subsubsection{fp32 against fp4}

The fp4 bundle carries the QAT weights de-quantised to fp32 tensors of the same shapes, so no runtime has a special path and none should be faster.
None is.
Over the 832 cells on the six machines that exist for both bundles, the throughput ratio $\Theta_{\mathrm{fp4}}/\Theta_{\mathrm{fp32}}$ has a median of 1.000 and an inter-quartile range of 0.985 to 1.016; the extremes (0.45 to 2.0) are all multi-thread cells on small files, where thread start-up and JIT state dominate.
Any claim of fp4 acceleration would need packed NVFP4 kernels, which no CPU runtime has, and the GPU path below has none either.
The fp4 checksums are identical across the languages present in all 211 fp4 cells, and the dump parity of the fp4 cells is the same 1.000 with $e^{\max} \leq 1.1 \times 10^{-12}$.

\subsubsection{GPU}

\begin{figure}[htbp]
  \centering
  \includegraphics[width=0.8\textwidth]{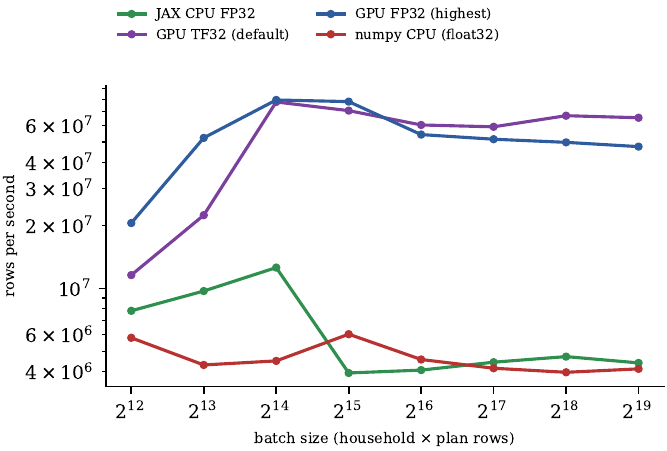}
  \caption{Outcome-head throughput on the DGX Spark against batch size (log$_2$; rows are (household, plan) pairs): JAX on the GB10 GPU in FP32 and TF32, JAX on the 20 CPU cores, and the numpy reference; median of three timed passes. TF32 fails parity and is shown for speed only.}
  \label{fig:bench-gpu-batch}
\end{figure}

Figure~\ref{fig:bench-gpu-batch} gives the one GPU measurement: the outcome head batched over 624,078 (household, plan) rows, 12,000 held-out households times their admissible plans, in JAX 0.5.3 on the GB10.

FP32 peaks at 79.4 million rows/s at a batch of 16,384, against 3.9 to 12.6 million for JAX on the 20 CPU cores and 4.0 to 6.0 million for numpy; that is about 1.5 million households per second for the outcome head alone.
But the head is one of three stages, and neither the encoder nor the planner is GPU-resident, so the end-to-end \emph{menu} figure of Table~\ref{tab:bench-latency} (4,182 rows/s in Rust on the same machine) is the operational number.
Parity settles which precision may be used: FP32 outputs are bit-identical to numpy on 100,000 rows, whereas TF32 puts only 51.4 per cent of rows within the $10^{-3}$ tolerance and must not be used for AllotNet's head.

\subsection{Behaviour on household datasets}
\label{sec:results-realdata}

The criteria of Section~\ref{sec:evaluation-bench-quality} were computed on the six datasets of Table~\ref{tab:bench-datasets} with $k = 15$ per cent and $B = 100$ bootstrap re-draws, on the v2.1 bundles exported on 30 August 2026.
Three settings distinguish this run from the v2 run it replaces.
Percentile-based encoder terms are ranked against the scored population rather than the training grid.
The encoder's terms carry the data-quality weights of the transformation step.
And the eligibility gate is evaluated as two extra methods in the cost table, at quotas of 0.15 and 0.30.
Before any of these numbers was quoted, the v2.1 bundles passed the parity gate on the DGX Spark and on the Apple M4 Pro, and the four cloud instances passed the same gate after their speed matrices, so all six machines gated the v2.1 bundles; the household-data numbers were computed on the DGX.
Two of the six datasets are controls: synth-10k is the world the model was trained on, and the DHIS2 harness inherits its joint distribution, so both show what agreement looks like when the encoder's assumptions hold.

\subsubsection{Rank agreement with standard indicators}

\begin{figure}[htbp]
  \centering
  \includegraphics[width=\textwidth]{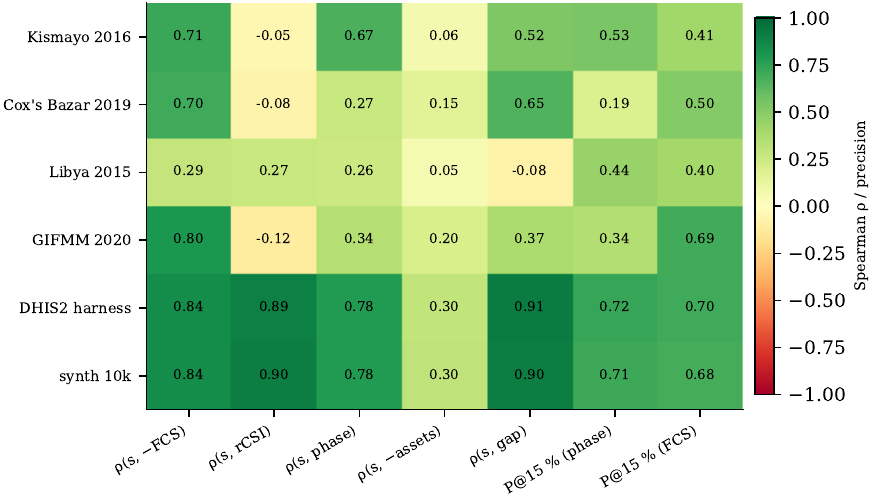}
  \caption{Rank agreement between the clean score and the standard indicators, datasets by indicators. The two synthetic-distribution sets (DHIS2 harness, synth-10k) sit at or near the top of every column; of the surveys, GIFMM comes closest on FCS (0.80 against 0.84) and edges them on P@15 per cent against FCS (0.69 against 0.68 to 0.70).}
  \label{fig:bench-rank-agreement}
\end{figure}

Table~\ref{tab:bench-quality} and Figure~\ref{fig:bench-rank-agreement} give the rank agreement between the clean score and the indicators.

\begin{table}[htbp]
  \setstretch{1}
  \centering
  \footnotesize
  \caption{Clean score (v2.1, population percentiles, data-quality weights) against standard indicators on household data: Spearman $\rho_s$ with FCS (sign reversed), rCSI, the CARI-style phase and the monthly gap, precision at $k = 15$ per cent against FCS and against the phase, monotonicity under perturbation ($M_{\mathrm{fcs}}$, $M_{\mathrm{gap}}$) and fp32 against fp4 agreement ($\rho_s$, top-decile overlap).}
  \label{tab:bench-quality}
  \setlength{\tabcolsep}{3pt}
  \resizebox{\textwidth}{!}{%
  \begin{tabular}{@{}lrrrrrrrrr@{}}
\toprule
Dataset & $n$ & $\rho_s(s,-\mathrm{FCS})$ & $\rho_s(s,\mathrm{rCSI})$ & $\rho_s(s,\text{phase})$ & $\rho_s(s,\text{gap})$ & P@15\% FCS & P@15\% phase & $M_{\mathrm{fcs}}$ / $M_{\mathrm{gap}}$ & $\rho_s(s^{32},s^{4})$ / $O_{10}$ \\
\midrule
Kismayo 2016 & 1,154 & 0.71 & -0.05 & 0.67 & 0.52 & 0.41 & 0.53 & 1.000 / 1.000 & 1.0000 / 1.000 \\
Cox's Bazar 2019 & 3,385 & 0.70 & -0.08 & 0.27 & 0.65 & 0.50 & 0.19 & 1.000 / 1.000 & 0.9999 / 0.985 \\
Libya 2015 & 500 & 0.29 & 0.27 & 0.26 & -0.08 & 0.40 & 0.44 & 1.000 / 1.000 & 1.0000 / 1.000 \\
GIFMM 2020 & 3,101 & 0.80 & -0.12 & 0.34 & 0.37 & 0.69 & 0.34 & 1.000 / 1.000 & 1.0000 / 0.997 \\
DHIS2 harness & 19,029 & 0.84 & 0.89 & 0.78 & 0.91 & 0.70 & 0.72 & 1.000 / 1.000 & 1.0000 / 0.994 \\
synth 10k & 10,000 & 0.84 & 0.90 & 0.78 & 0.90 & 0.68 & 0.71 & 1.000 / 1.000 & 1.0000 / 0.993 \\
\bottomrule
\end{tabular}
}
\end{table}

On the synthetic world, and on its DHIS2-shaped copy, the clean score tracks FCS and rCSI at $\rho_s$ of 0.84 to 0.90, as it must: the FCS deficit, rCSI and the gap ratio are encoder terms and the gap drives the deprivation index.
On the three surveys with a directly observed FCS and a proxy income the score now follows FCS at $\rho_s$ of 0.70 to 0.80 (Kismayo 0.71, Cox's Bazar 0.70, GIFMM 0.80; with v2, 0.60, 0.41 and 0.49).
The quality weights remove the hot-decked rCSI from the score and halve the weight of the expenditure- or wage-based income, so the one directly measured food-security indicator carries most of the score's food-security signal.
Precision at $k = 15$ per cent against FCS rises accordingly, to 0.41, 0.50 and 0.69 (v2: 0.38, 0.32 and 0.29).
Kendall $\tau_b$ against FCS is 0.51, 0.51 and 0.61 on those three files, 0.21 on Libya and 0.64 on the two synthetic sets.

The three negative rCSI correlations are informative, not bugs: in Cox's Bazar and GIFMM rCSI is entirely imputed (weight 0 in the score), and in Kismayo the direct five-item rCSI is only weakly related to FCS and to the gap.
For the same reason the CARI-style phase reference is contaminated by donor noise on the two files with imputed rCSI, and the FCS-only columns are the fair test there.
Libya is the counter-example: its income is wholly constructed, so the gap term leaves the score ($\rho_s(s, \mathrm{gap}) = -0.08$), and its FCS, weakly related to every other indicator in that file of 500 households, correlates only 0.29 with the score before and after the change.
Assets are close to uncorrelated with the score, by design: the asset score enters only the outcome head.

\subsubsection{Targeting errors against reference poor sets}

Table~\ref{tab:bench-targeting} reports the two targeting errors of Equation~\ref{eq:ee-ie} against two reference poor sets, CARI-style phase 3 or 4 and FCS below 35, once with the target set at $k = 15$ per cent and once at $|T| = |P|$, where exclusion and inclusion coincide.

\begin{table}[htbp]
  \setstretch{1}
  \centering
  \footnotesize
  \caption{Exclusion (EE) and inclusion (IE) errors of the v2.1 clean-score ranking against two indicator-defined poor sets $P$: at a target set of $k = 15$ per cent of households, and at $|T| = |P|$, where EE = IE.}
  \label{tab:bench-targeting}
  \resizebox{\textwidth}{!}{%
  \begin{tabular}{@{}l r l r r l r@{}}
    \toprule
    & \multicolumn{3}{c}{Phase $\geq 3$} & \multicolumn{3}{c}{FCS $< 35$} \\
    \cmidrule(lr){2-4} \cmidrule(l){5-7}
    Dataset & $|P|$ & EE / IE at $k$ & EE = IE & $|P|$ & EE / IE at $k$ & EE = IE \\
    \midrule
    Kismayo 2016 & 373 & 0.622 / 0.185 & 0.346 & 641 & 0.746 / 0.058 & 0.182 \\
    Cox's Bazar 2019 & 105 & 0.429 / 0.882 & 0.752 & 538 & 0.507 / 0.477 & 0.485 \\
    Libya 2015 & 10 & 0.600 / 0.947 & 0.800 & 19 & 0.474 / 0.867 & 0.632 \\
    GIFMM 2020 & 160 & 0.100 / 0.690 & 0.500 & 418 & 0.273 / 0.346 & 0.313 \\
    DHIS2 harness & 2,086 & 0.213 / 0.425 & 0.357 & 2,212 & 0.245 / 0.415 & 0.359 \\
    synth-10k & 1,079 & 0.222 / 0.440 & 0.365 & 1,153 & 0.256 / 0.428 & 0.373 \\
    \bottomrule
  \end{tabular}}
\end{table}

Against the FCS-below-35 reference with equal-size sets the clean score now misses 18 per cent of the indicator-poor on Kismayo, 31 per cent on GIFMM and 49 per cent on Cox's Bazar, against 24, 73 and 67 per cent with v2 and 36 to 37 per cent on the synthetic world.
GIFMM's phase-3 set is the one case where the score is a better poverty targeter than on synthetic data (EE 0.10 at $k$).
Cox's Bazar remains the hardest file: 60 per cent of its households have a gap, the income is an expenditure band, and the phase reference is rCSI-contaminated.
Libya's reference sets are 10 and 19 households; its errors are noise.

\subsubsection{Monotonicity, fp32 against fp4 and imputation sensitivity}

Both monotonicity checks of Equation~\ref{eq:monotonicity} return exactly 1.000 on all six datasets: adding five points of FCS never raises the score and removing USD~5 of monthly income never lowers it.
The quality scaling preserves the encoder's monotonicity because it rescales weights, not terms.
The fp4 bundle reproduces the fp32 ranking: $\rho_s(s^{32}, s^{4}) = 1.000$ on every dataset, a top-decile overlap $O_{10}$ of 1.000 on Kismayo and Libya, 0.997 on GIFMM, 0.994 on the DHIS2 harness, 0.993 on synth-10k and 0.985 on Cox's Bazar, and a maximum absolute score difference of 0.030 to 0.037, all inside the acceptance rule.
The clean score does not touch the retrained head, so the fp32 against fp4 difference is the moral core's QAT alone.

The imputation bootstrap (Table~\ref{tab:bench-bootstrap}) is now exactly stable, and that is the intended consequence of the quality weights, not a triumph.

\begin{table}[htbp]
  \setstretch{1}
  \centering
  \footnotesize
  \caption{Imputation bootstrap: $B = 100$ seeded re-draws of the hot-decked fields, top-15 per cent target set. Unstable means a membership probability strictly between 0.1 and 0.9; the Jaccard index is against the reference draw, with a percentile 95 per cent interval. The v2 columns are given for contrast.}
  \label{tab:bench-bootstrap}
  \setlength{\tabcolsep}{4pt}
  \begin{tabular}{@{}l l r l r r@{}}
    \toprule
    & & \multicolumn{2}{c}{v2.1} & \multicolumn{2}{c}{v2} \\
    \cmidrule(lr){3-4} \cmidrule(l){5-6}
    Dataset & Imputed fields (rows) & Unstable & Jaccard, mean [2.5\%, 97.5\%] & Unstable & Jaccard \\
    \midrule
    Cox's Bazar 2019 & rCSI (3,385) & 0.000 & 1.000 [1.000, 1.000] & 0.221 & 0.566 \\
    Libya 2015 & FCS, rCSI (176) & 0.000 & 1.000 [1.000, 1.000] & 0.108 & 0.721 \\
    GIFMM 2020 & rCSI (3,101) & 0.000 & 1.000 [1.000, 1.000] & 0.128 & 0.716 \\
    \bottomrule
  \end{tabular}
\end{table}

A field with quality 0 has weight 0, so re-drawing it cannot move the score.
The v2 instability (13 to 22 per cent of households unstable, Jaccard 0.57 to 0.72) measured how much the donor distribution had been deciding; the v2.1 score decides on the observed fields only.
Selection on Cox's Bazar and GIFMM is therefore no longer selection on donors.
It is selection on fewer indicators, which Table~\ref{tab:bench-targeting} shows.

\subsubsection{Planner efficiency against baselines}

\begin{figure}[htbp]
  \centering
  \includegraphics[width=\textwidth]{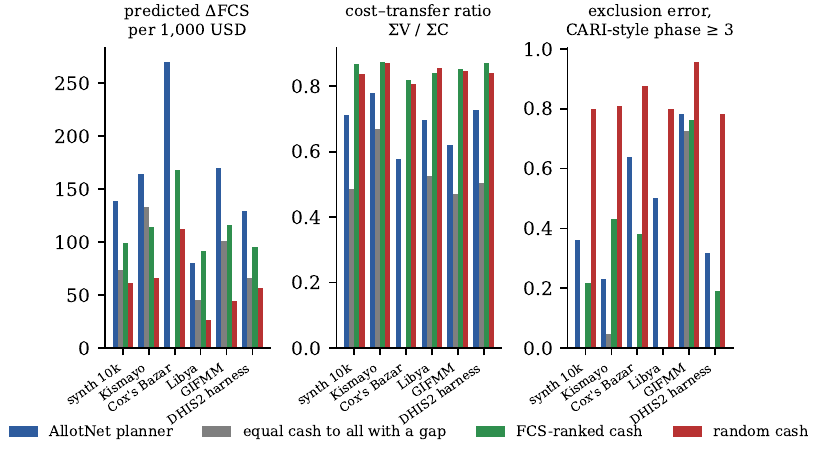}
  \caption{Predicted $\Delta$FCS per thousand USD, cost--transfer ratio and exclusion error against phase $\geq 3$ for the planner and the three baselines at $f = 0.25$, by dataset (v2.1).}
  \label{fig:bench-efficiency}
\end{figure}

\begin{figure}[htbp]
  \centering
  \includegraphics[width=0.8\textwidth]{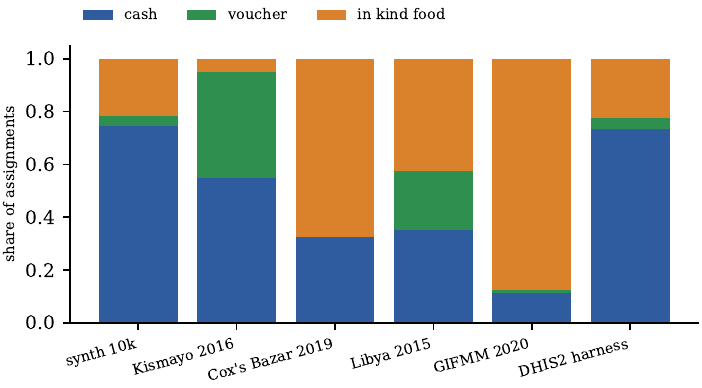}
  \caption{Modality shares of the planner's assignments at $f = 0.25$, by dataset (v2.1 head). Kismayo, Libya and GIFMM have market access fixed at `limited' because their surveys carry no market item; the in-kind share there answers a constant, not an observed market.}
  \label{fig:bench-planner-mix}
\end{figure}

Table~\ref{tab:bench-efficiency} compares the planner, the gated planner and the three cash baselines at a budget of a quarter of the population's three-month gap; Figure~\ref{fig:bench-efficiency} plots three of the columns and Figure~\ref{fig:bench-planner-mix} the modality mix.
Every outcome column is the v2.1 head's prediction, and every method is inside the head's training support, since the head was trained on coverages 0.1 to 1.2; CTR, $C_{hm}$, $\alpha$, GC, timeliness and the targeting errors are model-free.

\begin{table}[htbp]
  \setstretch{1}
  \centering
  \footnotesize
  \caption{Planner, gated planner (quotas 0.15 and 0.30) and three cash baselines at the same budget ($f = 0.25$ of the population's three-month gap), v2.1 head: households assigned, budget utilisation $U$, cost--transfer ratio, cost per household-month, alpha value of the in-kind share, predicted $\Delta$FCS and welfare per thousand USD, population gap coverage and exclusion error against phase $\geq 3$. Equal cash in Cox's Bazar assigns nothing because the per-household package falls below the USD~5 monthly floor.}
  \label{tab:bench-efficiency}
  \setlength{\tabcolsep}{4pt}
  \resizebox{\textwidth}{!}{%
  \begin{tabular}{@{}llrrrrrrrrr@{}}
\toprule
Dataset & Method & assigned & $U$ & CTR & $C_{hm}$ (USD) & $\alpha_{\text{in-kind}}$ & $\Delta$FCS/k\$ & $W$/k\$ & GC & EE$_{\text{phase}\ge3}$ \\
\midrule
synth 10k & AllotNet planner & 2,688 & 1.000 & 0.711 & 37.2 & 0.63 & 138.2 & 2.10 & 0.178 & 0.36 \\
 & planner, gate 15\,\% & 1,500 & 0.959 & 0.764 & 63.9 & 0.67 & 116.7 & 1.56 & 0.183 & 0.22 \\
 & planner, gate 30\,\% & 2,006 & 1.000 & 0.747 & 49.8 & 0.65 & 134.6 & 1.90 & 0.187 & 0.24 \\
 & equal cash & 6,781 & 1.000 & 0.486 & 14.7 & -- & 73.8 & 1.35 & 0.121 & 0.00 \\
 & FCS-ranked cash & 887 & 1.000 & 0.868 & 112.7 & -- & 98.9 & 1.29 & 0.217 & 0.22 \\
 & random cash & 1,372 & 1.000 & 0.837 & 72.8 & -- & 61.0 & 1.09 & 0.209 & 0.80 \\
\midrule
Kismayo 2016 & AllotNet planner & 479 & 1.000 & 0.781 & 55.6 & 0.63 & 163.9 & 3.09 & 0.195 & 0.23 \\
 & planner, gate 15\,\% & 173 & 0.533 & 0.793 & 82.0 & 0.68 & 124.5 & 2.54 & 0.106 & 0.62 \\
 & planner, gate 30\,\% & 345 & 0.991 & 0.801 & 76.5 & 0.68 & 124.6 & 2.55 & 0.198 & 0.38 \\
 & equal cash & 1,049 & 1.000 & 0.671 & 25.4 & -- & 132.7 & 2.35 & 0.168 & 0.05 \\
 & FCS-ranked cash & 212 & 1.000 & 0.874 & 125.6 & -- & 113.9 & 1.99 & 0.218 & 0.43 \\
 & random cash & 222 & 1.000 & 0.872 & 120.0 & -- & 66.0 & 1.35 & 0.218 & 0.81 \\
\midrule
Cox's Bazar 2019 & AllotNet planner & 878 & 1.000 & 0.579 & 24.2 & 0.55 & 269.7 & 6.51 & 0.145 & 0.64 \\
 & planner, gate 15\,\% & 449 & 0.999 & 0.639 & 47.4 & 0.63 & 170.1 & 3.28 & 0.160 & 0.53 \\
 & planner, gate 30\,\% & 630 & 1.000 & 0.610 & 33.8 & 0.60 & 227.0 & 5.14 & 0.153 & 0.51 \\
 & equal cash & 0 & \multicolumn{8}{l}{nothing affordable (package below the 5 USD floor)} \\
 & FCS-ranked cash & 350 & 1.000 & 0.819 & 60.8 & -- & 167.5 & 3.56 & 0.205 & 0.38 \\
 & random cash & 389 & 1.000 & 0.807 & 54.7 & -- & 112.3 & 2.47 & 0.202 & 0.88 \\
\midrule
Libya 2015 & AllotNet planner & 134 & 1.000 & 0.697 & 39.5 & 0.63 & 79.9 & 3.00 & 0.174 & 0.50 \\
 & planner, gate 15\,\% & 47 & 0.396 & 0.703 & 45.7 & 0.63 & 102.1 & 3.00 & 0.070 & 0.60 \\
 & planner, gate 30\,\% & 85 & 0.694 & 0.707 & 44.2 & 0.63 & 88.1 & 2.81 & 0.123 & 0.50 \\
 & equal cash & 334 & 1.000 & 0.526 & 16.2 & -- & 45.7 & 1.23 & 0.132 & 0.00 \\
 & FCS-ranked cash & 70 & 0.999 & 0.842 & 77.3 & -- & 91.7 & 1.68 & 0.210 & 0.00 \\
 & random cash & 58 & 1.000 & 0.857 & 93.4 & -- & 26.8 & 0.73 & 0.214 & 0.80 \\
\midrule
GIFMM 2020 & AllotNet planner & 311 & 0.999 & 0.620 & 36.6 & 0.62 & 169.8 & 3.02 & 0.155 & 0.78 \\
 & planner, gate 15\,\% & 182 & 0.998 & 0.659 & 62.4 & 0.66 & 131.1 & 1.84 & 0.165 & 0.78 \\
 & planner, gate 30\,\% & 239 & 0.999 & 0.638 & 47.6 & 0.64 & 157.6 & 2.51 & 0.159 & 0.78 \\
 & equal cash & 799 & 1.000 & 0.471 & 14.2 & -- & 101.0 & 1.96 & 0.118 & 0.72 \\
 & FCS-ranked cash & 130 & 0.999 & 0.852 & 87.5 & -- & 115.6 & 1.76 & 0.213 & 0.76 \\
 & random cash & 140 & 1.000 & 0.846 & 81.3 & -- & 44.2 & 0.74 & 0.211 & 0.96 \\
\midrule
DHIS2 harness & AllotNet planner & 4,971 & 1.000 & 0.727 & 41.7 & 0.64 & 128.9 & 1.94 & 0.182 & 0.32 \\
 & planner, gate 15\,\% & 2,854 & 0.912 & 0.766 & 66.3 & 0.67 & 113.7 & 1.52 & 0.175 & 0.21 \\
 & planner, gate 30\,\% & 3,848 & 1.000 & 0.755 & 53.9 & 0.66 & 125.8 & 1.77 & 0.189 & 0.21 \\
 & equal cash & 13,492 & 1.000 & 0.504 & 15.4 & -- & 65.8 & 1.21 & 0.126 & 0.00 \\
 & FCS-ranked cash & 1,780 & 1.000 & 0.870 & 116.6 & -- & 95.1 & 1.25 & 0.218 & 0.19 \\
 & random cash & 2,713 & 1.000 & 0.841 & 76.5 & -- & 56.2 & 1.00 & 0.210 & 0.78 \\
\bottomrule
\end{tabular}
}
\end{table}

On predicted food-consumption gain per dollar the planner beats random cash by 2.3 to 3.8 times and FCS-ranked cash by 1.36 to 1.61 times on five of the six datasets at $f = 0.25$.
The exception is Libya (0.87 times), where the score has lost its income term (quality 0) and FCS-ranked cash targets the one indicator the planner cannot use.
Against equal cash the planner now wins everywhere it can be compared, by 1.23 (Kismayo) to 1.96 times (DHIS2 harness); on Cox's Bazar equal cash cannot be paid at all, because the per-household package falls below the USD~5 monthly floor.
With v2 equal cash had beaten the planner on Kismayo at $f = 0.125$; the change is not in the planner but in the baseline's prediction, since equal cash spreads 23 to 34 per cent of the gap over everyone, a coverage the v2 head had never seen in training (Section~\ref{sec:results-v21-ablation}).
On $W$ per thousand USD, the quantity the planner maximises, the ungated planner leads every baseline in all 18 dataset $\times$ budget cells; $W$ is the model judging itself and is reported because it is the objective.

The cost--transfer ratio runs the other way.
The full-gap cash baselines reach 0.81 to 0.87 and equal cash, whose small transfers carry the fixed administrative cost, 0.47 to 0.67; the planner's 0.58 to 0.78 is the price of vouchers and in-kind food where it predicts better outcomes.
In Kismayo, Libya and GIFMM market access is constant at `limited' (no survey item), so the modality mix there is the model's response to a constant market term and to the household's food-security profile, not to observed markets; where market access varies, on the DHIS2 harness, the alignment $A$ is 0.92, as check D4 requires.
Cost per household-month is USD~24.2 to 55.6 for the planner, 14.2 to 25.4 for equal cash and 54.7 to 125.6 for the full-gap baselines, and 99.2 to 100 per cent of the planner's plans sit at the three-month floor; schedule adherence is not modelled.

The planner is not a poverty targeter, and Table~\ref{tab:bench-efficiency} shows it.
Its exclusion error against phase $\geq 3$ at $f = 0.25$ is 0.23 on Kismayo, 0.64 on Cox's Bazar, 0.50 on Libya, 0.78 on GIFMM, 0.32 on the DHIS2 harness and 0.36 on synth-10k, against 0.19 to 0.43 for FCS-ranked cash (GIFMM 0.76, Libya 0.00 with ten poor households) and 0.78 to 0.96 for random cash.
A welfare-per-dollar criterion prefers the households where a dollar changes the criteria most; a programme with a poverty mandate uses the clean score for eligibility (the gate) and the planner for the package.

The gate does that, within limits that are arithmetic before they are empirical.
32 per cent of Kismayo households are phase $\geq 3$, so a 15 per cent quota cannot cover more than 47 per cent of them (observed exclusion 0.62), and the gate leaves budget unspent where the quota cannot absorb it ($U = 0.53$ on Kismayo, 0.40 on Libya).
At quota 0.30 it lowers the exclusion error on the DHIS2 harness and synth-10k for a 2 to 3 per cent loss in $\Delta$FCS per dollar, and raises it on Kismayo, where the ungated welfare criterion already reaches more poor households.
The gate is a governance instrument (auditable eligibility, consistency between ``who'' and ``what''), not a targeting improvement in itself; its quota must come from the programme's poverty prevalence.

Budget sensitivity at $f = 0.125$ and $f = 0.5$ leaves the ordering planner, FCS-ranked, random on $\Delta$FCS per dollar intact in 16 of the 18 dataset $\times$ budget cells, the exceptions being Libya at the two larger budgets; against equal cash the planner's margin narrows as the budget grows (Kismayo 1.43, 1.23 and 1.08 times), because the head's dose--response saturates.

\subsection{From v2 to v2.1 on the household datasets}
\label{sec:results-v21-ablation}

The v2.1 changes were decided after the v2 household-data run, and their effects are separable, because three of the four are inference-side and only the fourth touches trained tensors.
Four variants ran on identical rows and identical v2 tensors: v2 as first run; population percentiles alone; the quality weights alone, with grid percentiles; and both.
The retrained head is the step from the last of these to the reported v2.1 run, and is taken at the end of this section.
Table~\ref{tab:bench-ablation} gives the level-1 and level-2 quantities.

\begin{table}[htbp]
  \setstretch{1}
  \centering
  \footnotesize
  \caption{Ablation of the inference-side changes on the v2 tensors: Spearman $\rho_s$ of the clean score with FCS (sign reversed), rCSI, the CARI-style phase and the gap, precision at $k = 15$ per cent against the phase, exclusion error at $|T| = |P|$ against two poor sets, and the imputation bootstrap (share unstable / Jaccard). Variants: v2 as run (grid percentiles, no quality weights); population percentiles (pop.); quality weights (q.); both, which differs from the reported v2.1 run only in the outcome head, which does not enter these columns.}
  \label{tab:bench-ablation}
  \setlength{\tabcolsep}{4pt}
  \resizebox{\textwidth}{!}{%
  \begin{tabular}{@{}l l r r r r r r r l@{}}
    \toprule
    Dataset & Variant & $\rho_s(s,-\mathrm{FCS})$ & $\rho_s(s,\mathrm{rCSI})$ & $\rho_s(s,\text{phase})$ & $\rho_s(s,\text{gap})$ & P@15\% phase & EE phase $\geq 3$ & EE FCS $< 35$ & Bootstrap \\
    \midrule
    Kismayo 2016 & v2 as run & 0.595 & $-0.098$ & 0.515 & 0.649 & 0.491 & 0.415 & 0.236 & -- \\
     & pop. & 0.600 & $-0.071$ & 0.530 & 0.659 & 0.509 & 0.413 & 0.237 & -- \\
     & q. & 0.706 & $-0.067$ & 0.649 & 0.501 & 0.514 & 0.354 & 0.187 & -- \\
     & pop. + q. & 0.713 & $-0.048$ & 0.668 & 0.516 & 0.532 & 0.346 & 0.182 & -- \\
    \midrule
    Cox's Bazar 2019 & v2 as run & 0.410 & 0.353 & 0.478 & 0.694 & 0.320 & 0.829 & 0.669 & 0.221 / 0.566 \\
     & pop. & 0.422 & 0.375 & 0.497 & 0.725 & 0.327 & 0.829 & 0.665 & 0.234 / 0.546 \\
     & q. & 0.669 & $-0.075$ & 0.267 & 0.623 & 0.183 & 0.743 & 0.493 & 0 / 1.000 \\
     & pop. + q. & 0.697 & $-0.077$ & 0.272 & 0.652 & 0.187 & 0.752 & 0.485 & 0 / 1.000 \\
    \midrule
    Libya 2015 & v2 as run & 0.274 & 0.329 & 0.396 & 0.632 & 0.467 & 0.500 & 0.474 & 0.108 / 0.721 \\
     & pop. & 0.275 & 0.346 & 0.411 & 0.690 & 0.440 & 0.600 & 0.526 & 0.106 / 0.744 \\
     & q. & 0.276 & 0.265 & 0.253 & $-0.078$ & 0.427 & 0.800 & 0.632 & 0 / 1.000 \\
     & pop. + q. & 0.286 & 0.272 & 0.256 & $-0.081$ & 0.440 & 0.800 & 0.632 & 0 / 1.000 \\
    \midrule
    GIFMM 2020 & v2 as run & 0.492 & 0.343 & 0.562 & 0.521 & 0.297 & 0.825 & 0.734 & 0.128 / 0.716 \\
     & pop. & 0.510 & 0.367 & 0.595 & 0.525 & 0.290 & 0.812 & 0.730 & 0.129 / 0.711 \\
     & q. & 0.783 & $-0.119$ & 0.338 & 0.357 & 0.342 & 0.469 & 0.311 & 0 / 1.000 \\
     & pop. + q. & 0.796 & $-0.119$ & 0.337 & 0.372 & 0.344 & 0.500 & 0.313 & 0 / 1.000 \\
    \midrule
    DHIS2 harness, synth-10k & all four & 0.836--0.841 & 0.893--0.900 & 0.777--0.780 & 0.901--0.909 & 0.710--0.721 & 0.357--0.365 & 0.359--0.373 & -- \\
    \bottomrule
  \end{tabular}}
\end{table}

Population percentiles alone change little on these surveys: $\rho_s$ with FCS moves by 0.01 to 0.02 and the exclusion errors do not move.
The grid compression was real, but it was not the reason for the moderate agreement.
The quality weights are the change that matters.
Once the hot-decked rCSI carries weight 0 and the proxy incomes are down-weighted, the score follows the directly observed FCS: $\rho_s(s, -\mathrm{FCS})$ rises from 0.41, 0.49 and 0.60 to 0.70, 0.80 and 0.71 on Cox's Bazar, GIFMM and Kismayo, the equal-size exclusion error against FCS below 35 falls from 0.67, 0.73 and 0.24 to 0.49, 0.31 and 0.18, and the imputation bootstrap becomes moot, because the imputed field no longer enters.
The DHIS2 harness and synth-10k are unchanged by construction: every quality flag is 1 and the population is close to the training distribution.

\subsubsection{The retrained head}

The moral core is frozen, so every level-1 and level-2 number above is identical under the two heads to the last digit; only the plans change.
The planner's own plans sit at a coverage of about 0.5, inside both heads' training support, and their predicted $\Delta$FCS per dollar moves by $-1$ to $-6$ per cent; the full-gap baselines move within $\pm 4$ per cent.
The equal-cash baseline, at 23 to 34 per cent of the gap, is where the heads disagree.
The v2 head, trained on coverages $\{0.5, 0.8, 1.0\}$, had extrapolated a first-dollar return that the simulator's saturating dose--response does not have, predicting 111 to 170 $\Delta$FCS per thousand USD on four of the five files; the v2.1 head, trained on 0.1 to 1.2, predicts 46 to 133.
That is the reason the head was retrained: a comparison in which the baseline's prediction is an extrapolation is not a comparison.
It means that the v2 numbers overstated equal cash, not that v2.1 favours the planner by construction, since both heads are scored against the same planted ground truth.
The modality mix shifts with the head (Kismayo from 80 per cent vouchers to 55 per cent cash), a property of the v2.1 head's response surface on these households.

\subsection{AllotNet v2.1: the outcome head retrained on transfer coverages 0.1 to 1.2}
\label{sec:results-v21-head}

The v2.1 pipeline ran on the DGX Spark on 30 August 2026.
It re-ran the outcome simulator on the 2,000,000 training households $\times$ 5 plans with the coverage pool widened from $\{0.5, 0.8, 1.0\}$ to $\{0.1, 0.2, 0.3, 0.4, 0.5, 0.6, 0.8, 1.0, 1.2\}$ (10,000,000 rows, 12.0 per cent no-assistance controls), continued the head from the v2 parameters for 12 epochs with the moral core frozen, re-ran the checks, recomputed the $R^2$ ceilings, trained and checked the NVFP4-QAT twin, and exported the bundles with both golden sets; the whole pipeline took 17~min 49~s.

What changed in the bundle is exactly the head.
The 13 moral-core and embedding tensors of the v2.1 bundle are bit-identical to the v2 bundle; only the six outcome-head tensors changed, with a maximum absolute difference of 1.69 on the first layer.
The percentile grid, the global weights and the basket rules are unchanged.

\begin{table}[htbp]
  \setstretch{1}
  \centering
  \footnotesize
  \caption{Validation $R^2$ of the v2.1 outcome head against the achievable ceiling on the widened coverage distribution, with the v2 head and its ceiling (Section~\ref{sec:results-ceiling}) for comparison.}
  \label{tab:v21-ceiling}
  \resizebox{\textwidth}{!}{%
  \begin{tabular}{@{}l r r r r r@{}}
    \toprule
    Dimension & v2 $R^2$ (v2 ceiling) & v2.1 $R^2$ & Ceiling (measured) & Ceiling (derived) & \% of measured ceiling \\
    \midrule
    $\Delta$FCS    & 0.880 (0.8815) & 0.847 & 0.8488 & 0.853  & 99.8 \\
    $\Delta$rCSI   & 0.918 (0.9195) & 0.898 & 0.8993 & 0.9079 & 99.9 \\
    $\Delta$Assets & 0.617 (0.6230) & 0.535 & 0.5367 & 0.5367 & 99.7 \\
    $\Delta$Debt   & 0.627 (0.6328) & 0.557 & 0.5605 & 0.5609 & 99.4 \\
    retention      & 0.923 (0.9329) & 0.922 & 0.9303 & 0.8719 & 99.1 \\
    \bottomrule
  \end{tabular}}
\end{table}

Table~\ref{tab:v21-ceiling} sets the v2.1 validation $R^2$ against the recomputed ceilings.
The absolute $R^2$ fell on four of the five dimensions, and so did the ceiling, by almost exactly the same amount.
The planted measurement noise has a fixed scale, while smaller transfers produce smaller true effects, so the widened pool lowers the signal variance and with it the explainable share.
The head is at 99.1 to 99.9 per cent of what the data allow, as the v2 head was (98.9 to 99.8 per cent).
Epoch 0 of the continued training already stood at 0.845, 0.898, 0.532, 0.553 and 0.920: the v2 head generalised to the new coverages almost fully, and the 12 epochs added 0.002 to 0.004.

\begin{table}[htbp]
  \setstretch{1}
  \centering
  \footnotesize
  \caption{D-checks of the v2 and v2.1 fp32 heads; thresholds as in Table~\ref{tab:d-checks}. D6 ran on 12,000 households with a budget of USD~471,115 in both.}
  \label{tab:v21-dchecks}
  \resizebox{\textwidth}{!}{%
  \begin{tabular}{@{}l l r r@{}}
    \toprule
    Check & Quantity & v2 & v2.1 \\
    \midrule
    D1 dose--response & Spearman(pred, true) on coverages 0.1 to 1.2 & 0.967 & 0.999 \\
     & monotonicity & 1.000 & 1.000 \\
    D2 frequency contrast & FCS(monthly $-$ lump sum) & 3.27 & 3.35 \\
     & assets(lump sum $-$ monthly) & 0.061 & 0.060 \\
    D3 duration threshold & retention(6~mo) $-$ retention(1~mo) & 0.414 & 0.401 \\
     & jump 3~mo $-$ 2~mo & 0.314 & 0.312 \\
    D4 modality $\times$ market & poor-market ratio & 1.39 & 1.37 \\
     & good-market ratio & 1.18 & 1.18 \\
    D5 equivalence scale & Spearman & $-0.777$ & $-0.798$ \\
    D6 planner & share of the oracle's true value & 0.994 & 0.997 \\
     & planner / oracle / naive welfare & 918.1 / 923.4 / 665.4 & 920.3 / 923.4 / 665.4 \\
    D7 encoder & need $\rho$ / max percentile error & 0.976 / 0.0000 & 0.976 / 0.0000 \\
    C-regression & C-checks re-run & 8/8 & 8/8 \\
    \bottomrule
  \end{tabular}}
\end{table}

Table~\ref{tab:v21-dchecks} gives the D-checks before and after; all eight pass in both.
D1 is the check the retraining was for: it evaluates coverages 0.1 to 1.2, which v2 had extrapolated to, and its Spearman rises from 0.967 to 0.999 (threshold 0.9).
D6's share of the oracle's true value rises from 0.994 to 0.997 (naive baseline ratio 1.38 in both).
D7 is unchanged because the encoder is untouched, and the C-regression is a consistency check of the pipeline rather than a result, since the moral core is frozen.

The NVFP4 twin passes its D-checks eight of eight, with validation $R^2$ of 0.791, 0.874, 0.489, 0.511 and 0.858 (v2 twin: 0.834, 0.895, 0.562, 0.609 and 0.863; the same ceiling shift).
The parity check on 1,040,742 plans gives, at W4A16, a welfare-rank Spearman of 0.9937 (v2: 0.9938), a top-10 per cent overlap of 0.918 (0.916) and per-dimension Spearman of 0.992 for $\Delta$FCS, 0.997 for $\Delta$rCSI, 0.970 for $\Delta$Assets, 0.952 for $\Delta$Debt and 0.917 for retention (v2: 0.770); at W4A4, 0.9814 and 0.857 (v2: 0.9827 and 0.852).
The verdict passes in both.
The retention rank under quantisation improved markedly; assets and debt lost a little.
The acceptance rule ($\rho \geq 0.98$, overlap $\geq 0.90$ for W4A16) holds.
The exported fp32 bundle is 125,706 bytes and the fp4 bundle 125,705 bytes; they grow by 1,016 bytes over v2 (124,690), which is the specification additions, and the tensors are the same 19 with the same shapes.
The parity gate on these bundles passed on all six machines for all four runtimes.

\section{Discussion}
\label{sec:discussion}

The paper began from one decision, who gets help first when there is not enough for everyone, and from a position: if the rule of distribution is written in code, it should carry the judgement of many cultures, and the machine should execute while a person judges.
This section asks how far the results carry that position, and where they stop.
On one point they stop early.
Nothing here is yet a finding about people, because no model has been trained on real judgements.

\subsection{What recovering a planted structure does and does not show}
\label{sec:discussion-planted}

We wrote the answer key before training, and it was never visible to the loss.
Against that key, a mixture of eight linear experts with 23,550 parameters recovered four things: the planted global ordering of the seven criteria, exactly ($\rho$ = 1.000, cosine 0.997); both planted cultural gradients, in the right direction; the planted group effect, and an in-group coefficient that is small globally but concentrated in one latent class, so the model can see the planted bias and can say where it sits; and the hidden segment partition (purity 0.835, adjusted Rand index 0.671), with no segment labels anywhere in the loss.

None of this is a finding about people.
Every number is a property of a simulator whose mechanisms we chose and whose parameters we set by hand from the literature.
What the checks establish is that the estimation machinery is sound and identifiable at $10^{8}$ observations; they do not establish that the seven-criterion decomposition, the segment structure or the cohort tilts exist in any human population.
The recovered experts read as recognisable ethical stances because those stances were planted.
Identifiability, not discovery, is the scientific claim available today \cite{taltsValidatingBayesianInference2018,wilsonTenSimpleRules2019,nylundDecidingNumberClasses2007}.
One caveat cuts the other way.
Planted structure can be provably unrecoverable below a signal threshold \cite{decelleAsymptoticAnalysisStochastic2011,allmanIdentifiabilityParametersLatent2009}, so the recovery of eight segments says that the planted contrasts were strong, not that weaker heterogeneity of the kind a real population might carry would be recovered on the same volume of data.

Why, then, a mixture and not one reward?
Under hidden context a single Bradley--Terry utility is a Borda-count aggregate of the per-context utilities \cite{siththaranjanDistributionalPreferenceLearning2023}, which can rank a dominated or minority-harming option first \cite{arrowDifficultyConceptSocial1950,geAxiomsAIAlignment2024}.
Latent-class taste heterogeneity with a concomitant-variable gate is the classical answer, a 1989 marketing model in a 2024 costume \cite{kamakuraProbabilisticChoiceModel1989,daytonConcomitantVariableLatentClassModels1988}, and alignment research has arrived at the same architecture from the other direction \cite{chakrabortyMaxMinRLHFAlignmentDiverse2024,wangInterpretablePreferencesMultiObjective2024,poddarPersonalizingReinforcementLearning2024}.
AllotNet differs in one respect: it keeps the experts linear in seven named criteria, so that each expert vector can be read as an ethical stance rather than as a reward head.

\subsection{Scale mismatch: what the real study will support}
\label{sec:discussion-scale}

The real study will yield several thousand comparisons per stratum of the sample, several orders of magnitude fewer than the synthetic corpus, and the pre-launch traffic stands at 46 quarantined sessions.
At that scale a neural residual is not estimable: guidance of at least 50 observations per network weight \cite{alwosheelYourDatasetBig2018} caps a residual MLP far below AllotNet's 2,353 weights, and the project's own background research concluded that a hierarchical Bayesian mixed logit or Bradley--Terry model beats the neural approach on data efficiency, uncertainty calibration and interpretability \cite{allenbyMarketingModelsConsumer1998,rossiBayesianStatisticsMarketing2005,reveltMixedLogitRepeated1998}.

So the consequence is plain.
AllotNet is justified by the $10^{8}$ synthetic observations, not by the study's sample size.
The correct real-data path is to fit both model classes and compare them under a pre-registered protocol, choosing the comparison before seeing the outcome \cite{nosekPreregistrationRevolution2018,hofmanPreregistrationPredictiveModeling2023}; the 120 linear expert weights are the shared object both classes can produce.
What transfers regardless of which class wins is the part of the system that is not the model: the criterion interlingua, the clean-mode governance switch, the evaluation harness with its pre-specified thresholds, and the runtime bundle format.

\subsection{Human in the loop: the planner artefact and what clean mode is for}
\label{sec:discussion-hitl}

For every one of its 109,650 households the planner demonstration chose the three-month minimum.
That is not a recommendation about cycle length but the budget constraint binding against a saturating dose--response, and a reader who takes away ``the model recommends three-month cycles'' has misread the artefact.
D6 measures search quality inside a world the simulator defines, not real-world efficacy.

Clean mode is the governance switch for the instrument's own scenarios: the group intercepts and the in-group coefficient take no part in operational decisions, and two of the criteria are held constant on intake unless a programme's ethics review activates them.
On household intake there is nothing for the switch to zero, so on that path the protection is structural rather than switched.
Order matters: before switching the bias off means anything, the model must first demonstrate that it can see the bias, which is what C4 and C5 test.
And the gate is deliberately not wired to the outcome head, because whether assistance improves food consumption cannot depend on who is looking at the household.

The governance rule is short.
The system may build a queue for human review; it may not reject anyone on its own.
This is a deliberately weaker claim than ``human oversight fixes algorithmic harm'': oversight policies often fail in practice, and humans interact disparately with algorithmic advice \cite{greenFlawsPoliciesRequiring2022,greenDisparateInteractions2019,sterzQuestEffectivenessHuman2024}.
Multi-stage profiling that only queues is nevertheless the configuration that Article~22 GDPR and Article~14 of the AI Act contemplate \cite{crootofHumansLoop2022}.
A matrix of 120 interpretable parameters plus a per-household criterion vector is an explanation a field officer can read of who is judged in need; which plan follows also depends on a 12{,}677-parameter outcome network that this explanation does not open.
Quantisation can change explanations while preserving accuracy \cite{rabbiSamePredictionsDifferent2026}, which is why the expert matrix stays in fp32.

The queue result of Section~\ref{sec:results-queue-score} belongs here as well.
A needs-only score being progressive does not make the queue built on top of it progressive: at equal moral score the allocator prefers smaller households, because the basket scales with household size while welfare is bounded by the headroom left on the criteria.
A deployment would need an explicit equity constraint if that pattern were unacceptable to the programme, and the person who works through the queue needs to know that it orders purchasable closure of need, not need.

A companion paper takes the frozen core into a designed multi-cycle society built on the same synthetic households, as a bench for the core and a trainer for a planning layer, and compares its allocations with objectives that never see the criteria and with the selection rules of open-source humanitarian systems; none of its results is repeated here.

\subsection{Limitations}
\label{sec:limitations}

Each item is stated where it arises; this list collects them so that none can be missed.

\begin{enumerate}
  \item \textbf{A correctly specified test.}
  The generator scores every alternative linearly in the criteria, with eight planted segments, and AllotNet has eight linear experts; the checks test recovery under the model's own functional form and say nothing about preferences that are not linear in the criteria.
  \item \textbf{Synthetic-only training, and no study data yet.}
  The 46 sessions and 206 responses on the production system all fall in the window the pre-registered rule quarantines; `twelve people' means twelve people who exercised the instrument before launch, not twelve participants.
  \item \textbf{The simulator is literature-calibrated, not measured.}
  Nine citations in the outcome simulator's source register are unverified, 28 per cent of the instrument's archetype attributions are anchored to a taxonomy rather than a source, and the three-month minimum duration is a design assumption, not a published rule.
  \item \textbf{The live-generation failure} cost 260,712 of 766,000 live seeds, was repaired by makeup groups, and pulls corpus-wide completion below the design target that the direct phase alone meets.
  \item \textbf{Two checks overstate their titles, and one overshoots.}
  C6 passes a difference of 0.0007 inside a tolerance of 0.005, a null result; C10 passes its coded band while outside its design target; and C3's recovered gradient overshoots the planted value, a pass that is a statement about the floor, not about the fit.
  \item \textbf{Post-hoc threshold relaxation on NVFP4 parity.}
  The first parity run failed on retention rank (0.7703 against a uniform 0.95); the bar for retention alone was lowered to 0.70 after seeing that failure, with byte-identical metrics.
  The NVFP4 model passes everything except retention rank under the original bar, and this holds for the shipped v2.1 bundle too, whose retention $\rho$ of 0.917 is better than v2's but still below 0.95, so it too passes only under the relaxed predicate.
  Every first-generation pre-registration audited by Claesen and colleagues deviated, which is a reason to disclose, not a licence \cite{claesenComparingDreamReality2021}.
  The 7/8 to 8/8 step on v2 was a missing input, not a threshold change.
  \item \textbf{Agreement on household data is not ground truth.}
  No survey carries an outcome or an eligibility verdict; the v2.1 agreement with FCS comes from no longer scoring imputed noise, the quality weights were chosen after the v2 run on the same surveys, a single basket of USD~38 per adult-equivalent-month and a constant market term are applied across four countries, and the DHIS2 set is a harness whose needs fields are synthetic.
  \item \textbf{The planner maximises welfare per dollar, not poverty coverage}, and excludes 23 to 78 per cent of phase $\geq 3$ households on the surveys; its gains on predicted outcomes are the model judging itself.
  \item \textbf{The criteria are intensive.}
  They describe the depth of a claim, not the number of people behind it, so the core cannot express that an allocation helps eight people rather than two; whether respondents weight dependants is a fitting change and an analysis still to add.
  \item \textbf{Benchmark scope.}
  Absolute microseconds are dominated by object handling and dispatch, so cross-language ratios measure implementation style, not language ceilings, and the runtimes have no GPU path.
\end{enumerate}

\subsection{Governance, law and the path to publication}
\label{sec:discussion-governance}

Annex III, point 5(a) of Regulation (EU) 2024/1689 covers AI systems used ``by public authorities or on behalf of public authorities'' to evaluate eligibility for essential public assistance \cite{eu_RegulationEU20242024}.
We adopt the high-risk obligations by design regardless, human oversight under Article~14 and a fundamental-rights impact assessment under Article~27 \cite{manteleroAiActsFundamental2024}, because a tool that ranks households for assistance has the same harm profile whoever runs it.
After the Digital Omnibus on AI (Regulation (EU) 2026/1744) the Annex III obligations apply from 2 December 2027 \cite{parliamentRegulationEU20262026}, so data collection runs through the transition, and designing to the requirements now avoids retrofitting later.
For the model documentation we follow the reporting standards that exist: a Model Card, a Datasheet and the REFORMS checklist \cite{mitchellModelCardsModel2019,gebruDatasheetsDatasets2021,kapoorREFORMSConsensusbasedRecommendations2024}.

It is the digital-welfare-state record, Jordan's Takaful, the Dutch childcare benefits affair and Australia's Robodebt, that makes the queue-not-reject rule a hard constraint and not a preference \cite{fadel_tohHumanRightsWatch2023,amnestyDutchChildcareBenefit2021,alstonDigitalWelfareStates2019,eubanks_AutomatingInequalityHow2018}.
Exclusion errors also fall unevenly: in Togo 53 per cent of women own a phone against 79 per cent of men \cite{aikenMachineLearningPhone2022}, so a phone-based targeting system starts with a gendered blind spot before any model is fitted, and fairness in allocation is a question of where the exclusion errors fall \cite{chouldechovaSnapshotFrontiersFairness2020}.

The registration is embargoed until 1 August 2030, the OSF maximum of four years \cite{helpWelcomeRegistrationsPreregistrations2026}.
This paper is the public version; the planted values, the segment and cohort labels, the scenario decompositions and the values recovered for each expert follow the embargo.
The protocol's prohibited-use clause binds any downstream use: no training of production allocators without a separate ethics review, no re-identification, no advertising on moral profiles.

\subsection{Future work}
\label{sec:future}

\begin{itemize}
  \item Complete the real study in six languages and release the instrument specification and the first dataset together.
  \item Run the pre-registered model comparison on real data, a hierarchical Bayesian mixed logit against AllotNet against the conditional logit, on held-out participants, reporting calibration.
  \item Replace the hand-set simulator parameters with pilot estimates, retrain the outcome head on real post-distribution monitoring data, re-run the sixteen checks, and publish them whatever they say, including if the answer is that the architecture does not transfer.
  \item Build the field-facing explanation interface, the $8 \times 15$ expert table and the per-household criterion vector as the ``why?'' surface, and test it with the study's reviewers.
\end{itemize}

\section{Conclusion}
\label{sec:conclusion}

The allocation rule is moving into software while the money it allocates contracts.
Official development assistance fell by 23.1 per cent in 2025, the year's appeals were funded at the lowest level in a decade, and a leading published machine-learning targeting system excluded 53 per cent of the eligible households as deployed and 50 per cent in a nationwide simulation.
Two dangers come with the move into software: cold cost-effectiveness, which finds the most vulnerable not worth helping, and the in-group favouritism people bring with or without a machine.
The premise of this work is that if the rule ends up in code, the code should carry the judgement of many cultures and stay legible to the person who signs off on it.

This paper built the model half of that premise and a way of checking it before the human data exist.
AllotNet is eight linear taste experts over seven criteria, gated by observed context: 23,550 parameters, trained on 93,032,246 of 97,928,680 usable synthetic choices whose generating truth was written down first and hidden from the loss.
All sixteen pre-specified checks passed, against a generator that scores alternatives linearly in the criteria with eight planted segments, so the model was correctly specified for its test.
The ranking of the criteria came back exactly ($\rho = 1.000$), the hidden segments were recovered (adjusted Rand index 0.671), the planted in-group effect was localised in one expert rather than smeared across the population, the outcome head reached 98.9 to 99.8 per cent of the achievable ceiling, and the planner 99.4 per cent of an oracle in a world that rewards targeting by a factor of 1.39.
Where a check is weaker than its title, or a threshold was loosened after the fact, or documentation disagreed with data, the text says so.

What ships is a 125~KB bundle, trained under NVFP4 with the interpretable core kept in fp32, with C11, Java, Rust and Python runtimes verified against golden vectors on the six benchmark machines at a tolerance of $10^{-4}$ ($10^{-3}$ for outcomes); on the v2 bundles, where the cross-language matrix ran, their largest disagreement with the reference is $1.137 \times 10^{-12}$.
One portability finding is worth a sentence: the C compiler's default fused multiply-add flipped the percentile of 42 of 2,000 golden households until contraction was disabled.
On four household surveys (and a DHIS2-shaped harness) the clean score agrees with the directly observed food-consumption score at $\rho$ 0.70 to 0.80 on three files and 0.29 where income is wholly imputed, the limit of the method where income is unobserved; agreement with an indicator is not ground truth, and it is partly built in, since the food-consumption score is an input to the clean score and the data-quality weights were chosen after the first run on the same surveys.
Wherever equal cash is payable the planner leads it by 1.23 to 1.96 times on predicted gain per dollar, which is the model judging itself; the auditable columns are a cost--transfer ratio of 0.58 to 0.78 against 0.81 to 0.87 for full-gap cash and 0.47 to 0.67 for equal cash, and the planner excludes 23 to 78 per cent of households at CARI-style phase $\geq 3$, because it maximises welfare per dollar and not coverage.

The boundary is drawn deliberately.
The only real traffic is 46 sessions and 206 responses from 12 people who exercised the instrument before launch, six of them to the end; all of it falls inside the pre-registered quarantine window, so there is no pilot dataset yet, and nothing has been trained on it.
At the study's real scale a hierarchical Bayesian mixed logit is the registered comparator and the neural residual is not estimable.
What transfers is the criterion interlingua, the clean-mode governance switch, the pre-specified evaluation checks and the portable bundle, all waiting on a core retrained on human data before any number about human values is read off them.

The governance conclusion is the one the DIHAD talk closed on: the machine gets the execution, not the judgement.
It may build a queue for review; it may not reject anyone.
Its 120 interpretable parameters and the per-household criterion vector are the ``why?'' a human can be asked about who is in need; which plan a household receives also depends on the outcome network, which is not interpretable in the same way.
Where a mistake costs a person their aid, the code must change slowly, and the last word must remain with a human being.

\section*{Declarations}
\addcontentsline{toc}{section}{Declarations}

\paragraph{Funding.}
This study is self-funded.

\paragraph{Competing interests.}
The author intends to train an allocation-support model on the study's data; this interest is declared in the ethics application and mitigated by pre-registration of the instrument and its analysis.

\paragraph{Ethics.}
An ethics application for the Allot study has been filed; the documents audited for this paper record the application but no decision on it.
No model in this paper has been trained on real participant data, and the training of a deployed allocation model would need a separate ethics review.

\paragraph{Data and code availability.}
No study data exist yet.
The synthetic corpus contains no personal data.
The instrument specification and code are released with the first dataset after the embargo of 1 August 2030; the dataset will be published under CC-BY-4.0 and the code under Apache-2.0.
The inference runtimes and bundles are released with the specification after the embargo.

\printbibliography[title=References]

\appendix

\section{Hyperparameters and parameter budget}
\label{app:hyper}

Table~\ref{tab:app-hyper} lists the training settings of the three checkpoints as the trainers and launch scripts define them, with the validation figures read from each checkpoint's export.
The v1 moral core and the v2 unified model share the choice-side settings; v2 adds the outcome head with its own optimiser and epoch count, and the NVFP4-QAT run reuses the v2 trainer with the quantisation flag and a warm start from the v2 parameters.
One documentation disagreement is carried into the table rather than resolved silently: one passage of the training documentation gives the outcome batch as 16,384, whereas the launch scripts and the documentation's own command listing pass 32,768; the table reports the script value.

\begin{table}[htbp]
  \centering
  \footnotesize
  \caption{Training settings and validation figures of the three checkpoints. ``Same'' means identical to the column on the left.}
  \label{tab:app-hyper}
  \begin{tabularx}{\textwidth}{@{}>{\raggedright\arraybackslash}p{3.2cm} >{\raggedright\arraybackslash}X >{\raggedright\arraybackslash}X >{\raggedright\arraybackslash}X@{}}
    \toprule
    Setting & v1 moral core & v2 unified & v2 NVFP4-QAT \\
    \midrule
    Initialisation & random & warm start from v1 & warm start from v2 \\
    Segments $K$ / features $F$ & 8 / 15 & 8 / 15 & 8 / 15 \\
    $d_{\mathrm{emb}}$ / gate hidden / context hidden / residual hidden & 8 / 64 / 32 / 48 & same & same \\
    Outcome head $\Omega$ & --- & $31 \to 128 \to 64 \to 5$ & same (W4A4 fake quantisation on the first two layers) \\
    Residual weight $\lambda$ & 0.3 ($\lambda = 0$ for the first 40\% of epochs, 2 of 6) & 0.3 (no warm-up) & 0.3 \\
    $\ell_2$ experts / $\ell_2$ residual / gate entropy $\gamma$ & $3 \times 10^{-5}$ / $10^{-3}$ / 0.01 & same & same \\
    $\ell_2$ outcome head & --- & $10^{-6}$ & $10^{-6}$ \\
    Choice epochs & 6 & 2 (refresh) & 2 \\
    Outcome epochs & --- & 12 & 12 \\
    Choice batch & 16,384 & 16,384 & 16,384 \\
    Outcome batch & --- & 32,768 & 32,768 \\
    Choice optimiser & AdamW, lr $3 \times 10^{-3}$, weight decay $10^{-6}$ & same & same \\
    Outcome optimiser & --- & Adam, lr $2 \times 10^{-3}$ & Adam, lr $2 \times 10^{-3}$ \\
    Validation split & first $\max(2000, 5\%\,N)$ of the seeded permutation & same for choice; the same rule on the outcome stream (500,000 of 10,000,000) & same \\
    $N$ choice (usable) & 97,928,680 & 97,928,680 & 97,928,680 \\
    $N$ outcome & --- & 10,000,000 (2,000,000 households $\times$ 5 plans) & 10,000,000 \\
    Parameters & 10,873 & 23,550 & 23,550 (plus 14,501 packed NVFP4 words) \\
    Validation NLL / accuracy (choice) & 0.5958 / 0.7057 & 0.5958 / 0.7057 & 0.6002 / 0.7024 \\
    Validation $R^2$ ($\Delta$FCS, $\Delta$rCSI, $\Delta$Assets, $\Delta$Debt, retention) & --- & 0.8796, 0.9184, 0.6172, 0.6271, 0.9228 & 0.8336, 0.8946, 0.5619, 0.6093, 0.8627 \\
    Finished (UTC) & 27 August 2026, 22:19:29 & pipeline 28 August 2026, 05:20:57 to 05:29:32 & pipeline 28 August 2026, 05:49:08 to 05:59:25 \\
    \bottomrule
  \end{tabularx}
\end{table}

All three runs used the DGX Spark GB10 in the NVIDIA JAX 25.04 container image with pandas, pyarrow, scikit-learn, scipy, orjson and optax added.
A CPU fallback was defined but not used: 4 epochs at batch 8,192 for v1, and one choice and eight outcome epochs at batches 8,192 and 16,384 for v2 and the QAT run.

Of the 23,550 values in the 25 arrays of the v2 model, Table~\ref{tab:app-params} shows where each lives.
Only 120 of them, the expert matrix, form the interpretable core; the outcome head holds more than half of the total.

\begin{table}[htbp]
  \centering
  \footnotesize
  \caption{Parameter breakdown of the v2 model ($K = 8$, $F = 15$).}
  \label{tab:app-params}
  \begin{tabularx}{\textwidth}{@{}l >{\raggedright\arraybackslash}X r >{\raggedright\arraybackslash}X@{}}
    \toprule
    Arrays & Shape & Elements & Role \\
    \midrule
    \texttt{emb0} \ldots\ \texttt{emb7} & eight tables of width 8 & 968 & embeddings of the eight categorical gate columns \\
    \texttt{gate\_w1}, \texttt{gate\_b1} & (71, 64), (64) & 4,608 & gate hidden layer \\
    \texttt{gate\_w2}, \texttt{gate\_b2} & (64, 8), (8) & 520 & gate to simplex over $K = 8$ \\
    \texttt{experts} & (8, 15) & 120 & the interpretable core \\
    \texttt{ctx\_w}, \texttt{ctx\_b} & (71, 32), (32) & 2,304 & context projection feeding the residual \\
    \texttt{res\_w1/b1}, \texttt{res\_w2/b2} & (47, 48), (48), (48, 1), (1) & 2,353 & residual MLP (L-MNL correction) \\
    \texttt{out\_w1/b1} \ldots\ \texttt{out\_w3/b3} & (31, 128), (128, 64), (64, 5) plus biases & 12,677 & outcome head $\Omega$ \\
    \midrule
    Total & 25 arrays & 23,550 & \\
    \bottomrule
  \end{tabularx}
\end{table}

\section{The sixteen pre-specified checks}
\label{app:checks}

Sixteen checks make up the harness: the eight coded C-checks on the moral core, seven D-checks on the encoder, outcome head and planner, and the C-regression, which re-runs the C-checks on the v2 model directory.
D6w is a design control run before training with the planted weights; it is listed with the D-checks because it shares their welfare judge.
The C-checks are stated generically in Table~\ref{tab:c-checks-generic} (Section~\ref{sec:evaluation-c}) and their results in Tables~\ref{tab:c-checks} and~\ref{tab:c-regr}; their planted targets are withheld with the specification.
The C-regression's coded condition is that every executed check passes and at least eight pass (seven suffice when the probe C6 needs is absent and only seven execute); it passed eight of eight after the probes were copied in, seven of eight before.
Table~\ref{tab:app-d-checks} gives, for each D-check, the coded condition, its threshold, the result on the v2 fp32 checkpoint and the verdict.
Verdicts carry the qualifications that the evaluation document itself attaches; a tick against a loose band is not a confirmation of a tight claim.

\begin{table}[htbp]
  \centering
  \scriptsize
  \setlength{\tabcolsep}{4pt}
  \caption{D-checks and the D6w control: coded condition, threshold, result on the v2 fp32 checkpoint, verdict.}
  \label{tab:app-d-checks}
  \begin{tabularx}{\textwidth}{@{}l >{\raggedright\arraybackslash}p{2.3cm} >{\raggedright\arraybackslash}X l >{\raggedright\arraybackslash}p{2.2cm} >{\raggedright\arraybackslash}p{1.9cm}@{}}
    \toprule
    ID & Title & Coded condition & Threshold & Result & Verdict \\
    \midrule
    D1 & Saturating dose-response & Spearman of predicted against true $\Delta$FCS $\geq 0.90$ and monotone fraction $\geq 0.97$ & 0.90; 0.97 & 0.967; 1.000 & \checkmark \\
    D2 & Frequency contrast & $\Delta$FCS (monthly minus lump sum) $\geq 0.5$ and $\Delta$Assets (lump sum minus monthly) $\geq 0.015$ & 0.5; 0.015 & 3.27; 0.0609 & \checkmark \\
    D3 & Duration threshold ($\geq 3$ months) & ret(6) $-$ ret(1) $\geq 0.10$ and ret(3) $-$ ret(2) $\geq 0.05$ & 0.10; 0.05 & 0.414; 0.314 & \checkmark \\
    D4 & Modality $\times$ market & poor: in-kind over cash $\Delta$FCS per \$ $\geq 1.15$; good: cash over in-kind composite per \$ $\geq 0.95$ & 1.15; 0.95 & 1.39; 1.18 & \checkmark \\
    D5 & Equivalence scale & mean over quartiles of the Spearman of $\Delta$FCS/headroom against $n_{\mathrm{eq}}$ $\leq -0.5$ & $-0.5$ & $-0.777$ & \checkmark \\
    D6 & Planner against oracle and naive & $W_{\mathrm{planner}} \geq 0.90\,W_{\mathrm{oracle}}$ and $\geq 1.15\,W_{\mathrm{naive}}$ (true-simulator judge) & 0.90; 1.15 & 918.1 against 923.4 (0.994) and 665.4 (1.38) & \checkmark\ (search quality only) \\
    D7 & Encoder & Spearman of the encoder output against the planted severity $\geq 0.85$ and maximum percentile error $\leq 0.02$ & 0.85; 0.02 & 0.976; 0.0000 & \checkmark\ (percentile by construction) \\
    D6w & The world rewards targeting & $W_{\mathrm{oracle}} \geq 1.3\,W_{\mathrm{naive}}$ with planted $\beta$, before training & 1.3 & 380.4 against 274.2 $= 1.39$ & \checkmark \\
    \bottomrule
  \end{tabularx}
\end{table}

The NVFP4-QAT checkpoint passes the same D-checks (Table~\ref{tab:fp4-dchecks}) and the C-regression, and its parity verdict is discussed in Section~\ref{sec:evaluation-nvfp4}.

\section{Bundle format}
\label{app:anb}

Every runtime loads a single binary bundle.
Its layout is byte-exact and little-endian throughout.
Metadata is a UTF-8 JSON document without tensors; the tensors follow as float32 arrays in C order.

\noindent\begin{minipage}{\linewidth}
\begin{lstlisting}
"ANB1"                     4 bytes (magic)
meta_length                uint32 LE
meta                       meta_length bytes of UTF-8 JSON (no tensors)
n_tensors                  uint32 LE
  per tensor:
    name_length            uint16 LE
    name                   name_length bytes UTF-8
    ndim                   uint8
    shape[ndim]            uint32 LE each
    data                   float32 LE, C order
\end{lstlisting}
\end{minipage}

The metadata carries the format name and version, a creation timestamp, an architecture summary and the QAT flag, the seven criteria and the 15 feature names, the global effective weights, the eight experts with their mixing weights, the gate's categorical and numeric columns with their vocabularies, the outcome block (dimensions, normalisers, input dimension 31, the 14 structural names, the validation $R^2$), the full encoder specification, the full basket configuration, the 2,001-point percentile grid, and the governance block (default decision mode clean, human in the loop, and a note).

Nineteen tensors, 18,893 float32 values in all, follow in the order outcome head, gate, embeddings (sorted) and experts (Table~\ref{tab:app-tensors}).
Neither the context projection nor the residual MLP is exported: the runtimes compute the clean score from the experts alone.

\begin{table}[htbp]
  \centering
  \footnotesize
  \caption{Tensors in the bundle, in file order.}
  \label{tab:app-tensors}
  \begin{tabularx}{\textwidth}{@{}l >{\raggedright\arraybackslash}X@{}}
    \toprule
    Section & Tensors (shape) \\
    \midrule
    Outcome head & \texttt{out\_w1} (31, 128), \texttt{out\_b1} (128), \texttt{out\_w2} (128, 64), \texttt{out\_b2} (64), \texttt{out\_w3} (64, 5), \texttt{out\_b3} (5) \\
    Gate & \texttt{gate\_w1} (71, 64), \texttt{gate\_b1} (64), \texttt{gate\_w2} (64, 8), \texttt{gate\_b2} (8) \\
    Embeddings (sorted) & \texttt{emb0} \ldots\ \texttt{emb7}, eight tables of width 8, 968 values \\
    Experts & \texttt{experts} (8, 15) \\
    \midrule
    Total & 19 tensors, 18,893 float32 values \\
    \bottomrule
  \end{tabularx}
\end{table}

The fp32 v2 bundle is 124,690 bytes (metadata 48,814 bytes) and the fp4 bundle 124,688 bytes (metadata 48,812); the JSON twins are 457,671 and 463,184 bytes, and the golden-vector files 18,516 and 18,521 bytes.
For the fp32 file the byte budget reconciles exactly: 8 bytes of magic and metadata length, 48,814 of metadata, 4 for the tensor count, 292 for the nineteen headers (103 name bytes, 57 fixed bytes and 132 shape bytes) and 75,572 for the data ($18{,}893 \times 4$), which sum to 124,690.

NVFP4 weights are not part of the bundle; they travel in a packed sidecar of 29,698 bytes, and the CPU runtimes use fp32 throughout.
Each golden-vector case holds an intake, its eight derived economic quantities, its seven criterion values, its clean score, and three plans with their costs and five predicted outcomes.

\section{Benchmark environment and raw tables}
\label{app:bench}

This appendix is the reference record behind the benchmark of Section~\ref{sec:evaluation-bench}: the reference benchmarks, the burstable-instance record, the invocations that failed and were kept, the GPU environment, peak memory by language and the parity-gate record for the v2.1 bundles.

\subsection{Reference benchmarks}

Table~\ref{tab:app-bench-universal} gives the medians of three runs of the reference benchmarks on the six machines.

\begin{table}[htbp]
  \centering
  \footnotesize
  \caption{Reference benchmarks, medians of three runs. The sysbench cpu figures of the M4 Pro are not comparable with the other rows (see text).}
  \label{tab:app-bench-universal}
  \setlength{\tabcolsep}{4pt}
  \resizebox{\textwidth}{!}{%
  \begin{tabular}{@{}lrrrrrrr@{}}
\toprule
Machine & \multicolumn{2}{c}{sysbench cpu (events/s)} & \multicolumn{2}{c}{7-Zip (MIPS)} & \multicolumn{2}{c}{OpenSSL SHA-256 (MB/s)} & FP64 GEMM \\
\cmidrule(lr){2-3}\cmidrule(lr){4-5}\cmidrule(lr){6-7}
 & 1 thread & all & 1 thread & all & 1 thread & all & GFLOPS (BLAS) \\
\midrule
EC2 t3.small (2 thr.) & 1,038 & 1,647 & 3,255 & 5,408 & 388 & 424 & 58.8 (openblas) \\
EC2 t4g.small (2 thr.) & 2,805 & 5,608 & 3,528 & 7,913 & 1,554 & 3,105 & 36.7 (openblas) \\
EC2 t3.large (2 thr.) & 1,055 & 1,648 & 3,335 & 5,355 & 394 & 423 & 57.6 (openblas) \\
EC2 t4g.large (2 thr.) & 2,804 & 5,607 & 3,533 & 7,844 & 1,553 & 3,097 & 36.5 (openblas) \\
DGX Spark GB10 (20 thr.) & 5,554 & 88,338 & 7,223 & 123,485 & 2,456 & 45,285 & 413.8 (openblas) \\
Apple M4 Pro (12 thr.) & 16,719,173 & 13,665,456 & 9,601 & 99,881 & 3,184 & 29,503 & 720.3 (accelerate) \\
\bottomrule
\end{tabular}
}
\end{table}

The GEMM column is numpy FP64 at $n = 4096$ on the machine's BLAS: Accelerate on the Mac, OpenBLAS on the others.
The large instances reproduce their small siblings' per-core numbers within 2.5 per cent on every reference benchmark, so the 100k and 400k cells measured on them are comparable with the small-instance cells.
The Homebrew build of sysbench for macOS/arm64 optimises the prime loop away, which is why sysbench cpu is not used for normalisation.
The t3.small's two vCPUs are hyper-threads of one physical core, and it shows: all-thread scaling is 1.6$\times$ on sysbench and 1.7$\times$ on 7-Zip, against 2.0 to 2.2$\times$ on the two Graviton2 cores.

\subsection{Burstable-instance record}

Both small instances ran with unlimited CPU credits; monitoring shows the launch credits exhausted and surplus credits accruing (129.9 on the t3.small, 95.4 on the t4g.small), so the measurements are full-speed but not free of the variance of the burstable class.
The four instances ran for 14.2 to 18.2 hours at about 2.82~USD of compute plus about 0.90~USD of surplus credits, and a scripted teardown verified that no tagged instance or volume remained.

\subsection{Failed invocations kept in the record}

Java ran out of heap under the harness default of 1~GiB.
On the DGX, \emph{plan} on synth-100k (fp32 and fp4) and \emph{axes} and \emph{menu} on synth-400k ended with a Java heap-space error; on the Mac, \emph{plan} on synth-100k did the same.
The cause is structural rather than a leak: the Java driver keeps every parsed row as a map, and the planner holds the 81-plan menu of every household.
The cells were re-run with a 4~GiB heap, the failed records stay in the archive, and the harness's heap rule is now 40 per cent of RAM capped at 4~GiB.

The Mac matrix was started three times; one run is kept.
The first overlapped with a rebuild of the runtimes (91 of its results came from drivers rebuilt mid-run) and was discarded wholesale; the second began on battery and was stopped after five minutes; the third, on mains power with no builds or background work, is the one reported, and all 632 of its cells completed with no failures.

\subsection{GPU environment and raw numbers}

The GPU measurements ran in the training container with JAX 0.5.3 on the GB10 (CUDA 13.0, compute capability 12.1); the workload is the outcome head alone with the bundle's fp32 tensors, over an input matrix of 624,078 rows built once outside every timing, with one warm-up and three timed passes per batch size.
The GPU reference GEMM gives 18,015~GFLOPS at $n = 4096$ in true FP32 and 29,171 in TF32; at $n = 8192$, 19,665 and 34,074.
Table~\ref{tab:app-bench-gpu} lists the throughput of the median pass by batch size.

\begin{table}[htbp]
  \centering
  \footnotesize
  \caption{Outcome-head throughput on the GB10 GPU in millions of rows per second, median of three timed passes, by batch size and JAX precision (TF32 fails parity).}
  \label{tab:app-bench-gpu}
  \begin{tabular}{@{}r r r@{}}
    \toprule
    Batch size & FP32 & TF32 \\
    \midrule
    4,096 & 20.5 & 11.6 \\
    8,192 & 52.3 & 22.4 \\
    16,384 & 79.4 & 77.7 \\
    32,768 & 78.0 & 70.6 \\
    65,536 & 54.3 & 60.3 \\
    131,072 & 51.6 & 59.1 \\
    262,144 & 49.8 & 66.8 \\
    524,288 & 47.5 & 65.3 \\
    \bottomrule
  \end{tabular}
\end{table}

Parity against the numpy reference on 100,000 rows: FP32 is bit-identical, with a maximum $|\Delta|$ of 0.0 and 100 per cent of rows within $10^{-3}$.
TF32 has a maximum $|\Delta|$ of 0.0485, a maximum relative error of 101 and only 51.4 per cent of rows within $10^{-3}$; it fails the parity gate and is reported for speed only.

\subsection{Peak memory per language and dataset}

Table~\ref{tab:app-bench-memory} gives the maximum resident set size on the DGX (fp32, one thread).

\begin{table}[htbp]
  \centering
  \footnotesize
  \caption{Peak resident memory in MB on the DGX (fp32, one thread). The three Java columns are the harness default heap of 1~GiB and the re-runs at 4 and 16~GiB; OOM: heap exhausted; a dash means the cell was not re-run at that heap. The 400k \emph{menu} figure at 4~GiB is the full heap plus JVM overhead, not the driver's requirement.}
  \label{tab:app-bench-memory}
  \begin{tabular}{@{}l l r r r r r r@{}}
    \toprule
    Dataset & Mode & C11 & Rust & Java 1~GiB & Java 4~GiB & Java 16~GiB & Python \\
    \midrule
    synth-10k & axes & 47 & 38 & 278 & -- & -- & 67 \\
    synth-10k & menu & 79 & 59 & 538 & -- & -- & 121 \\
    synth-10k & plan & 128 & 166 & 1,104 & -- & -- & 330 \\
    \addlinespace
    synth-100k & axes & 447 & 363 & 957 & -- & -- & 365 \\
    synth-100k & menu & 631 & 536 & 1,184 & 2,272 & -- & 848 \\
    synth-100k & plan & 1,257 & 1,637 & OOM & 4,288 & -- & 2,999 \\
    \addlinespace
    synth-400k & axes & 1,783 & 1,444 & OOM & 3,558 & -- & 1,375 \\
    synth-400k & menu & 2,658 & 2,314 & OOM & 4,459 & 8,955 & 3,953 \\
    \bottomrule
  \end{tabular}
\end{table}

All four drivers keep every parsed row in memory, by contract, and the planners keep the 81-plan menu of every household; the Java driver's map-based rows and boxed plan objects roughly double the footprint.
A 2~GiB instance therefore cannot run the 400k set in any language, so the 100k and 400k cells were re-run on 8~GiB instances of the same core count with a 5~GiB JVM heap.
At a 4~GiB heap, Java \emph{plan} on synth-100k produces the same checksum and plan summary (27,303 households assigned, USD~2,980,005.36 spent) as the other three languages; at 400k the 4~GiB heap is full and the serial collector dominates (1,079 rows/s on the DGX), and with a 16~GiB heap the cell uses about 9~GB and runs at 4,183 rows/s single-threaded and 53,626 on 20 threads, so the collapse was heap size alone.

\subsection{Parity-gate record for the v2.1 bundles}

The v2.1 bundles were gated before any household-data number was quoted, over the grid and population golden sets of 8 and 2,000 cases, fp32 and fp4, and the legacy v2 sets, at the tolerances of Section~\ref{sec:evaluation-bench-parity}, on all six machines; the planner assignments were identical across machines and languages on every golden file (for instance 72 households and 4,994.84 on the fp32 2,000-case population set).
A first attempt on the DGX failed on two test-harness errors, both fixed before the record; the runtimes themselves passed every file at the first attempt.
The one failure the cloud-instance gates found was also in the test suite: on the t3.small (x86-64 with AVX-512) three tests asserted bit equality between the vectorised numpy path and the row-wise path, which holds on every ARM host in the study but not under AVX-512, where the same weighted sums differ by one ulp; those assertions were relaxed to $10^{-12}$ absolute, and the tolerance-based golden gates had passed on that machine throughout.

\end{document}